\documentclass[12pt,a4paper]{article}

\usepackage{amsfonts}
\usepackage{amsmath}
\usepackage{amsbsy}
\usepackage{theorem}
\usepackage{graphicx}
\newtheorem{definition}{Definition}
\newtheorem{theorem}{Theorem}

\newtheorem{lemma}{Lemma}

\newtheorem{remark}{Remark}

\begin{document}
\sf

\title{\bfseries Time-adaptive infinite-dimensional Gaussian process  regression on manifolds
}

\author{M. D. Ruiz-Medina, A.E. Madrid, A. Torres-Signes and J.M. Angulo}

\date{}

 \maketitle

\begin{abstract}
This paper proposes  a new formulation of functional Gaussian Process   regression on manifolds,  based on an Empirical Bayes approach,  in the spatiotemporal random field context. We apply the machinery of  tight Gaussian measures in separable Hilbert spaces, exploiting the invariance property of covariance kernels under the group of isometries of the manifold. The identification via characteristic function of these measures with the  infinite product of one-dimensional Gaussian measures is then obtained,  in terms of the
eigenfunctions of the Laplace--Beltrami operator on the manifold. The involved time-varying angular spectrum constitutes the key tool for dimension reduction in the implementation of this regression approach,  adopting a suitable truncation scheme depending on the functional sample size. The simulation study and synthetic data application  illustrate the performance of the proposed functional regression predictor.
\end{abstract}

\medskip


\section{Introduction}\label{sec1}
Gaussian Process (GP) regression allows high flexibility in nonlinear modelling, being adopted in a  quite extensive variety of theoretical and applied disciplines, including machine learning and statistics (see, e.g.,  \cite{Carlin}; \cite{Deisenroth}; \cite{Diggle13}; \cite{Henning}; \cite{Kaufman10}; \cite{Quinonero05}; \cite{Williams06}; \cite{Rice91}; \cite{Riutort23}; \cite{Wilson14}, among others).
The specification of the   nonlinear random regression function based on a GP is achieved  through their conditional prior Gaussian distribution in a multivariate framework (see, e.g.,  \cite{Neal97}; \cite{Williams06}).  In practice,  to avoid the computational complexity of the  fully Bayesian framework, an Empirical  Bayes approach is implemented to approximate the   hyperparameter vector characterizing the covariance matrix.  The finite-dimensional  mean vector and covariance matrix   encode  prior assumptions, related to the functional relationship,  local regularity, scale properties, memory, among other features.

The main computational drawback of GP regression  lies in the calculation of the posterior, defined in terms of  the inverse of the covariance matrix of the observations. Several approaches have been implemented to solve this issue. Sparse GPs are based on low rank approximation of the covariance matrix. A unified overview of this technique can be found in  \cite{Quinonero05} and  \cite{Williams06}.
Basis function approximation constitutes a competitive alternative, especially  when it is supported by the spectral analysis or series expansion of the GP, including its sparse approximation and variational methods  in the frequency domain (see, e.g., \cite{Adler81}; \cite{Hensman17}; \cite{Hoffmann26};   \cite{Lazaro10}; \cite{RuizMedina2022}).
We also refer to the precision-matrix-based approach frequently arising in spatial data analysis (see, e.g., \cite{Lindgren22}).

 In the present paper, an infinite-dimensional framework is adopted (see  \cite{Riutort23}; \cite{Solin13}; \cite{Solin20}) in the spatiotemporal data analysis context (see \cite{Andersen}; \cite{Carlin}; \cite{Diggle13}, and the references therein).  Specifically, we work under a Functional Gaussian Process  (FGP) model. Particularly, we restrict our attention to time-correlated functional data  supported on a manifold, covering the case of long-range dependence (LRD) in time. The corresponding family of covariance kernels, indexed by time, is assumed to be invariant under the group of isometries of the manifold, defined by a connected and compact two-point homogeneous space (see, e.g., \cite{Gine1975}; \cite{MaMalyarenko}). The eigenfunctions of the Laplace--Beltrami operator on the manifold then provide a spectral diagonalization of these kernels at each time.

 We adopt an Empirical Bayes approach. By applying  Monte Carlo numerical integration over the replicates of the FGP, the loss function, given by the  infinite-dimensional  marginal likelihood of the data,  is obtained conditioning   to the FGP covariance hyperparameters and the variance of the additive observation noise.    Adopting the FGP  prior model, given the observed values of its covariance  hyperparameter vector, to compute the posterior  marginals, the conditional conjugate infinite-dimensional  Gaussian distribution is evaluated on  the  time-varying Maximum Likelihood II (ML-II) estimates of the covariance hyperparameter vector and of the observation noise variance. Specifically,  since conditioning to the  RKHSs is considered in the computation of the marginal posteriors,  normalization is achieved in terms of  the Fredholm determinant.
 The  FGP posterior marginal covariance operators  involve  the computation of  the   inverse of the corresponding covariance operator  of the functional observation process.
 To reduce computational burden arising in the calculation of this predictive distribution, we work in the time-varying purely point spectral domain, exploiting invariance properties of covariance kernels, leading to their diagonalization in terms of the Laplace--Beltrami operator eigenfunctions.
  Note that this domain allows the identification, in distributional sense via the characteristic function, of the measure induced by a Gaussian random element with values in a separable Hilbert space having trace covariance operator, and an infinite product of one-dimensional Gaussian measures on $\left(\mathbb{R}^{\infty},\mathcal{B}(\mathbb{R}^{\infty})\right)$ (see Theorem 1.2.1 and Proposition 1.2.8 in \cite{DaPrato}).
  The logarithm of the functional sample size is usually considered as the value of the  truncation parameter for dimension reduction  in a consistent way, providing a trade-off between computational complexity and accuracy for large functional sample sizes. Thus, we work with the corresponding  finite sequence of truncated spatial  likelihood functions indexed by time.

 To illustrate the implementation of the FGP regression methodology proposed, we focus on the  Gneiting class of spatiotemporal covariance functions restricted to the sphere. Two subfamilies are analyzed in the simulation study in Section  \ref{ss},  displaying different local regularity and memory properties. Particularly, we pay attention to the Cauchy covariance  subfamily displaying LRD in time (see also \cite{Hensman17} for the purely spatial case, where Mat\'ern covariance  kernel is considered). The performance
 of the  posterior functional predictor regarding bias and variability are illustrated. Different truncation strategies depending on the functional sample size are  tested, distinguishing between logarithmic-  and power-law truncation schemes. For large functional sample sizes, a good performance is observed under both truncation  criteria. For small sample sizes, when higher spatial  local singularity and memory in time  are displayed by the spatiotemporal restricted covariance model,    higher truncation orders  are required, leading to a better performance of the  power-law truncation  scheme. A synthetic daily spherical functional dataset, displaying strong temporal correlation,  is generated to analyze the nonlinear interaction between    downward solar radiation flux and atmospheric pressure at high cloud bottom. The 5-fold cross-validation technique  is  implemented to test the accuracy of the proposed time-adaptive Empirical-Bayes-FGP based  posterior predictor, which is compared with a functional linear predictor, non-incorporating the posterior covariance operator structure.

 The remainder  of the paper is structured as follows. In Section \ref{preliminaries}, background,  notation, and the main preliminary elements and results  are  established.
 The proposed time-adaptive Empirical Bayes  FGP (EBFGP) regression   is introduced in Section \ref{EBGPFREGprev}.  The simulation study  in Section \ref{ss} illustrates the interaction between tuning parameters, and local regularity and memory FGP model parameters, testing different  truncation schemes, for  small and large functional datasets.  The  5-fold cross-validation technique  is  implemented in Section \ref{sda} to evaluate the performance of the  posterior   functional predictor of
 downward solar radiation flux from atmospheric pressure at high cloud bottom. Section \ref{FC}, on final comments,  provides some future research lines, and  the computational aspects arising  in the implementation of  the FGP regression technique proposed.

 \section{Preliminaries}
\label{preliminaries}

In the exposition of the preliminary elements, we begin by focusing on basic definitions and results on Gaussian measures on a separable Hilbert space in Section \ref{GMSHS}. Time-varying  pure point  spectral analysis on manifolds is introduced in Section \ref{TVSAM}. The basic elements and notation in the formulation of our  time-adaptive  EBFGP regression approach are provided in Section \ref{PEBGPFREG}.

In what follows we consider that all random variables are defined on the basic probability space $(\Omega,\mathcal{A}, P)$.

\subsection{Gaussian measures in separable Hilbert spaces}
\label{GMSHS}

This section provides the basic definitions and results on Gaussian measures in separable Hilbert spaces applied in the formulation and  implementation of the time-adaptive EBFGP regression approach proposed.

 Hereafter,  $\mu_{\mathcal{R}} $ denotes  a centered non-degenerate Gaussian  measure  in a separable Hilbert space  $\left(\mathbb{H}, \left\langle \cdot,\cdot\right\rangle_{\mathbb{H}}\right)$ with covariance operator $\mathcal{R}$ in the space of positive trace-class operators on  $\mathbb{H}$, denoted as $L_{1}^{+}(\mathbb{H}),$ with $L_{1}(\mathbb{H})$ being the space of trace-class operators on $\mathbb{H}$.

 For a probability measure  $\mu_{\mathcal{R}} $ in $\mathbb{H}$, its characteristic function   is given by  (see, e.g., Section 1.2.1 in \cite{DaPrato})
$$\phi_{\mu_{\mathcal{R}} }(x) = \int_{\mathbb{H}}\exp\left( i\left\langle x,y\right\rangle_{\mathbb{H}}\right)\mu_{\mathcal{R}}(dy),\quad x\in \mathbb{H}.$$
Theorem \ref{thgm} and Definition \ref{def1} below
will be applied in the computation of likelihood functions in our context of EBFGP regression.
Specifically, the next result (see Theorem 1.2.1 in  \cite{DaPrato}) provides the identification via characteristic function  of a tight Gaussian measure $\mu_{a,\mathcal{R}}$ on a separable Hilbert space  $\left(\mathbb{H}, \left\langle \cdot,\cdot\right\rangle_{\mathbb{H}}\right)$, with the infinite product of one-dimensional Gaussian measures, given by  $\widetilde{\mu}$ on $\left(\mathbb{R}^{\infty},\mathcal{B}(\mathbb{R}^{\infty}) \right)$, where $\mathcal{B}(\mathbb{R}^{\infty})$ denotes the Borel $\sigma$-algebra on $\mathbb{R}^{\infty}$.
\begin{theorem}
\label{thgm}
Let $a\in \mathbb{H}$, and $\mathcal{R}\in L_{1}^{+}(\mathbb{H})$.  There exists a unique Gaussian probability measure $\mu_{a,\mathcal{R}}$ on $(\mathbb{H},\mathcal{B}(\mathbb{H}))$, with $\mathcal{B}(\mathbb{H})$ being the Borel  $\sigma$-algebra on the separable Hilbert space
$\left(\mathbb{H}, \left\langle \cdot,\cdot\right\rangle_{\mathbb{H}}\right)$,  such that, for each $h\in \mathbb{H},$
$$\phi_{\mu_{a,\mathcal{R}} }(h) = \int_{\mathbb{H}}\exp\left(i\left\langle h,x\right\rangle_{\mathbb{H}} \right)\mu_{a,\mathcal{R}}(dx)=\exp(i\left\langle a,h\right\rangle_{\mathbb{H}})\exp\left(-\frac{1}{2}\left\langle \mathcal{R}(h),h\right\rangle_{\mathbb{H}} \right).$$ This measure $\mu_{a,\mathcal{R}}$ is the restriction to $\mathbb{H}$ (isometrically identified in the $l^{2}$ sense) of the infinite-product measure $\widetilde{\mu}=\prod_{k=1}^{\infty }\mu_{k}$ on $\left(\mathbb{R}^{\infty},\mathcal{B}(\mathbb{R}^{\infty}) \right)$, with $\mu_{k}(dy)=f_{N_{k}}(y)dy$, being $f_{N_{k}}(y)=\frac{1}{\sqrt{2\pi \lambda_{k}}}\exp\left(-\frac{(y-a_{k})^{2}}{2\lambda_{k}}\right)$, $a_{k}=\left\langle a, e_{k}\right\rangle_{\mathbb{H}},$   $k\geq 1$,  $y\in \mathbb{R}$. The sequence $\left\{\lambda_{k},\ k\geq 1\right\}$
satisfies $\mathcal{R}(e_{k})=\lambda_{k}e_{k}$, $k\geq 1$, for an orthonormal basis of eigenfunctions $\{e_{k},\ k\geq 1\}$ of $\mathbb{H}$.
We respectively refer to $a$ and $\mathcal{R}$ as the functional mean and covariance operator of measure $\mu_{a,\mathcal{R}}$.
\end{theorem}

\begin{remark}
\label{rem1}
As commented, in Theorem \ref{thgm} we have applied the  identification in distributional sense of the functional random variables, inducing the infinite-dimensional Gaussian probability measures $\mu_{a,\mathcal{R}}$ and  $\widetilde{\mu},$ from the equality of  their  characteristic functions. In particular, for $a\underset{\mathbb{H}}{=}0,$
\begin{eqnarray}\phi_{\mu_{0,\mathcal{R}} }(h) &=&  \int_{\mathbb{H}}\exp\left( i\left\langle h,y\right\rangle_{\mathbb{H}}\right)\mu_{\mathcal{R}}(dy)\nonumber\\
&=&\exp\left(-\frac{1}{2}\left\langle \mathcal{R}(h),h\right\rangle_{\mathbb{H}} \right)=\prod_{k\geq 1}\exp\left(-\frac{1}{2}
\lambda_{k}\left(\mathcal{R}\right)[h(\phi_{k})]^{2}\right)\nonumber\\ &=& \phi_{\widetilde{\mu}}(\{h(\phi_{k})\}_{k\geq 1}).\nonumber\\
\label{eqcharact}\end{eqnarray}

\noindent Thus, the probability distribution function $F_{\mu_{0,\mathcal{R}}}(h)$ of  $\mu_{0,\mathcal{R}}$ is given by
\begin{eqnarray}
&&F_{\mu_{0,\mathcal{R}}}(h)=\prod_{k\geq 1}\Phi\left(\frac{h(\phi_{k})}{\sqrt{\lambda_{k}(\mathcal{R})}}\right)>0,
\label{eqPDF}
\end{eqnarray}
\noindent since $\sum_{k\geq 1}1-\Phi\left(\frac{h(\phi_{k})}{\sqrt{\lambda_{k}(\mathcal{R})}}\right)<\infty,$ for  $h\in \mathcal{R}^{1/2}(\mathbb{H}).$
Here,  $\Phi$ denotes the  cumulative probability distribution of a standard normal random variable, and $\mathcal{R}\phi_{k}=\lambda_{k}(\mathcal{R})\phi_{k},$ $k\geq 1$.
\end{remark}

In what follows, we assume that $X\sim \mu_{0,\mathcal{R}}$ is such that   \begin{equation}P\left( X\in \overline{\mathcal{R}^{1/2}(\mathbb{H})}^{\mathbb{H}}\right)=1,\label{eqrkhs}
\end{equation}\noindent  which holds under the condition that the injection $\mathcal{R}^{1/2}(\mathbb{H})\hookrightarrow \mathbb{H}$ is a Hilbert-Schmidt operator. Let $\mathbb{M}_{d}$ be a connected and compact two-point  homogeneous space with  topological dimension $d$  in $\mathbb{R}^{d+1}$, and $d\nu$ let denote the normalized Riemannian measure.
 Theorem \ref{thgm} and Definition \ref{def1} below  will be applied to  the scale of fractional Sobolev spaces constructed from $L^{2}(\mathbb{M}_{d},d\nu).$  Hence,
  we will restrict our attention to the RKHS of $X$ being the fractional  Sobolev space $H^{r}(\mathbb{M}_{d})$  of order $r$ on $\mathbb{M}_{d},$ such that for $\mathbb{H}=H^{s}(\mathbb{M}_{d}),$ with $r-s>d/2,$ condition (\ref{eqrkhs}) holds, with
  $$\overline{\mathcal{R}^{1/2}(\mathbb{H})}^{\mathbb{H}}=\overline{H^{r}(\mathbb{M}_{d})}^{H^{s}(\mathbb{M}_{d})}=H^{s}(\mathbb{M}_{d}).$$
  \noindent See \cite{Triebel78, Triebel97}  on embedding theorems between fractional Sobolev spaces.

In the normalization of the RKHS-restrictions of   infinite-dimensional Gaussian measures  on $\mathcal{B}(\mathbb{H})$, the concept of Fredholm determinant of a trace-class operator  plays a crucial role.
This concept provides  a complex-valued function which generalizes the
determinant of a matrix  to the case of a  covariance operator, as given in the following definition.

\begin{definition}
\label{def1} (See, for example, \cite{Simon},  Chapter 5, pp.47-48,
equation (5.12))  The Fredholm determinant of $A$ is
\begin{equation}
\mathcal{D}(\omega )=\det(I-\omega A)=\exp\left(-\sum_{k=1}^{\infty }%
\frac{\left\|A^{k}\right\|_{L_{1}(\mathbb{H})}}{k}\omega^{k}\right)=\exp\left(-\sum_{k=1}^{\infty
}\sum_{l=1}^{\infty}[\lambda_{l}(A)]^{k}\frac{\omega^{k}}{k}\right),
\label{fdf}
\end{equation}
\noindent for $\omega \in \mathbb{C}$, and $|\omega |\|A\|_{L_{1}(\mathbb{H})}< 1,$ with $\left\|\cdot \right\|_{L_{1}(\mathbb{H})}$ denoting  the norm in the space of trace-class operators.  Note
that $\| A^{m}\|_{L_{1}(\mathbb{H})}\leq \|A\|_{L_{1}(\mathbb{H})}^{m}$, with  $A$ being a  trace-class   operator on a separable Hilbert
space $\left(\mathbb{H}, \left\langle \cdot,\cdot\right\rangle_{\mathbb{H}}\right)$.
\end{definition}
\begin{remark}
As given in (\ref{eqPDF}), the infinite-dimensional Gaussian   likelihood functions, introduced in   Sections \ref{EBGPFREG}, \ref{ss} and \ref{sda},  are computed conditionally to the functional observations in $\mathcal{R}^{1/2}(\mathbb{H})\subset \mathbb{H},$ leading  to the following normalization constant:
    \begin{eqnarray}
    && \int_{\mathbb{H}}\exp\left(\frac{-1}{2}\left\langle \mathcal{R}^{1/2}x, \mathcal{R}^{-1}\mathcal{R}^{1/2}x\right\rangle_{\mathbb{H}}\right)
    \mu_{\mathcal{R}}(dx)\nonumber\\ &&=\prod_{k\geq 1}\int_{\mathbb{R}}\exp\left(\frac{-\lambda_{k}(\mathcal{R})}{2}z_{k}^{2}\right)\mu_{Z_{k}}(dz_{k})\nonumber\\
    &&=\prod_{k\geq 1}\frac{1}{\sqrt{1+\lambda_{k}(\mathcal{R})}}=\mathcal{D}(-1),
    \label{srest}
    \end{eqnarray}
    \noindent  for $\|\mathcal{R}\|_{1}<1,$ where $Z_{k}\sim \mathcal{N}(0,1),$ $k\geq 1,$  where $\mathcal{D}$ is the Fredholm determinant of $\mathcal{R},$  introduced in Definition \ref{def1}, with  $\omega =-1,$ and $A=\mathcal{R}$. In (\ref{srest}),  we have applied Theorem \ref{thgm}, and condition (\ref{eqrkhs}) ensuring that the RKHS is dense in $\mathbb{H},$   allowing  the change of variable $x(\phi_{k})=\left\langle x,\phi_{k}\right\rangle_{\mathbb{H}}=\lambda_{k}z_{k},$ $k\geq 1.$
\end{remark}

 Recall that, along the paper, $\mathbb{M}_{d}$ constitutes an example of a  manifold, with isometrically equivalent  properties to the sphere, locally resembling an Euclidean space.  Examples of Riemannian manifolds of this type are the projective spaces $\mathbb{P}_{d}$ over the fields $\mathbb{R}$ and $\mathbb{C},$ over the skew field $\mathbb{H}$ of quaternions, and over the algebra $\mathbb{O}$ of octonions (see, e.g., Table 1 in \cite{MaMalyarenko}).

\subsection{FGP time-varying angular spectral analysis on manifolds}
\label{TVSAM}

As commented in the Introduction, a key feature for dimension reduction of the EBFGP regression approach presented  consists in working in the time-varying purely point spectral domain, exploiting invariance of covariance kernels involved in the definition of the time-indexed family of integral covariance operators, characterizing the second-order structure of  the  FGP model. This section provides some basic results and elements characterizing this time-varying purely  point spectral domain.

The next lemma provides the Karhunen--Lo\`eve expansion of a mean-square continuous centered spatiotemporal random field on $\mathbb{M}_{d}\times [0,T]$, $T>0$  (see Theorem 1 in the Supplementary Material in  \cite{Ovalle23}). This result is based on the following  diagonal expansion of a spatially  invariant spatiotemporal covariance kernel $C_{\mathbb{M}_{d}}$ on $\mathbb{M}_{d}\times \mathbb{M}_{d}$, in terms of the eigenfunctions  $\left\{ S_{n,j}^{(d)}, j=1,\dots , \Gamma (n,d),\ n\in \mathbb{N}_{0}\right\}$  of the Laplace--Beltrami operator $\Delta_{d}$ on $L^{2}(\mathbb{M}_{d},d\nu)$:
\begin{eqnarray}&&\hspace*{-1cm} C_{\mathbb{M}_{d}}(\mathbf{x},\mathbf{y},|t-s|)= C_{\mathbb{M}_{d}}(\mathbf{x},\mathbf{y},\tau)
  =\sum_{n\in \mathbb{N}_{0}}
B_{n}(\tau )\sum_{j=1}^{\Gamma (n,d)}S_{n,j}^{(d)}(\mathbf{x})S_{n,j}^{(d)}(\mathbf{y}),\label{klexpc2}
\end{eqnarray}
\noindent where\begin{equation}
\int_{\mathbb{M}_{d}}C_{\mathbb{M}_{d}}(\mathbf{x},\mathbf{y},\tau)S_{n,j}^{(d)}(\mathbf{y})d\nu(y)=B_{n}(\tau )S_{n,j}^{(d)}(\mathbf{x}), \ j=1,\dots, \Gamma (n,d),
\label{eieq}
\end{equation}
\noindent and $$\Gamma (n,d)=\frac{(2n+\alpha +\beta +1)\Gamma (\beta +1)\Gamma (n+\alpha+\beta +1)\Gamma (n+\alpha +1)}{\Gamma (\alpha +1)\Gamma (\alpha +\beta +2)\Gamma (n+1)\Gamma(n+\beta +1)}$$

\medskip

\noindent provides the dimension of the $n$th eigenspace of $\Delta_{d},$ corresponding to the eigenvalue  $$\lambda_{n}=- \varepsilon n(\varepsilon n+\alpha +\beta +1),\quad n\in \mathbb{N}_{0},$$
\noindent with $\alpha =(p+q-1)/2$ and $\beta =(q-1)/2,$ being $\varepsilon =2$ for $\mathbb{M}_{d}=\mathbb{P}_{d}(\mathbb{R}),$ and $\varepsilon =1$ otherwise. Here, $p$ and $q$ are the dimensions of some root spaces connected with the Lie algebras of the groups $G$ and $K,$ that respectively represent the connected component of the group of isometries of $\mathbb{M}_{d},$ and the stationary subgroup of a fixed point in $\mathbb{M}_{d}$.
\begin{lemma}
\label{lemmat5ma}
Let $Z_{T}=\{ Z(\mathbf{x},t),\ \mathbf{x}\in \mathbb{M}_{d},\ t\in [0,T]\}$   having covariance kernel  (\ref{klexpc2}),
and
\begin{equation}\sum_{n\in \mathbb{N}_{0}}B_{n}(0)\Gamma (n,d)<\infty.\label{eqtrace}
\end{equation}
\noindent Then,  $Z_{T}$ admits the following orthogonal expansion: For $\mathbf{x}\in \mathbb{M}_{d},$  $t\in [0,T],$

 \begin{eqnarray}
&&Z_{T}(\mathbf{x},t)\underset{L^{2}(\Omega\times\mathbb{M}_{d}\times[0,T],P(d\omega )\otimes d\nu\otimes dt)}{=}\sum_{n\in \mathbb{N}_{0}} \sum_{j=1}^{\Gamma (n,d)}Z_{n,j}(t)S_{n,j}^{(d)}(\mathbf{x}),\nonumber\\
\label{klexp}
\end{eqnarray}

\noindent where the convergence of the series is considered in the norm of the space $L^{2}(\Omega\times\mathbb{M}_{d}\times[0,T],P(d\omega )\otimes d\nu\otimes dt)$. Note that, in the spatiotemporal  component, we have considered convergence in the space $L^{2}\left(\mathbb{M}_{d}\times[0,T],\right.$ \linebreak
 $\left. d\nu\otimes dt\right)$. The sequence
 $\left\{Z_{n,j}(t), \ t\in  [0,T], \ j=1,\dots, \Gamma(n,d), \  n\in \mathbb{N}_{0} \right\}$ is constituted by centered  independent Gaussian processes on $[0,T]$ given by
 \begin{equation}
 Z_{n,j}(t)=\int_{\mathbb{M}_{d}}Z_{T}(\mathbf{y},t)S_{n,j}^{(d)}(\mathbf{y})d\nu(\mathbf{y}),\  t\in  [0,T] , \  j=1,\dots, \Gamma (n,d),\ n\in \mathbb{N}_{0},
 \label{fcA}
 \end{equation}
 \noindent in the mean-square sense, where, for every $t,s\in [0,T],$  $T\geq 2,$ $\mathbb{E}\left[Z_{n,j}(t)Z_{n,j}(s)\right]= B_{n}(|t-s|)=B_{n}(\tau ),$ with $B_{n}(\tau)$ satisfying (\ref{klexpc2})-(\ref{eieq}).
\end{lemma}

 In the next section, we will adopt a parametric framework in the characterization of the time-varying angular spectrum  $\left\{Z_{n,j}(t),\ j=1,\dots,\Gamma(n,d),\ n\in \mathbb{N}_{0},\right.$ \linebreak $\left.\ t\in [0,T]\right\}$ of $Z,$ for implementation of the proposed
 time-adaptive EBFGP regression.

\subsection{Functional observation model in EBFGP regression}
\label{PEBGPFREG}
Assume the  conditional functional observation process $Y_{t,\boldsymbol{\theta }(t),\sigma (t)}$ on the manifold $\mathbb{M}_{d}$ satisfies
\begin{equation}
Y_{t,\boldsymbol{\theta }(t),\sigma (t)}(\mathbf{x})=Z_{t,\boldsymbol{\theta }}(\mathbf{x})+\varepsilon_{t,\sigma(t)}(\mathbf{x}),\quad \mathbf{x}\in \mathbb{M}_{d},\ t\in \mathbb{T},
\label{obGP}
\end{equation}

\medskip

\noindent where $\mathbb{T}= [0,T]$  is the observable temporal interval, and  $\mathbb{D}=\left\{t_{1},t_{2},\dots,t_{n}\right\}\subset \mathbb{T},$ for some $n\in \mathbb{N},$ with $t_{1}=0$ and $t_{n}=T,$ denotes the set of time instants at which the spatial sampling information about the system is updated. We have respectively denoted by   $Z_{\boldsymbol{\theta }}:=\left\{ Z_{t,\boldsymbol{\theta }}, \ t\in  \mathbb{T}\right\}$   and $\varepsilon_{\sigma(t)}:= \left\{ \varepsilon_{t,\sigma(t)}, \  t\in  \mathbb{T}\right\}$ the conditional FGP and  functional Gaussian white noise process,  given the observed values of the random hyperparameters   $\boldsymbol{\theta }$ and  $\sigma^{2}(t)$, $t\in \mathbb{T}$.
Specifically, given the observed trajectory of the stochastic process $\sigma:=\left\{\sigma(t),\ t\in \mathbb{T}\right\},$ with $\mathcal{P}(\sigma(t)\in \Sigma =\mathbb{R}_{+})=1,$ $t\in \mathbb{T},$ consider $\varepsilon_{\sigma(t)}=\left\{ \varepsilon_{t,\sigma(t)}, \  t\in  \mathbb{T}\right\},$ such that
 $$\varepsilon_{t,\sigma(t)}:=\varepsilon_{t}/\sigma(t):\left(\Omega ,\mathcal{A}, P\right)\to \left(L^{2}(\mathbb{M}_{d},d\nu),\left\langle \cdot,\cdot \right\rangle_{L^{2}(\mathbb{M}_{d},d\nu)}\right),$$
   \noindent is a measurable mapping, for each $t\in \mathbb{T}.$ The associated conditional covariance operator family  on $\left(L^{2}(\mathbb{M}_{d},d\nu),\left\langle \cdot,\cdot \right\rangle_{L^{2}(\mathbb{M}_{d},d\nu)}\right)$ is defined as  \begin{equation}\mathcal{R}_{\varepsilon_{t,\sigma(t)}}=E\left[\varepsilon_{t,\sigma(t)}\otimes \varepsilon_{t,\sigma(t)}\right]=\sigma^{2}(t)
   \mathbb{I}_{L^{2}(\mathbb{M}_{d},d\nu)},\quad t\in \mathbb{T},\label{uet}
   \end{equation}
 \noindent with $\mathbb{I}_{L^{2}(\mathbb{M}_{d},d\nu)}$ denoting the identity operator on $\left(L^{2}(\mathbb{M}_{d},d\nu),\left\langle \cdot,\cdot \right\rangle_{L^{2}(\mathbb{M}_{d},d\nu)}\right).$
Specifically, these random objects can be properly introduced in the weak sense, via white noise isometry, considering an arbitrary $L^{2}(\mathbb{M}_{d}, d\nu)$-valued   Gaussian random variable $X$, inducing an infinite-dimensional Gaussian  measure $\mu_{Q}$ on the Borel $\sigma$-algebra $\mathcal{B}(L^{2}(\mathbb{M}_{d}, d\nu))$ of $\left(L^{2}(\mathbb{M}_{d},d\nu),\left\langle \cdot,\cdot \right\rangle_{L^{2}(\mathbb{M}_{d},d\nu)}\right),$
with trace covariance operator $Q.$ Assume that $Q$ is invariant under the group of isometries of $\mathbb{M}_{d}$.  Then, for each $t\in \mathbb{T},$ given the observed value $\sigma(t),$
\begin{equation}\varepsilon_{t,\sigma(t)}\underset{D}{=}\sum_{n\in \mathbb{N}_{0}}\sum_{j=1}^{\Gamma (n,d)}\left\langle \sigma(t) Q^{-1/2}X,  S_{n,j}^{(d)}\right\rangle_{L^{2}(\mathbb{M}_{d}, d\nu)}S_{n,j}^{(d)},\quad t\in \mathbb{T},\label{eq22}
\end{equation}
\noindent where $\underset{D}{=}$ denotes the equality in distributional sense, and, as before,   \linebreak $\left\{ S_{n,j}^{(d)}, j=1,\dots , \Gamma (n,d),\ n\in \mathbb{N}_{0}\right\}$ denotes the system of eigenfunctions of the Laplace--Beltrami operator $\Delta_{d}$ on $\left(L^{2}(\mathbb{M}_{d},d\nu),\left\langle \cdot,\cdot \right\rangle_{L^{2}(\mathbb{M}_{d},d\nu)}\right)$. The observation model (\ref{obGP}) usually arises in the  spherical functional time series framework (see, e.g., \cite{RMCR25} and \cite{RMCR26}).

 Conditionally to the observed value of the  hyperparameter random vector $\boldsymbol{\theta },$ with $\mathcal{P}\left[\boldsymbol{\theta }\in \Theta \right]=1,$ $Z_{\boldsymbol{\theta }}:=\left\{ Z_{t,\boldsymbol{\theta }}, \ t\in  \mathbb{T}\right\}$ is interpreted as a family of stationary time-correlated  $L^{2}(\mathbb{M}_{d},d\nu)$-valued random   parameter sequence,    whose   conditional $L^{2}(\mathbb{M}_{d}, d\nu)$-valued Gaussian marginal probability measures are  respectively  induced by $Z_{t,\boldsymbol{\theta }}:=Z_{t} | \boldsymbol{\theta },$ $t\in  \mathbb{T}.$ Under stationarity in time, the marginal characteristic function is  given by (\ref{eqcharact}), considering the  covariance operator $\mathcal{R}_{0}(\boldsymbol{\theta})=\mathbb{E}\left[Z_{t,\boldsymbol{\theta }}\otimes Z_{t,\boldsymbol{\theta }}\right]$ on  $\left(L^{2}(\mathbb{M}_{d},d\nu),\left\langle \cdot,\cdot \right\rangle_{L^{2}(\mathbb{M}_{d},d\nu)}\right)$ having kernel  $C_{\mathbb{M}_{d}}(\mathbf{x},\mathbf{y},0),$ $\mathbf{x},\mathbf{y}\in \mathbb{M}_{d},$  introduced in  (\ref{klexpc2}).  In particular, from Theorem \ref{thgm}, the infinite-dimensional Gaussian marginals of $Z_{\boldsymbol{\theta}}$ are characterized by the probability measure $\mu_{\mathcal{R}_{0}(\boldsymbol{\theta })}$ on the Borel $\sigma$-algebra $\mathcal{B}\left(L^{2}(\mathbb{M}_{d}, d\nu)\right)$ of $\left(L^{2}(\mathbb{M}_{d},d\nu),\left\langle \cdot,\cdot \right\rangle_{L^{2}(\mathbb{M}_{d},d\nu)}\right),$ whose restriction
to $\mathcal{R}_{0}^{1/2}(\boldsymbol{\theta })[L^{2}(\mathbb{M}_{d}, d\nu)]$ satisfies (\ref{srest}), with
$\mu_{\mathcal{R}_{0}(\boldsymbol{\theta })}=\prod_{n=0}^{\infty}[\mu_{\lambda_{n}(\mathcal{R}_{0}(\boldsymbol{\theta }))}]^{\Gamma (n,d)},$ in the sense of (\ref{eqcharact}), and, for $n\in \mathbb{N}_{0},$ $$\mu_{\lambda_{n}(\mathcal{R}_{0}(\boldsymbol{\theta }))}(dy_{n})=f_{N_{\lambda_{n}(\mathcal{R}_{0}(\boldsymbol{\theta }))}}(y_{n})dy_{n}=\frac{1}{\sqrt{2\pi B_{n}(0,\boldsymbol{\theta })} }\exp\left(-\frac{y_{n}^{2}}{2B_{n}(0,\boldsymbol{\theta })}\right)dy_{n},$$
\noindent where $\lambda_{n}(\mathcal{R}_{0}(\boldsymbol{\theta }))=B_{n}(0,\boldsymbol{\theta }),$ $n\in \mathbb{N}_{0}.$
\noindent  Note that
\begin{equation} Z_{n,j}(t,\boldsymbol{\theta })=\left\langle Z_{t,\boldsymbol{\theta }},S_{n,j}^{(d)}\right\rangle_{L^{2}(\mathbb{M}_{d},d\nu)}\underset{D}{=}\sqrt{B_{n}(0,\boldsymbol{\theta })}\mathcal{Z}_{n,j}(t, \boldsymbol{\theta }),\label{ptvas}\end{equation}
\noindent with  $\mathcal{Z}_{n,j}(t,\boldsymbol{\theta })\sim \mathcal{N}(0,1),$   and, for $t,s\in \mathbb{T},$ $ j=1,\dots,\Gamma (n,d),$  $n\in \mathbb{N}_{0},$ \begin{equation}\mathcal{C}_{\mathcal{Z}_{n,j}}\left(t,s, \boldsymbol{\theta }\right)=\mathbb{E}\left[\mathcal{Z}_{n,j}(t,\boldsymbol{\theta })\mathcal{Z}_{n,j}(s,\boldsymbol{\theta })\right]=\frac{B_{n}(|t-s|,\boldsymbol{\theta })}{B_{n}(0,\boldsymbol{\theta })}=\frac{B_{n}(\tau ,\boldsymbol{\theta })}{B_{n}(0,\boldsymbol{\theta })}.\label{linearcorr}
\end{equation}

 \noindent Thus, the
FGP  $Z_{\boldsymbol{\theta }}$  displays  functional linear correlation in time, characterized by the family of covariance operators $\left\{ \mathcal{R}_{\tau }(\boldsymbol{\theta }),\ \tau \in \mathbb{T}\right\},$ respectively having covariance kernels
\begin{eqnarray} &&\mathbb{E}\left[Z_{t,\boldsymbol{\theta }}\otimes Z_{s,\boldsymbol{\theta }}(\mathbf{x}, \mathbf{y})\right]=C_{\mathbb{M}_{d}}(\mathbf{x},\mathbf{y},|t-s|, \boldsymbol{\theta })\nonumber\\ &&= C_{\mathbb{M}_{d}}(\mathbf{x},\mathbf{y},\tau,\boldsymbol{\theta })
  =\sum_{n\in \mathbb{N}_{0}}
B_{n}(\tau , \boldsymbol{\theta })\sum_{j=1}^{\Gamma (n,d)}S_{n,j}^{(d)}(\mathbf{x})S_{n,j}^{(d)}(\mathbf{y}),\quad \mathbf{x},\mathbf{y}\in \mathbb{M}_{d},\ t,s\in \mathbb{T}.\nonumber\\ \label{klexpc2bb}
\end{eqnarray}

 Note that, for each $t\in \mathbb{T}$, given a realization $\left\{Z_{t,\boldsymbol{\theta }}(\mathbf{x}), \ \mathbf{x}\in \mathbb{M}_{d}\right\}$ of the functional marginal $Z_{t,\boldsymbol{\theta }}$ and of $\sigma^{2}(t)$,
\begin{equation}Y_{t,\boldsymbol{\theta }(t),\sigma (t)}(\mathbf{x}) | Z_{t,\boldsymbol{\theta }}(\mathbf{x}),\sigma (t)\sim \mathcal{N}(Z_{t,\boldsymbol{\theta }}(\mathbf{x}),\sigma^{2}(t)),\ \mathbf{x}\in \mathbb{M}_{d},\ t\in \mathbb{T}.\label{YYY}
\end{equation}

\begin{remark}
We have paid  attention to the stationary FGP framework, where  the covariance operator hyperparameter $\boldsymbol{\theta }$ does not depend on time, ensuring parameter identifiability for these models.
'Beyond stationary-in-time condition of the FGP model, one can assume, as a first approximation for identifiability, that the random hyperparameter vector  $\boldsymbol{\theta }$ becomes a vector stochastic process $\left\{\boldsymbol{\theta }(t),\ t\in \mathbb{T}\right\}$, which is assumed to be   strictly stationary in time. Hence,  its finite-dimensional  vector probability distributions are invariant under translations, leading to a substantial dimension reduction of this more complex random parameter setting.
\end{remark}

\section{The FGP regression methodology}
\label{EBGPFREGprev}

This section describes the two main steps involved in the implementation of the time-adaptive  FGP regression approach proposed, based on sequential  Empirical Bayes parameter estimation. Section \ref{EBGPFREG} introduces the computation of the ML-II estimate $(\widehat{\boldsymbol{\theta }}_{t},\widehat{\sigma }(t))$  of the  hyperparameter vector  $(\boldsymbol{\theta },\sigma (t)).$ We denote by $g(\boldsymbol{\theta },\sigma(t))$ the
prior joint probability density, with absolutely continuous marginals
 $f(\boldsymbol{\theta }),$   and $h(\sigma(t))$ for each time $t\in \mathbb{T}.$
 Section \ref{cigf}
provides the posterior distribution computed from the ML-II parameter estimates through time, applying conditional conjugation in  the framework of infinite-dimensional  Gaussian measures. Finally,  Section \ref {FBV} describes the corresponding  analysis of functional bias and variance.

\subsection{Empirical Bayes approach}
\label{EBGPFREG}

For each $t\in \mathbb{T}$, the ML-II estimate $(\widehat{\boldsymbol{\theta }}_{t},\widehat{\sigma }(t))$  of the  hyperparameter vector  $(\boldsymbol{\theta },\sigma (t))$ satisfies
 \begin{eqnarray}&&(\widehat{\boldsymbol{\theta }}_{t},\widehat{\sigma}(t))=\mbox{arg max}_{(\boldsymbol{\theta },\sigma (t))}p_{Y_{t}(\cdot) | (\boldsymbol{\theta }, \sigma (t))}(y_{t}(\cdot) | \boldsymbol{\theta }, \sigma (t))\nonumber\\ &&=\mbox{arg max}_{(\boldsymbol{\theta },\sigma (t))}\int_{\mathbb{H}}p_{Y_{t}(\cdot) | (Z_{t, \boldsymbol{\theta }}(\cdot),\sigma(t))}(y_{t}(\cdot) | (z_{t,\boldsymbol{\theta }}(\cdot),\sigma (t)) )p_{Z_{t,\boldsymbol{\theta }}(\cdot) | \boldsymbol{\theta }}(dz_{t}(\cdot) | \boldsymbol{\theta }),\nonumber\end{eqnarray}
\noindent where, for each $t\in \mathbb{T},$  $p_{Y_{t}(\cdot) | (\boldsymbol{\theta }, \sigma(t))}(y_{t}(\cdot) | (\boldsymbol{\theta }, \sigma (t)))$ denotes the marginal likelihood of the data    $Y_{t}(\cdot) | (\boldsymbol{\theta }, \sigma (t)),$ characterized by the covariance operator
 \begin{eqnarray}&&\hspace*{-2.5cm} \mathbb{E}\left[Y_{t}(\cdot) | (\boldsymbol{\theta }, \sigma (t))\otimes Y_{t}(\cdot) | (\boldsymbol{\theta }, \sigma (t)) \right]=\mathcal{R}_{0}(\boldsymbol{\theta })+\sigma^{2}(t)\mathbb{I}_{L^{2}(\mathbb{M}_{d},d\nu)}.\nonumber\end{eqnarray}
\noindent
In practice,  for each $t\in \mathbb{T}$, a truncated version of the  integral
\begin{equation}\int_{\mathbb{H}}p_{Y_{t}(\cdot) | Z_{t}(\cdot),\sigma(t)}(y_{t}(\cdot) | (z_{t}(\cdot),\sigma (t)) )p_{Z_{t}(\cdot) | \boldsymbol{\theta }}(dz_{t}(\cdot) | \boldsymbol{\theta })\label{eqmca}
\end{equation}
\noindent is computed by projection, applying  Monte Carlo numerical integration, from the generated replicates  of  $Z_{t,\boldsymbol{\theta }},$  based on  its truncated   Karhunen-Lo\'eve expansion (see  equations (\ref{klexp}), (\ref{obGP}) and (\ref{ptvas})). In the  simulation study in Section
\ref{ss}, and in the  synthetic data  application
in Section \ref{sda}, we apply this numerical approximation, keeping in mind that $Y_{t}(\cdot) | Z_{t,\boldsymbol{\theta }}(\cdot),\sigma(t)\sim \mathcal{N}\left(Z_{t,\boldsymbol{\theta }}(\cdot), \sigma^{2}(t)\mathbb{I}_{L^{2}(\mathbb{M}_{d},d\nu)}\right),$ and $Z_{t,\boldsymbol{\theta }}\sim \mathcal{N}(0,\mathcal{R}_{0}(\boldsymbol{\theta })).$  Hence, at each time $t\in \mathbb{T},$ the marginalization computed in (\ref{eqmca}) is updated, in terms of  the observed value $\sigma (t)$ and the generated replicates of $Z_{t,\boldsymbol{\theta }},$ in the  support of $p_{Z_{t}(\cdot) | \boldsymbol{\theta }},$ defining the conditional mean of $Y_{t}(\cdot) | (Z_{t,\boldsymbol{\theta }}(\cdot),\sigma(t)).$

\subsection{Conjugate infinite-dimensional Gaussian families}
\label{cigf}

For each $t\in \mathbb{T},$ given the values of the ML-II estimate $(\widehat{\boldsymbol{\theta }}_{t},\widehat{\sigma }(t)),$ the Gaussian posterior $p_{Z_{t} | (\widehat{\boldsymbol{\theta}}_{t},\widehat{\sigma }(t), Y_{t})}(z_{t} | (\widehat{\boldsymbol{\theta}}_{t},\widehat{\sigma} (t)), y_{t})$  is characterized by the  posterior mean $\mathbb{E}\left[Z_{t} | (\widehat{\boldsymbol{\theta }}_{t},\widehat{\sigma }(t), Y_{t})\right]$ and covariance operator $\mathcal{R}^{Z | (\widehat{\boldsymbol{\theta }}_{t},\widehat{\sigma} (t), Y_{t})}(\widehat{\boldsymbol{\theta }}_{t},\widehat{\sigma }(t))$:
\begin{eqnarray}
&&\hspace*{-0.7cm}\mathbb{E}\left[Z_{t} | (\widehat{\boldsymbol{\theta }}_{t},\widehat{\sigma }(t), Y_{t})\right]=\mathcal{R}^{Z,Y}_{t}(\widehat{\boldsymbol{\theta }}_{t})[\mathcal{R}^{Y}_{t}(\widehat{\boldsymbol{\theta }}_{t},\widehat{\sigma }(t))]^{-1}Y_{t},\quad t\in \mathbb{T}\nonumber\\
&&\hspace*{-0.5cm}\mathcal{R}^{Z | (\widehat{\boldsymbol{\theta }}_{t},\widehat{\sigma} (t), Y_{t})}(\widehat{\boldsymbol{\theta }}_{t},\widehat{\sigma }(t))=\mathcal{R}^{Z}_{t}(\widehat{\boldsymbol{\theta }}_{t})\nonumber\\ &&\hspace*{1.5cm}-\mathcal{R}^{Z,Y}_{t}(\widehat{\boldsymbol{\theta }}_{t}) [\mathcal{R}^{Y}_{t}(\widehat{\boldsymbol{\theta }}_{t},\widehat{\sigma} (t))]^{-1}\mathcal{R}^{Y,Z}_{t}(\widehat{\boldsymbol{\theta }}_{t}),\ t\in \mathbb{T}.\nonumber\\
\label{eqccong}
\end{eqnarray}

Applying Spectral Theorem on spectral calculus (see, e.g., pp. 112--140 of \cite{DautrayLions85}), equation (\ref{eqccong}) can be rewritten, by projection onto the orthonormal  basis $\left\{ S_{n,j}^{(d)}, \  j=1,\dots , \Gamma (n,d),\ n\in \mathbb{N}_{0}\right\}$ and the spectral kernel $ \sum_{n\in \mathbb{N}_{0}}\sum_{j=1}^{\Gamma(n,d)}S_{n,j}^{(d)}\otimes S_{n,j}^{(d)},$ as follows:
\begin{eqnarray}
&&\hspace*{-0.7cm}\mathbb{E}\left[Z_{n,j}(t,\widehat{\boldsymbol{\theta }}_{t}) | (\widehat{\boldsymbol{\theta }}_{t},\widehat{\sigma }(t), Y_{n,j}(t,\widehat{\boldsymbol{\theta }}_{t}))\right]=B_{n}(t,\widehat{\boldsymbol{\theta }}_{t})\left[B_{n}(0,\widehat{\boldsymbol{\theta }}_{t})\right.\nonumber\\
&&\hspace*{2.5cm}+\left.\widehat{\sigma}^{2}(t)\right]^{-1}Y_{n,j}(t,\widehat{\boldsymbol{\theta }}_{t}), \ j=1,\dots,\Gamma (n,d),\ n\in \mathbb{N}_{0},\  t\in \mathbb{T}\nonumber\\
&&\hspace*{-0.5cm}\lambda_{n}\left(\mathcal{R}^{Z | (\widehat{\boldsymbol{\theta }}_{t},\widehat{\sigma} (t), Y_{t})}(\widehat{\boldsymbol{\theta }}_{t},\widehat{\sigma }(t))\right)=B_{n}(t,\widehat{\boldsymbol{\theta }}_{t})\nonumber\\ &&\hspace*{1.5cm}-B_{n}(t,\widehat{\boldsymbol{\theta }}_{t}) [B_{n}(0,\widehat{\boldsymbol{\theta }}_{t})+\widehat{\sigma}^{2}(t)]^{-1}B_{n}(t,\widehat{\boldsymbol{\theta }}_{t}),\ n\in \mathbb{N}_{0},\ t\in \mathbb{T},\nonumber\\
\label{eqccong2}
\end{eqnarray}
\noindent where, for  each $t\in \mathbb{T}$, $\left\{\lambda_{n}\left(\mathcal{R}^{Z | (\widehat{\boldsymbol{\theta }}_{t},\widehat{\sigma} (t), Y_{t})}(\widehat{\boldsymbol{\theta }}_{t},\widehat{\sigma }(t))\right),\ n\in \mathbb{N}_{0}\right\}$ denotes the angular spectrum
of the posterior covariance operator   $\mathcal{R}^{Z | (\widehat{\boldsymbol{\theta }}_{t},\widehat{\sigma} (t), Y_{t})}(\widehat{\boldsymbol{\theta }}_{t},\widehat{\sigma }(t)),$ which characterizes  the   corresponding posterior Gaussian measure, and satisfies
$$\sum_{n\in \mathbb{N}_{0}}\Gamma (n,d)\lambda_{n}\left(\mathcal{R}^{Z | (\widehat{\boldsymbol{\theta }}_{t},\widehat{\sigma} (t), Y_{t})}(\widehat{\boldsymbol{\theta }}_{t},\widehat{\sigma }(t))\right)<\infty .$$

\subsection{Functional bias and variance analysis}
\label{FBV}
From equation (\ref{eqccong}),  and  triangle inequality, the following almost surely (a.s.) inequality is considered, for each $t\in \mathbb{T}:$
 \begin{eqnarray}
&&\left\|Z_{t,\boldsymbol{\theta }_{0}}-\mathbb{E}\left[Z_{t} | (\widehat{\boldsymbol{\theta }}_{t},\widehat{\sigma }(t), Y_{t})\right]
\right\|_{L^{2}(\mathbb{M}_{d},d\nu)}\nonumber\\
&&\leq \left\|Z_{t,\boldsymbol{\theta }_{0}}-\mathcal{R}^{Z,Y}_{t}(\widehat{\boldsymbol{\theta }}_{t})[\mathcal{R}^{Y}_{t}(\widehat{\boldsymbol{\theta }}_{t},\widehat{\sigma }(t))]^{-1}Z_{t,\boldsymbol{\theta }_{0}}
\right\|_{L^{2}(\mathbb{M}_{d},d\nu)}\nonumber\\
&&+\left\|\mathcal{R}^{Z,Y}_{t}(\widehat{\boldsymbol{\theta }}_{t})[\mathcal{R}^{Y}_{t}(\widehat{\boldsymbol{\theta }}_{t},\widehat{\sigma }(t))]^{-1}Z_{t,\boldsymbol{\theta }_{0}} -\mathbb{E}\left[Z_{t} | (\widehat{\boldsymbol{\theta }}_{t},\widehat{\sigma }(t), Y_{t})\right]\right\|_{L^{2}(\mathbb{M}_{d},d\nu)}\nonumber\\
&&=S_{1}+S_{2},\label{decbv}
 \end{eqnarray}
  \noindent where  $\boldsymbol{\theta }_{0}$ denotes the theoretical value,  and  $S_{1}$ provides the $L^{2}(\mathbb{M}_{d},d\nu)$-norm of the  bias term in the approximation of    $Z_{t,\boldsymbol{\theta }_{0}}$  by the posterior predictor $\mathbb{E}\left[Z_{t} | (\widehat{\boldsymbol{\theta }},\widehat{\sigma }(t), Y_{t})\right]$.
  Here,  $\mathbb{E}\left[S_{2}^{2}\right]$ provides residual variability.

From (\ref{eqccong2}), for each $n\in \mathbb{N}_{0},$

\begin{eqnarray}
&&\lambda_{n}\left(\mathcal{R}^{Z | (\widehat{\boldsymbol{\theta }}_{t},\widehat{\sigma} (t), Y_{t})}(\widehat{\boldsymbol{\theta }}_{t},\widehat{\sigma }(t))\right)= B_{n}(t,\widehat{\boldsymbol{\theta }}_{t}) \left(1-
\frac{B^{2}_{n}(t,\widehat{\boldsymbol{\theta }}_{t})}{[B_{n}(0,\widehat{\boldsymbol{\theta }}_{t})+\widehat{\sigma}^{2}(t)]B_{n}(t,\widehat{\boldsymbol{\theta }}_{t})}\right),\nonumber\\ \label{eqfvdtvs}
  \end{eqnarray}
\noindent leading to the usual one-dimensional variability decomposition formula in the linear case.

\section{Simulation study}
\label{ss}

In this section, we illustrate the performance of time-adaptive EBFGP regression, from  large and small functional sample sizes, under sparsely and densely discretely observed spherical functional data, adopting different truncation schemes depending on the functional sample size.  We pay special attention  to the analysis of the interaction between  these  tuning parameters and the local spatial  regularity and memory in time of the spherical functional time series models analyzed.

Functional observations in equation (\ref{obGP}) are drawn from a FGP prior, characterized by a  spatiotemporal covariance kernel  in the Gneiting class   (see \cite{Gneiting02}),
\begin{equation}C(\left\Vert \mathbf{z}\right\Vert ,\tau )=\frac{\widetilde{\sigma}^{2}}{[\psi (\tau ^{2})]^{d/2}}
\varphi \!  \left( \frac{\left\Vert \mathbf{z}\right\Vert ^{2}}{\psi (\tau ^{2})}\right),
\text{ }\widetilde{\sigma} ^{2}\geq 0,\text{ }(\mathbf{z},\tau )\in \mathbb{R}^{3}\times \mathbb{R},\label{eqgc}
\end{equation}
\noindent  restricted to the unit sphere $\mathbb{S}_{2}(1)$ in $\mathbb{R}^{3}$,
 \begin{eqnarray}&&
  \hspace*{-0.7cm} C\left( 2\sin \left( \frac{\theta }{2}\right),\tau \right)=\frac{\widetilde{\sigma}^{2}}{[\psi (\tau ^{2})]^{d/2}}
\varphi \! \left( \frac{\left[2\sin \left( \frac{\theta }{2}\right)\right]^{2}}{\psi (\tau ^{2})}\right),\
\widetilde{\sigma} ^{2}\geq 0,\ \theta \in [0,\pi],\ \tau \in \mathbb{R},\nonumber\\
  \label{rgneiting}
  \end{eqnarray}
\noindent where  $\varphi $ is a completely monotone function, and $\psi $ is
a positive function  with  completely monotone derivative.
 Specifically, for $\mathbf{x}, \mathbf{x}^{\prime}\in \mathbb{S}_{2}(1)$, the restriction (\ref{rgneiting}) is obtained from   (\ref{eqgc}), by considering  $\mathbf{z}=\mathbf{x}-\mathbf{x}^{\prime }$,  and  $\Vert \mathbf{x}-\mathbf{x}^{\prime }\Vert =2\sin \left( \frac{\theta }{2}\right)$,
 with $\theta $ now denoting the angle between $\mathbf{x}$ and $\mathbf{x}^{\prime }$, and $\|\cdot\|$  being the Euclidean distance.
In particular, we consider the restriction to $\mathbb{S}_{2}(1)\subset \mathbb{R}^{3}$  of the following two subfamilies of covariance functions   in the Gneiting class:

\begin{eqnarray} &&\varphi (u)=\frac{1}{(1+cu^{\gamma })^{\nu }},\quad u>0,c>0,\ 0<\gamma \leq 1,\ \nu >0\nonumber\\
&&\psi (u)=(1+au^{\alpha })^{\beta },\quad u\geq 0, a>0,\ 0<\alpha \leq 1,\ 0<\beta \leq 1,\ \nonumber\\
\label{A1}
\end{eqnarray}

\noindent and

\begin{eqnarray}
\varphi (u)&=&\left( 2^{\varpi -1}\Gamma \left( \varpi \right) \right) ^{-1}\left(
cu^{1/2}\right) ^{\varpi }K_{\varpi }\left( cu^{1/2}\right) ,\text{ } u>0, \text{ } c>0,\text{ }%
\varpi >0\nonumber\\
\widehat{\varphi }(\lambda )&=& \mathcal{M}
 \left( c^{2}+\left\Vert
\lambda \right\Vert ^{2}\right) ^{-\left( \varpi  +\frac{3}{2}\right) },\quad \lambda
\in \mathbb{R}^{3},\quad \mathcal{M}>0\nonumber\\
\psi (u)&=&(1+au^{\alpha })^{\beta },\text{ }u\geq 0, \text{ }a>0,\text{ }0<\alpha \leq 1,\text{
}0<\beta \leq 1.\nonumber\\
\label{A2}
\end{eqnarray}

Local spatial regularity is respectively reflected by the  random   parameters $(\gamma ,\nu )$ and $\varpi $ in subfamilies (\ref{A1})
and (\ref{A2}), and memory in time is reflected by the vector parameter $(\alpha, \beta)$ in both subfamilies.
In our simulation study, the remaining parameters are considered to be degenerated taking the  value 1.

The  diagonal coefficients $\left\{ B_{n}(\tau ), \ n\in \mathbb{N}_{0}\right\}$ are computed by  the  projection of the restricted covariance function (\ref{rgneiting})   onto the  selected elements of the   spherical harmonic basis $\left\{ S_{n,j}^{(2)}, \  j=1,\dots , \Gamma (n,2),\ n\in \mathbb{N}_{0}\right\}.$ Both, the restricted covariance function (\ref{rgneiting}) and the selected elements of the   spherical harmonic basis, are  evaluated at the same regular  spherical grid. We assume axial or zonal isotropy in the generations of the FGP, reflected in the choice of a finite set of  zonal functions, corresponding to the  finite set of eigenspaces of the Laplace--Beltrami operator selected, in the finite-dimensional approximations
  involved in the simulation study. Note that the number of zonal functions depends on the functional sample size through a logarithmic or power function.  The resulting axial or zonal isotropy allows to illustrate the performance of our approach under weaker invariance conditions, leading to a substantial dimension reduction.

\subsubsection*{Generation of priors of hyperparameters}
\begin{figure}[h]
\begin{center}
\includegraphics[width=3.5cm,height=2.5cm]{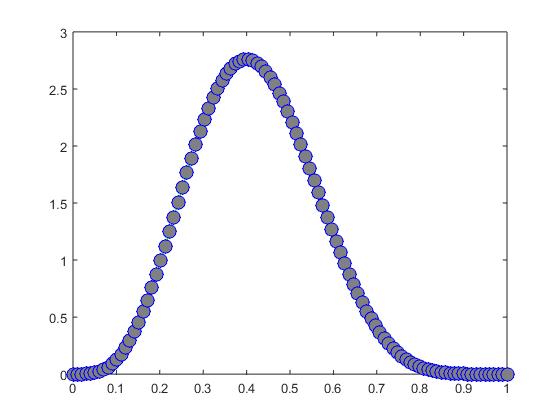}
\includegraphics[width=3.5cm,height=2.5cm]{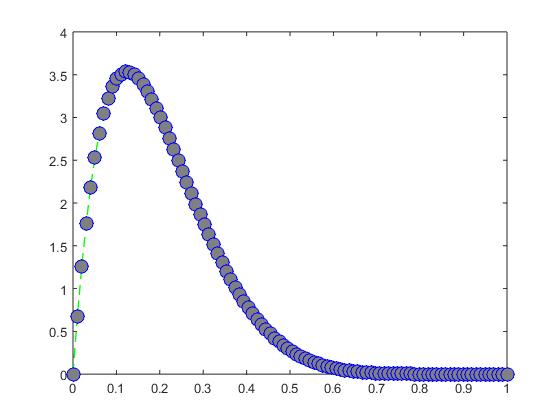}
\includegraphics[width=4.5cm,height=2.5cm]{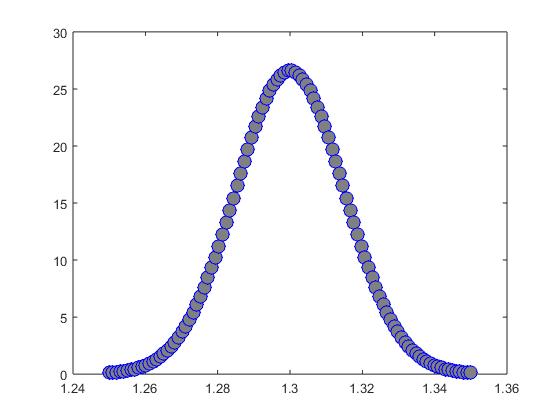}
\includegraphics[width=3.5cm,height=2.5cm]{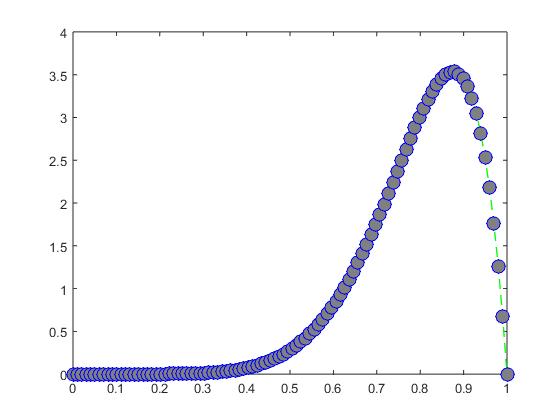}
\includegraphics[width=3.5cm,height=2.5cm]{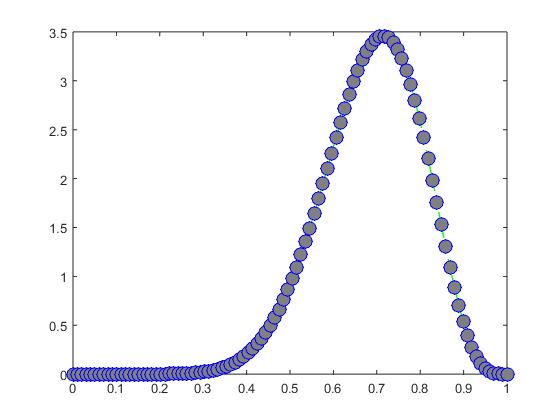}
\includegraphics[width=3.5cm,height=2.5cm]{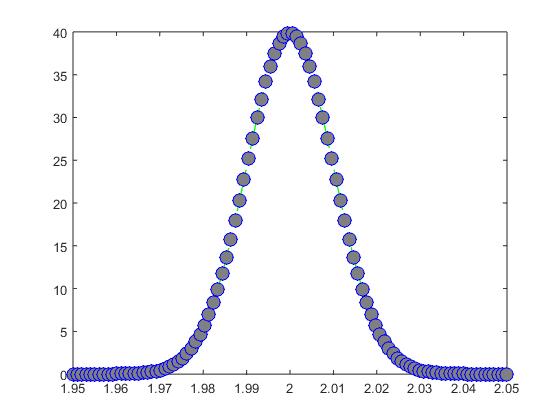}
\end{center}
\caption{\scriptsize{Informative priors for hyperparameters characterizing spatiotemporal covariance function subfamilies 1 and 2. Regularity hyperparameter priors for $\gamma$ (top-left), $\nu$ (top-center) and $\varpi$ (top-right). Memory and noise hyperparameter priors for $\beta$ (bottom-left), $\alpha$ (bottom-center), and $\sigma (t)$ (bottom-right)}}
 \label{F1P}
\end{figure}

  Samples of the priors of the independent components of the  hyperparameter vector $\boldsymbol{\theta },$ as well as of the prior of the parameter $\sigma^{2}(t)$, are generated. Our choice of  the supports and shapes of the hyperparameter  informative priors is based on   a  previous  exploratory data analysis, where bias and variability properties of the posterior predictor are tested,  under different   hyperparameter scenarios of the $\mathbb{S}_{2}(1)$-restricted  subfamilies (\ref{A1}) and (\ref{A2}).
  Figure \ref{F1P} displays the hyperparameter  priors in the beta  family for $\gamma\sim \mbox{Beta} (5,7)$, $\nu \sim \mbox{Beta}(2,8)$, in equation  (\ref{A1}), and     $\varpi \sim N(1.3,0.015)$, in equation   (\ref{A2}). In both subfamilies,    $\beta
 \sim \mbox{Beta} (8,2)$, $\alpha \sim \mbox{Beta}(11,5)$, and  $\sigma (t) \sim \mathcal{N}(1/4,0.01)$ (truncated at 0) are considered.

\subsection{Subfamily 1}

\begin{figure}[h]
\begin{center}
\includegraphics[width=4cm,height=6.5cm]{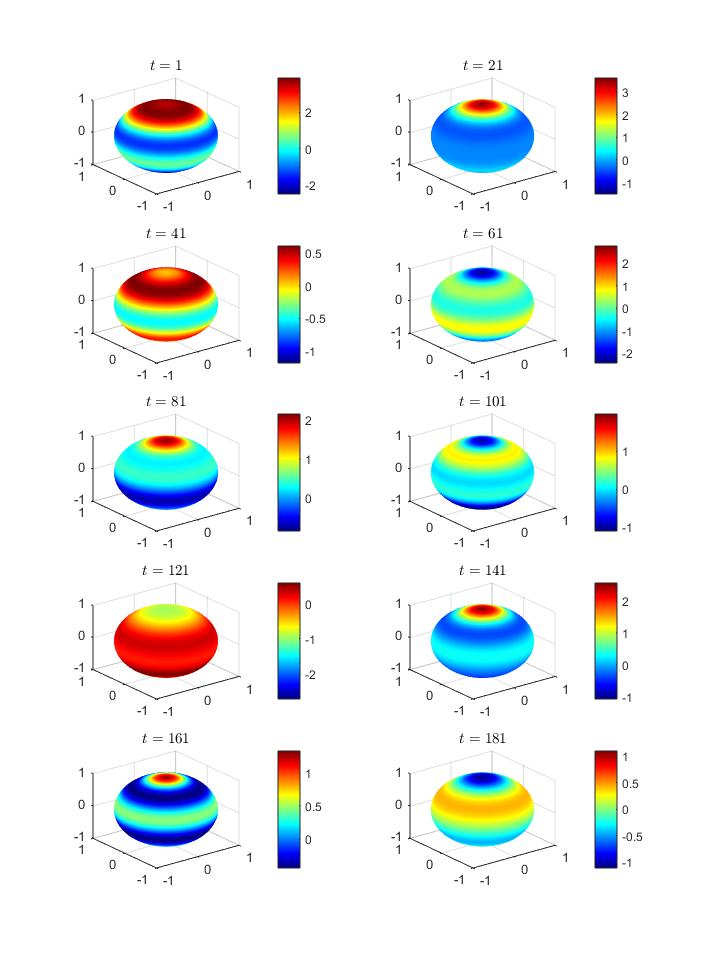}
\includegraphics[width=4cm,height=6.5cm]{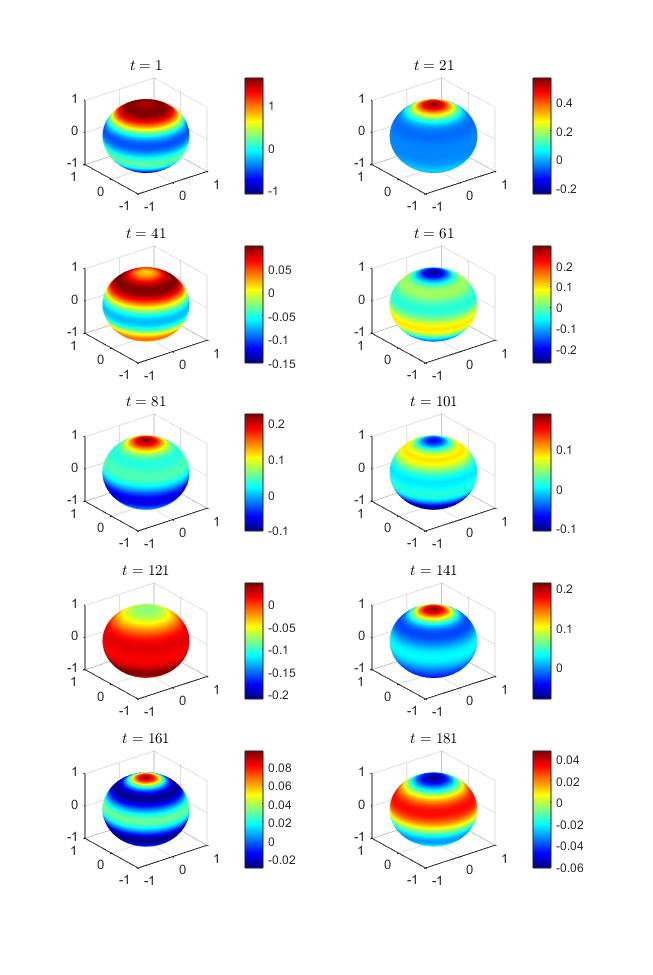}
\includegraphics[width=4cm,height=6.5cm]{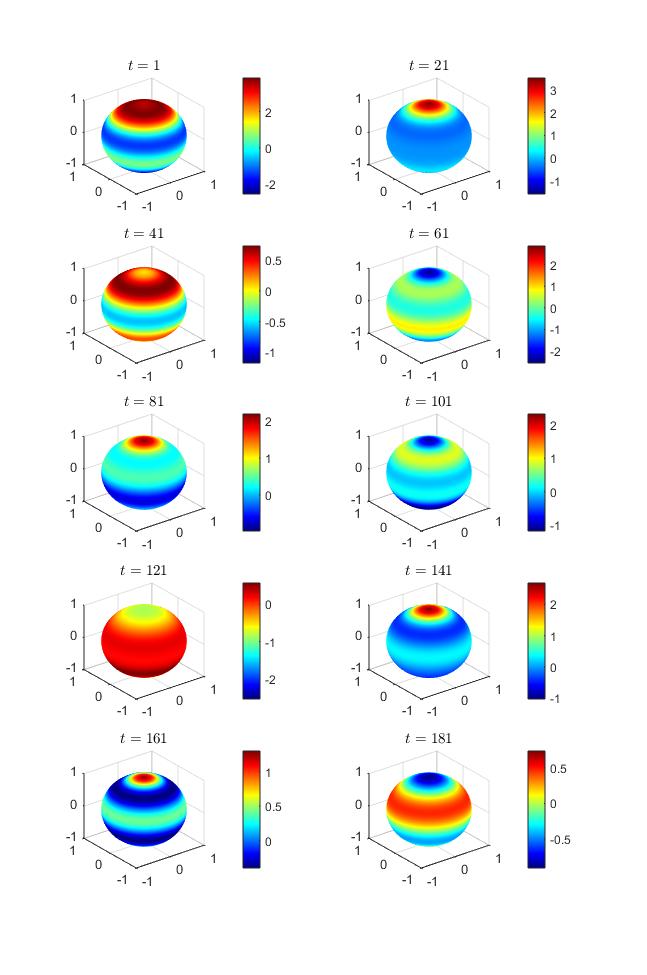}
\end{center}
\caption{\scriptsize{\emph{Subfamily 1}. Realization of conditional  FGP spherical functional time series model (left), spherical functional posterior mean (center), and observation spherical functional time series model  (right), $\mathbb{D}=\left\{1, 21, 41, 61, 81, 101, 121, 141, 161, 181\right\}\subset \mathbb{T}$ }}
 \label{F2P}
\end{figure}

Figure \ref{F2P}  displays, at the left,  one  realization of FGP prior in $\mathbb{S}_{2}(1)$ over   $\mathbb{D}=\left\{1, 21, 41, 61, 81, 101, 121, 141, 161, 181\right\}\subset \mathbb{T}$ in subfamily  (\ref{A1}). The corresponding spherical  functional observation affected by additive noise with intensity $\sigma (t)$ is plotted at the right. Both of them are
 generated conditionally to
  the  covariance   hyperparameter values.

In Figure \ref{F2P} at the center,  the posterior spherical functional mean  over  $\mathbb{D}$  is also displayed. As given in Section \ref{EBGPFREGprev}, the posterior is  computed conditionally to  the ML-II estimates of the hyperparameter vector
$(\gamma,\nu,\alpha ,\beta , \sigma (t))$ obtained,   based on $R=300$ replicates,  considering the sample sizes $T=300$ and $M=50$, with  spatial discretization  given by  $N=150$   spherical  nodes. Note that, here, we have implemented a power-function truncation scheme  $TR=[T^{\varrho}]_{-}$, $\varrho=1/2.45$,  with $[\cdot]_{-}$ denoting the integer part.

\begin{figure}[h]
\begin{center}
\vspace*{0.5cm}
\includegraphics[width=4.5cm,height=6cm]{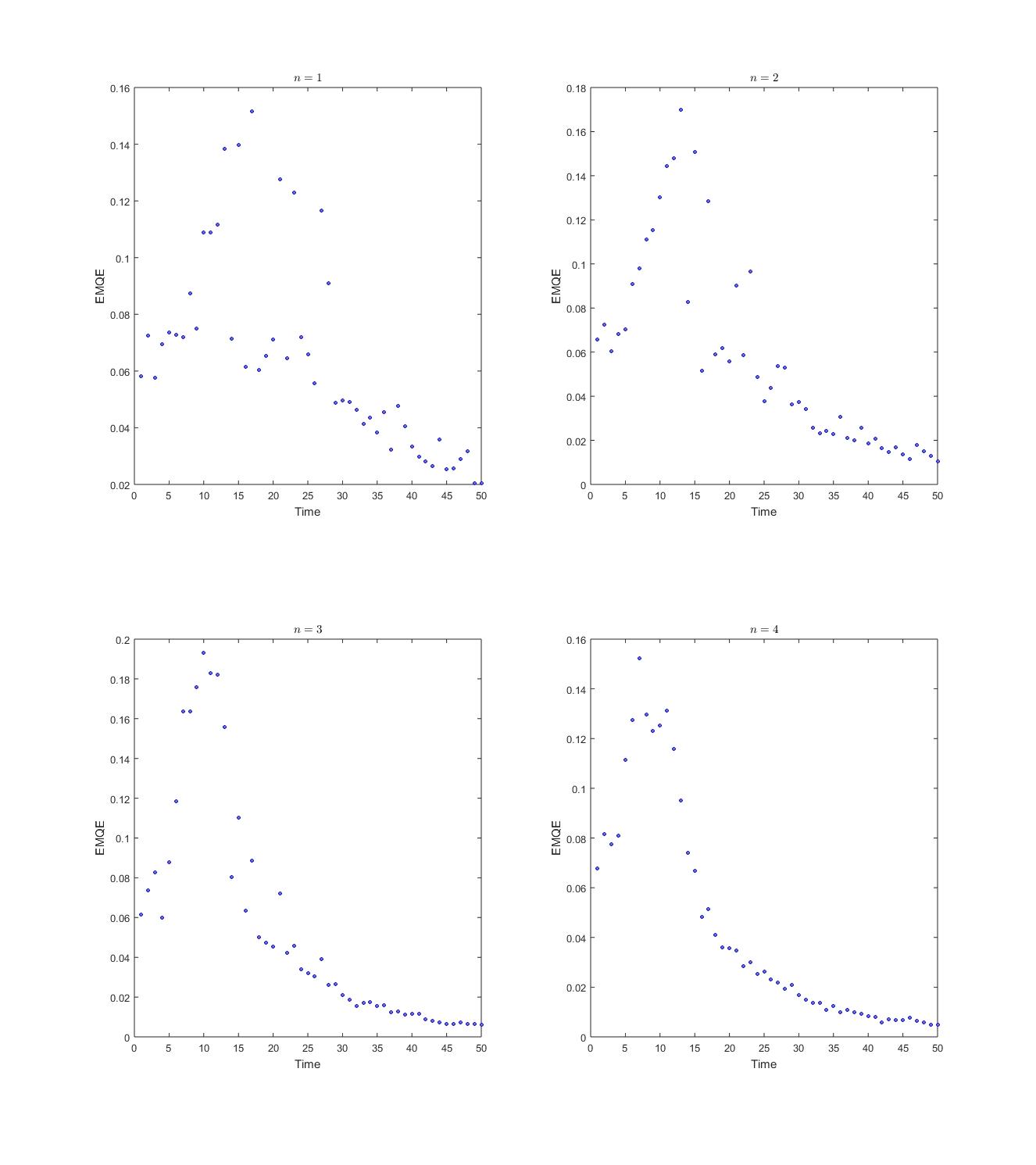}
\includegraphics[width=4.5cm,height=6cm]{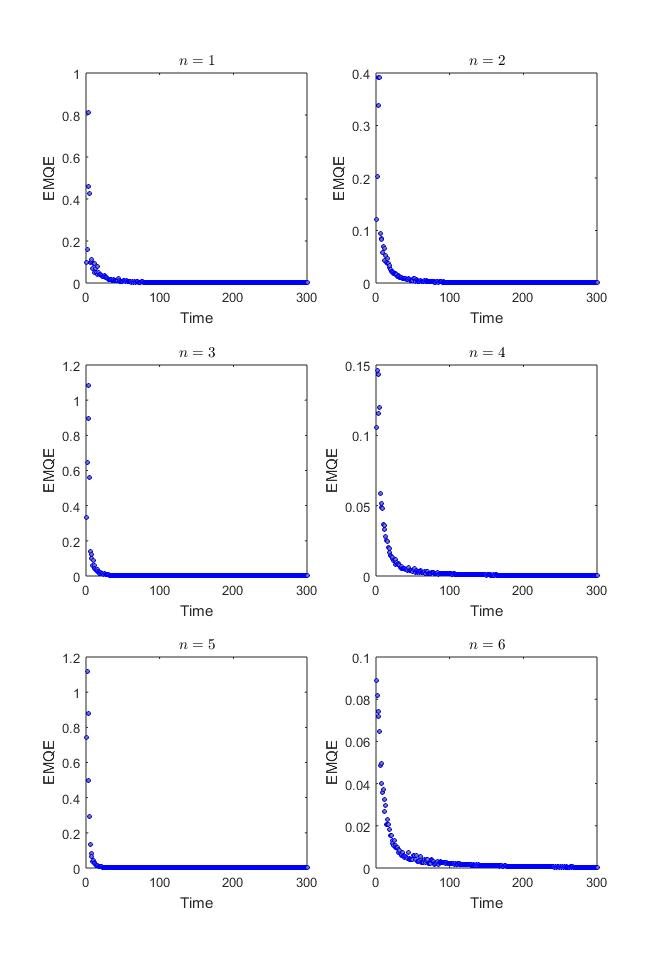}
\includegraphics[width=4.5cm,height=6cm]{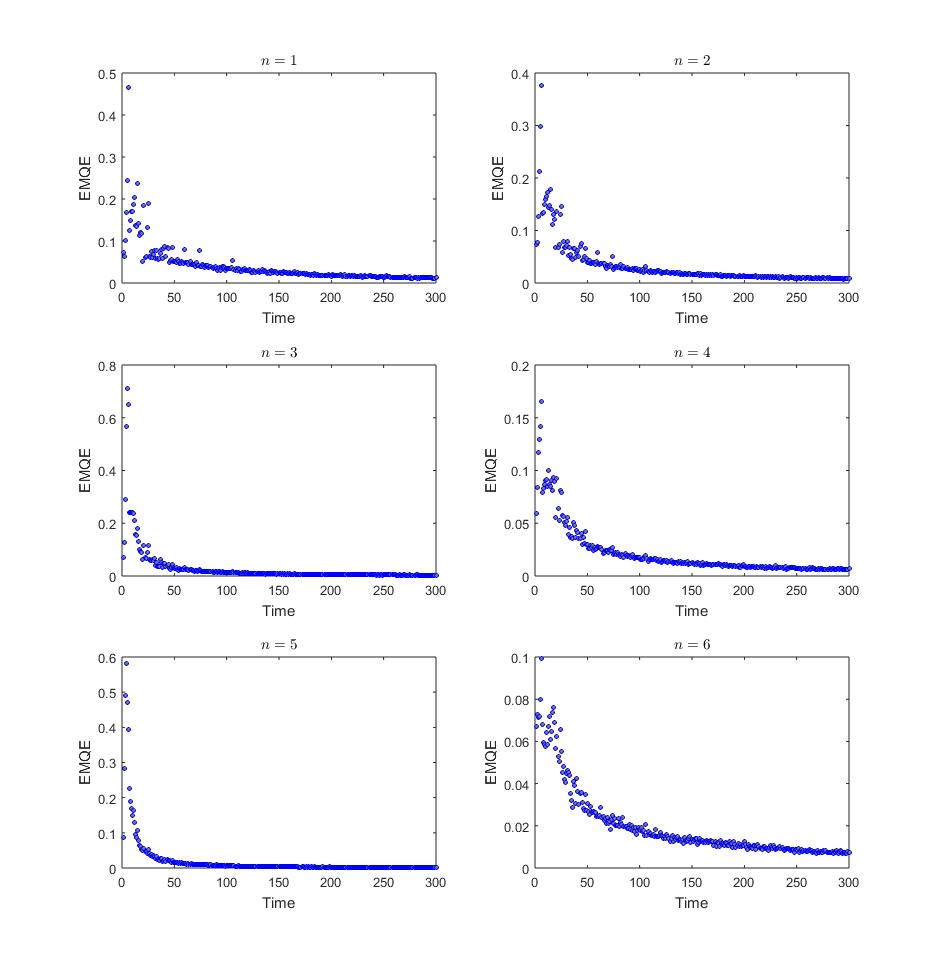}
\includegraphics[width=4.5cm,height=6cm]{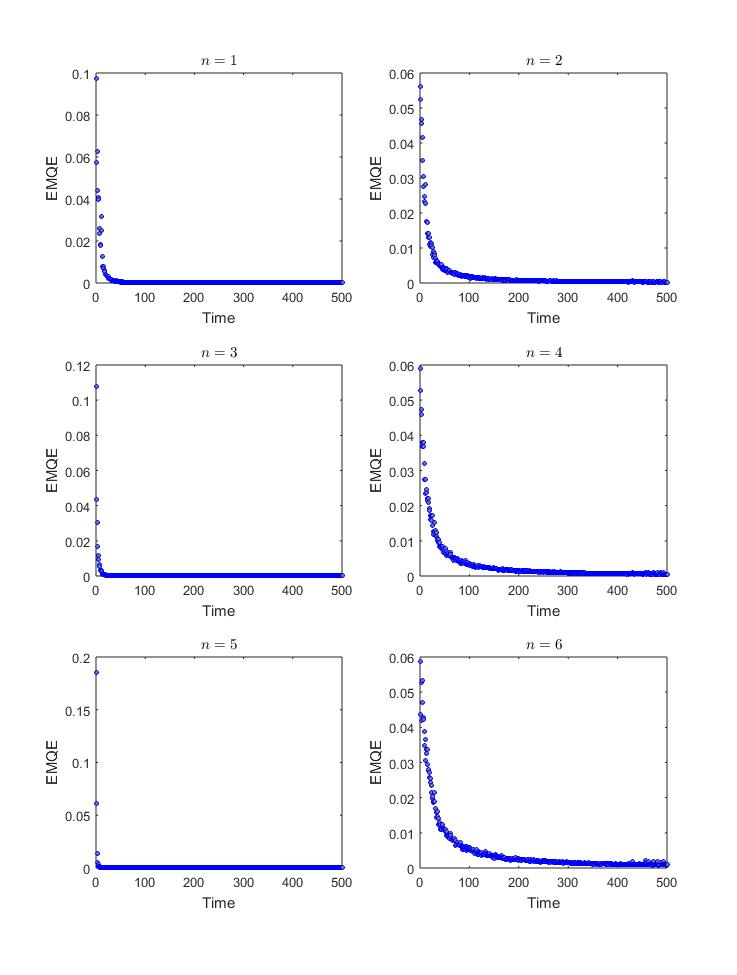}
\includegraphics[width=4.5cm,height=6cm]{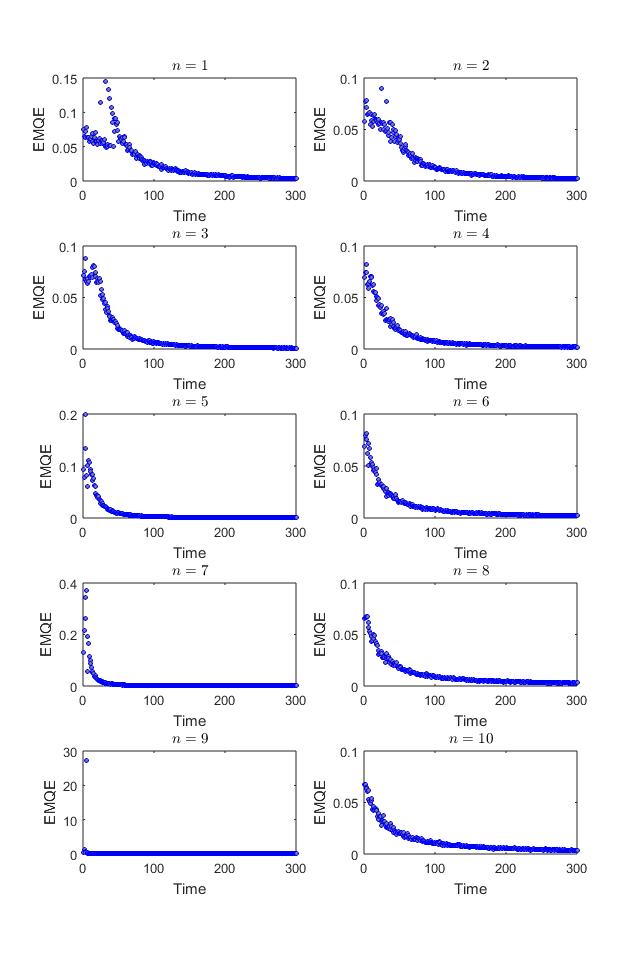}
\includegraphics[width=4.5cm,height=6cm]{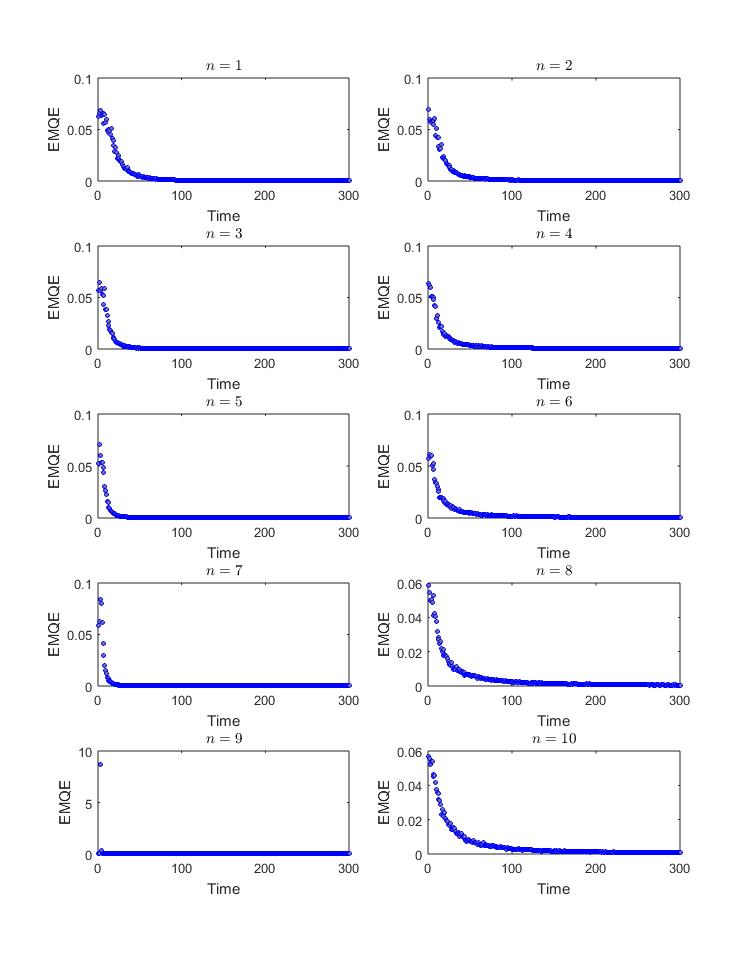}
\end{center}
\caption{\scriptsize{\emph{Subfamily 1}. The time-varying Empirical Mean Quadratic Errors (EMQEs) are displayed for the following values of the  functional sample size $T$, number of spherical nodes  $N$, truncation parameter  $TR$, prior hyperparameter sample size $M$, and
 number of repetitions $R$:
 $T=50,$   $N=500,$  $TR=\log(T)\simeq  4,$  $M=100,$ $R=200$
 (top-left); $T=300,$   $N=150,$  $TR=\log(T)\simeq 6,$  $M=50,$  $R=400$  (top-right);
 $T=300,$  $N=200,$  $TR= \log(T)\simeq 6,$   $M=50,$   $R=400$  (center-left);
 $T=500,$   $N=50,$  $TR= \log(T)\simeq 6,$  $M=100,$  $R=400$   (center-right);  $T=300,$   $N=200,$ $ TR=[T^{\varrho}]_{-},$ with $\varrho=1/2.45,$  $M=50,$   $R=400$     (bottom-left);    $T=300,$   $N=250,$  $TR=[T^{\varrho}]_{-},$ with  $\varrho=1/2.45,$    $M=50,$  $R=400$
   (bottom-right)}}
 \label{F5P}
\end{figure}

 In what follows we analyze the performance of the proposed EBFGP methodology under subfamily 1,  covering  the  combinations of the values  $T=50,300,500$,  \linebreak $N=50,150,200,250,500$,  $TR=\log(T),$ $TR=[T^{\varrho}]_{-}$, $\varrho=1/2.45$, $M=50,100$, and $R=200, 400$. Parameter $\varrho$ is fitted to  decay velocity of covariance operator eigenvalues following the rules stated in Theorems 8.7--8.8 in  \cite{Bosq00} about consistency of projection correlation operator estimators.
    The time-varying Empirical Mean Quadratic Errors (EMQEs), at each spherical  Laplace--Beltrami operator eigenspace,  are computed by applying a truncated version of Parseval identity. They are displayed in
  Figure \ref{F5P},  showing  the most significant combinations of these tuning parameter values, regarding observed differences  on the performance of the proposed methodology, with the aim of illustrating their interaction and effect. Specifically,
comparing the top-left and top-right plot groups in this figure, we can observe the effect of increasing parameters $N$ and $M$.
These parameters are  respectively  involved
in the discrete spatial  numerical projections and the Monte Carlo approximation in the numerical integration performed,  in the  time-adaptive Empirical Bayes computation of the hyperparameter ML-II estimates.  In both plot groups, a logarithmic truncation rule is applied ensuring a similar performance according to the functional sample size, which is larger at the top-right. On the other hand, when large functional sample sizes are considered ($T=500$),  under a sparse spatial scenario ($N=50$), the logarithmic truncation scheme is perfectly suited for fitting the  local spatial variation, allowing a substantial improvement when increasing the size of the hyperparameter prior samples from $M=50$ to $M=100$.  This improvement is observed at the center-right, where in both plot groups the same number of replicates $R=400$, of the FGP and functional observations, are generated  for the Monte Carlo numerical integration involved in the time-adaptive  Empirical Bayes implementation. Note that  the center-left  corresponds to a smaller value $T=300,$  with a larger number $N=200$ of spherical nodes, requiring  a higher value of the truncation parameter  than the one here applied, provided by the logarithm of the functional sample size.  Finally, we illustrate that, as expected, the error induced by spatial discretization  involved in the computation of spatial projections is less significant in the implementation of the FGP regression approach than the effect of the magnitudes $M$ and $TR$, with the truncation order $TR$  depending on  $T$ and $N$.   This fact is illustrated in the two plot groups at the bottom of Figure \ref{F5P}, where a similar performance is observed for equal  parameter values
  $T, TR, M, R$, and  an increment of $50$ spherical nodes at the bottom-right  with respect to bottom-left. This increment in the number of spherical nodes is reflected on a slightly better performance at higher resolution levels at the bottom-right, including a  smaller time edge effect. Thus, at coarser scales where  large scale properties are reproduced we obtain almost the same performance.

\begin{figure}[h]
\begin{center}
\includegraphics[width=4.5cm,height=6cm]{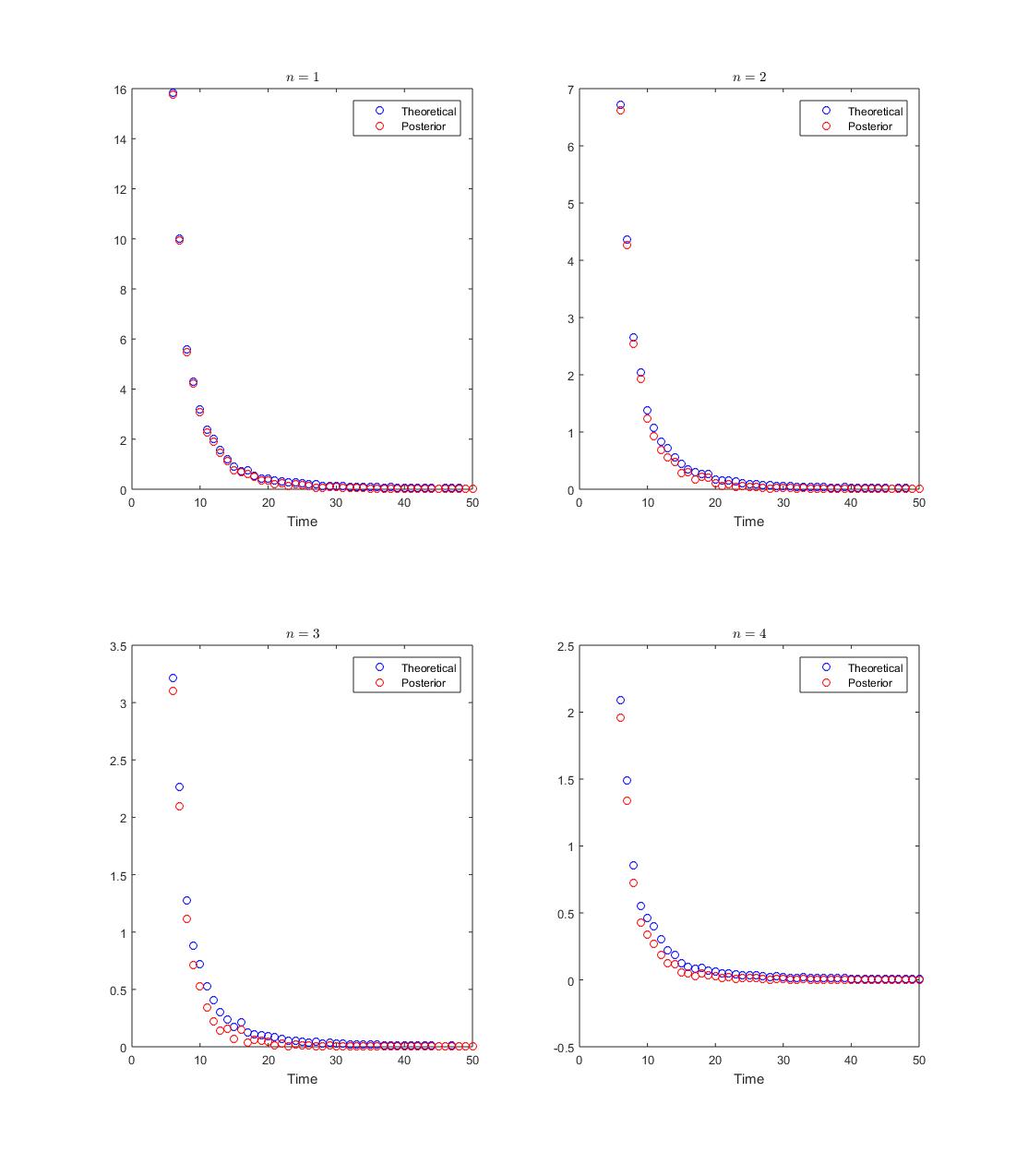}
\includegraphics[width=4.5cm,height=6cm]{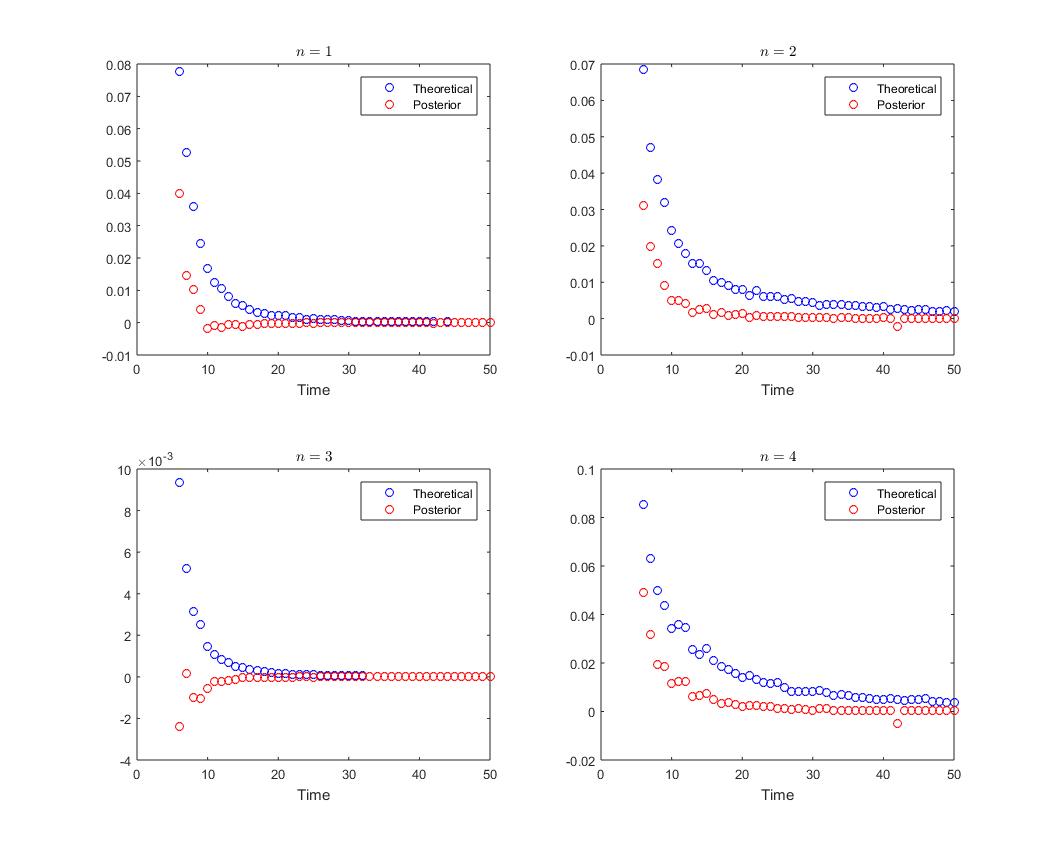}
\includegraphics[width=4.5cm,height=6cm]{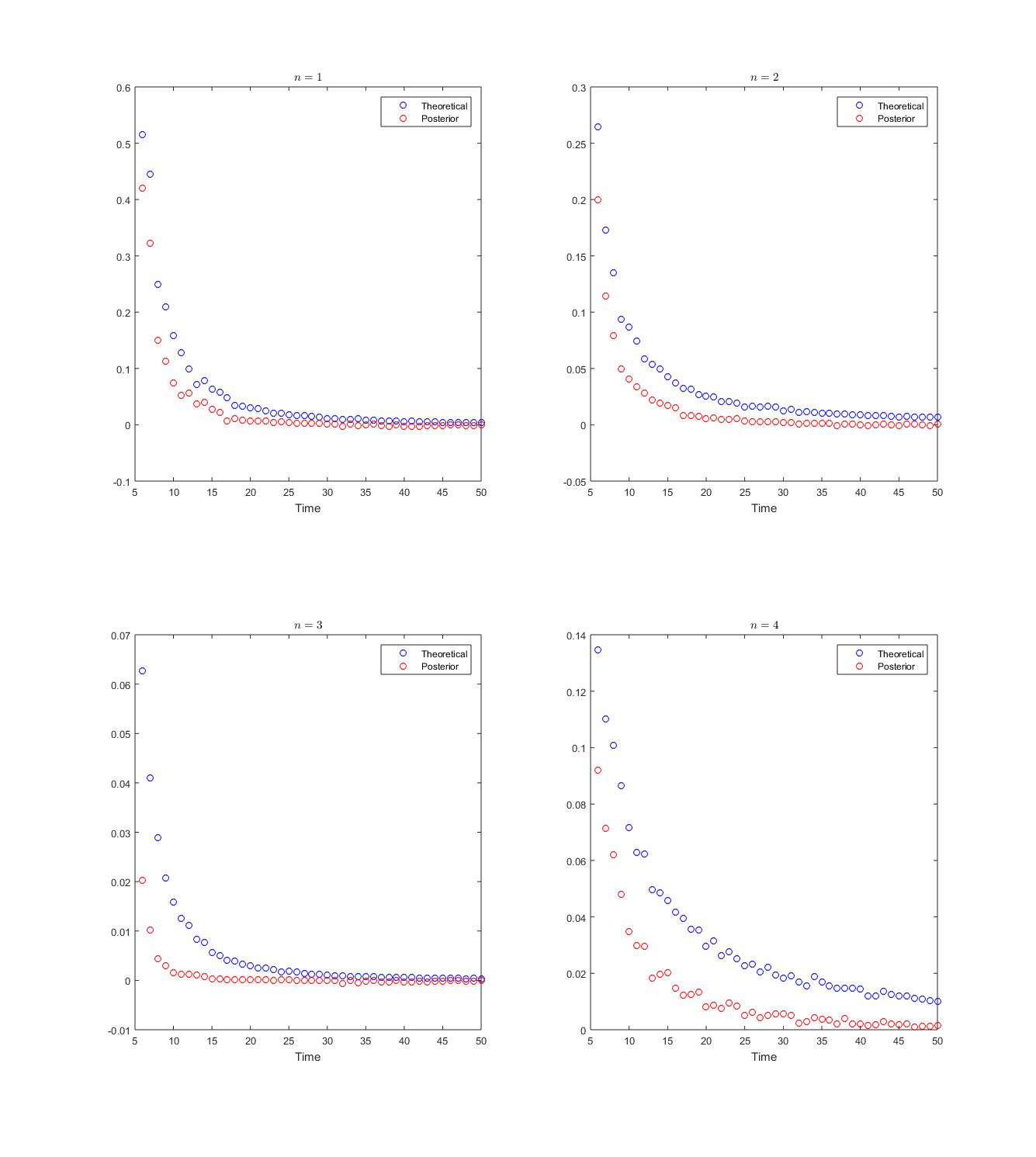}
\includegraphics[width=4.5cm,height=6cm]{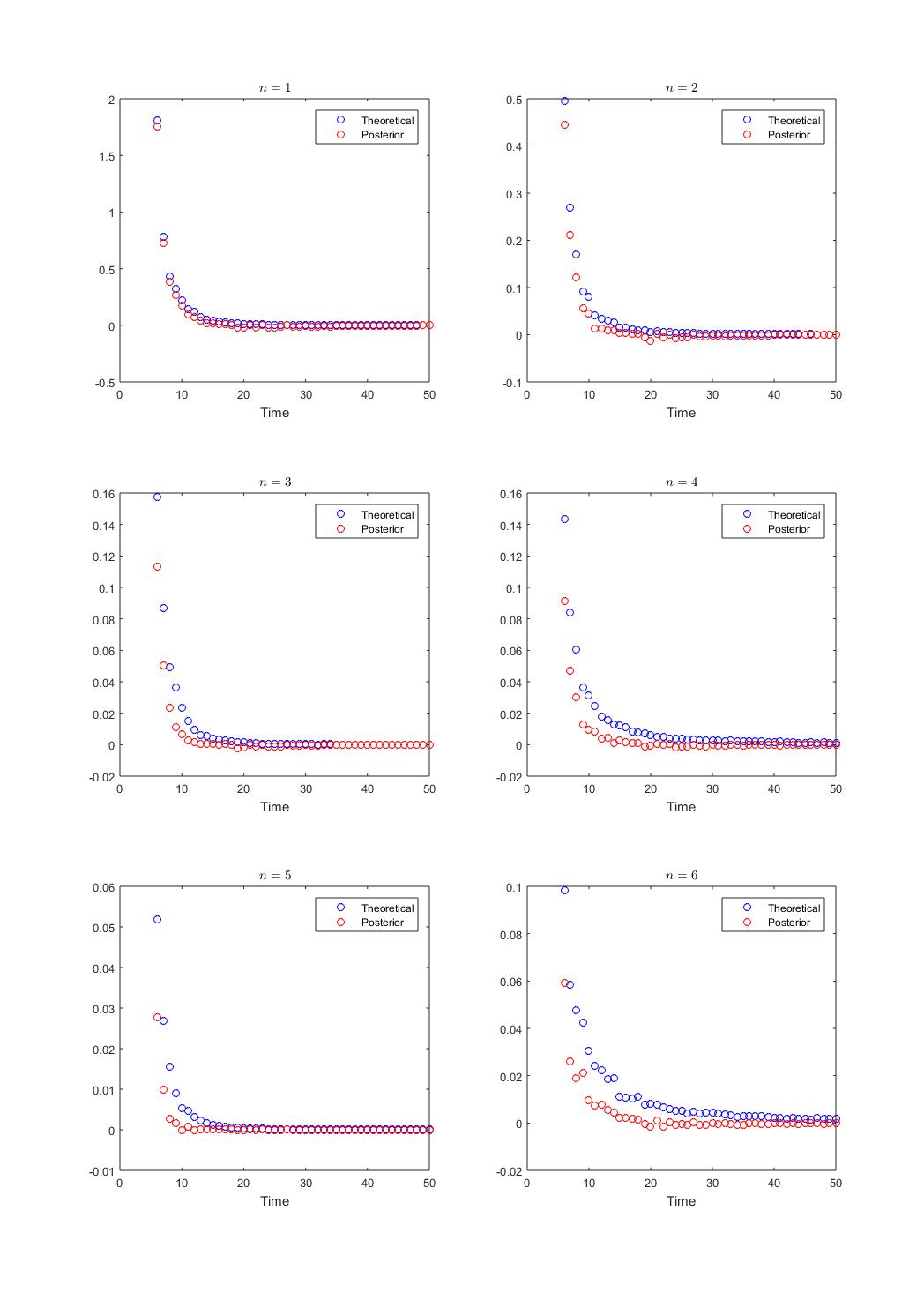}
\includegraphics[width=4.5cm,height=6cm]{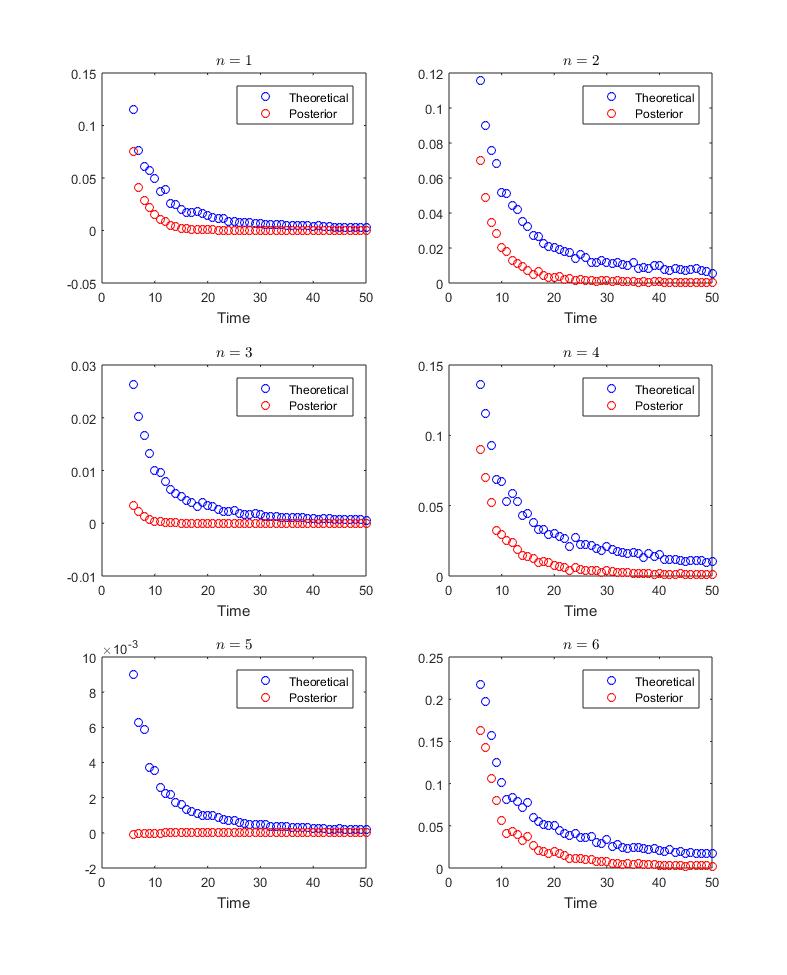}
\includegraphics[width=4.5cm,height=6cm]{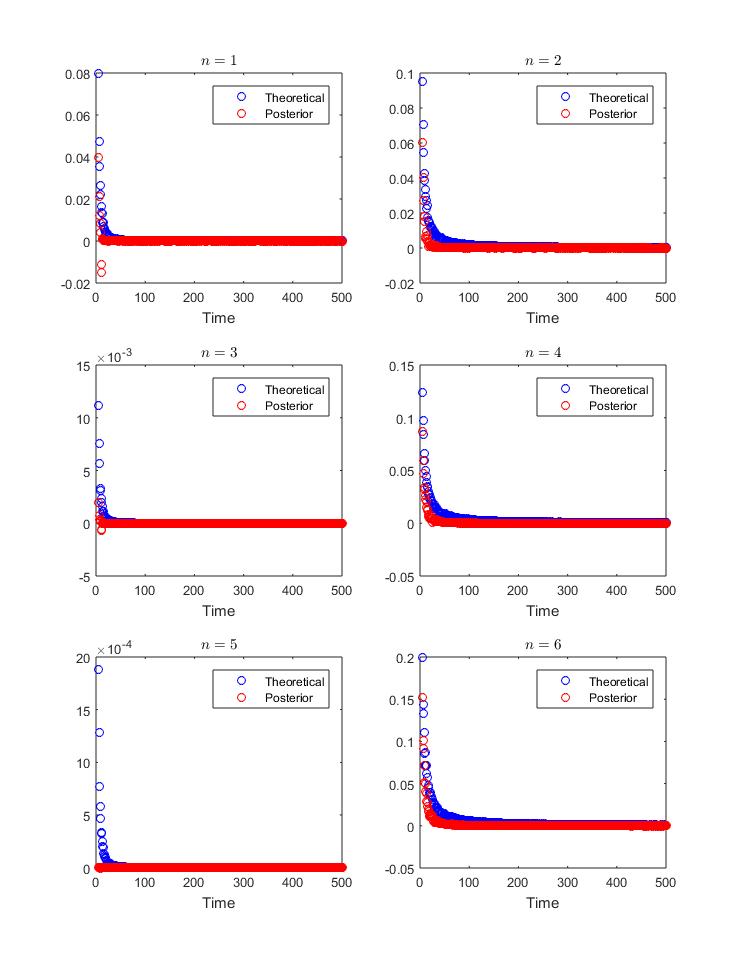}
\end{center}
\caption{\scriptsize{\emph{Subfamily 1}. Theoretical (blue dotted line) and posterior (red dotted line) time linear correlation of the projected FGP model, at the eigenspaces specified by the logarithmic and power-function truncation schemes analyzed.  The following values of the  functional sample size $T$, number of spherical nodes  $N$, truncation parameter  $TR$, prior hyperparameter sample size $M$, and
 number of replicates $R$ are displayed:
$T=50, N=500, TR=\log(T)\simeq 4, M=100, R=200$ (top-left); $T=50, N=50, TR=\log(T)\simeq 4, M=50, R=400$  (top-right);
$T=50, N=50,   TR=\log(T)\simeq 4, M=50,  R=500$  (center-left); $T=50, N=500,  TR=6, M=100, R=200$  (center-right);
$T=50, N=50, TR=6,  M=50,  R=400$ (bottom-left); $T=500, N=50,  TR=6,  M=100,   R=400$  (bottom-right)}}
 \label{F6P}
\end{figure}

The discretized  version of  temporal linear correlation  $\mathcal{C}_{\mathcal{Z}_{n,j}}\left(t,s, \boldsymbol{\theta }\right)$   in (\ref{linearcorr}), based on the theoretical  $\boldsymbol{\theta } $ value, of $Z_{n,j}(t,\boldsymbol{\theta })$ in (\ref{ptvas}), is displayed in blue dotted line in Figure \ref{F6P}. The corresponding posterior approximation, computed via equation (\ref{eqccong2}) (see also equation  (\ref{eqfvdtvs})),  is displayed in  red dotted line in this figure.  Again, increasing $N$ and $M$, a better performance of  EBFGP is observed at the top-left of Figure \ref{F6P}, with respect the top-right. Note that, beyond the minimum threshold $R=200$, Monte Carlo numerical integration produces suitable approximations in the implementation of time-adaptive Empirical Bayes. This fact can also be checked at the center-right, where a better performance is obtained than at the center-left. The improvement obtained by increasing $T$ and $M$ can also be observed at the bottom-right, with respect to the bottom-left.

\begin{figure}[h]
\begin{center}
\includegraphics[width=5cm,height=6.5cm]{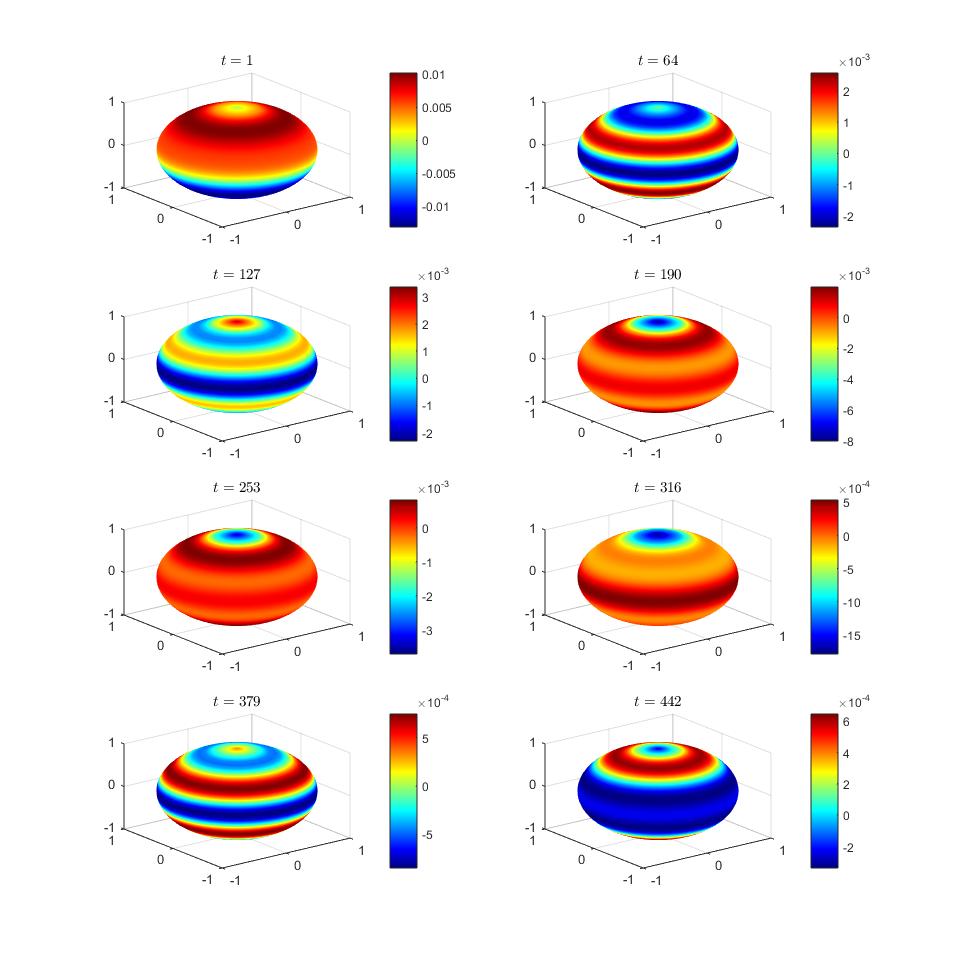}
\includegraphics[width=5cm,height=6.5cm]{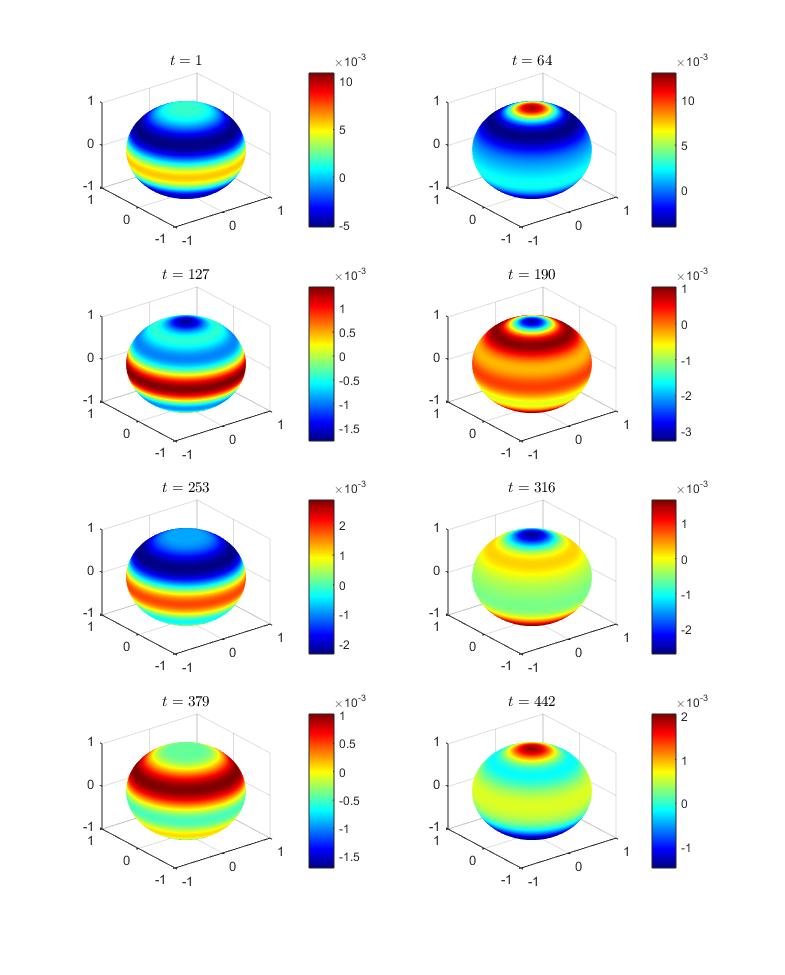}
\hspace*{0.2cm}
\includegraphics[width=4cm,height=6.5cm]{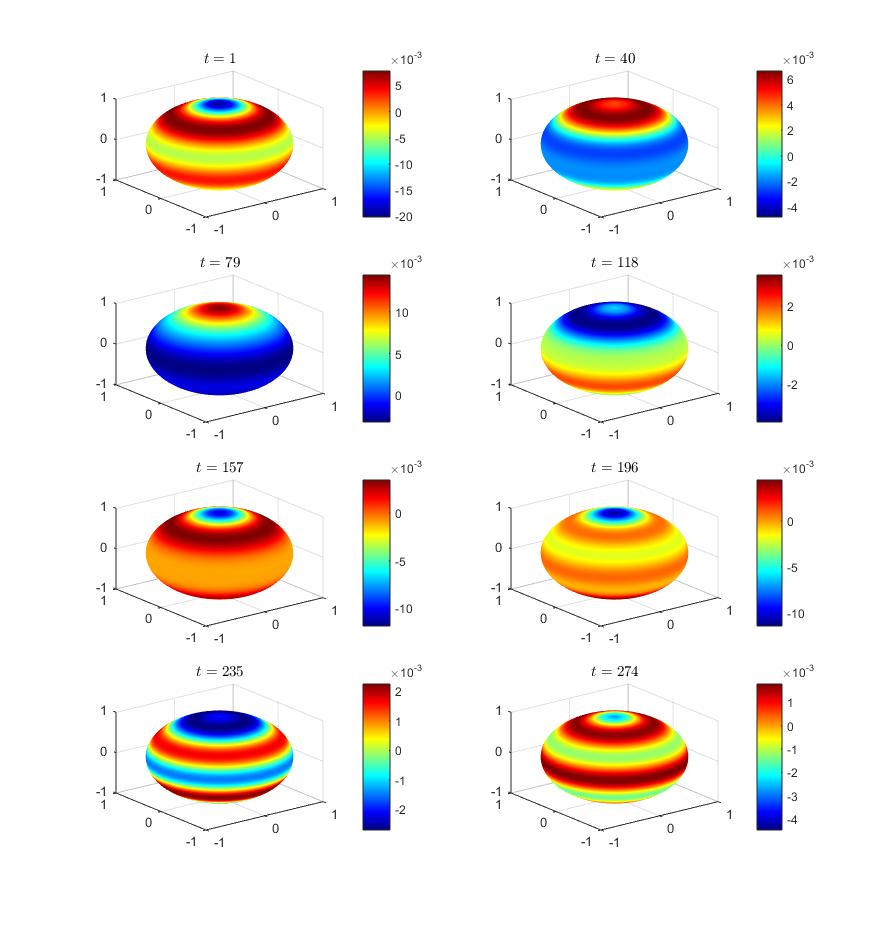}
\includegraphics[width=4cm,height=6.5cm]{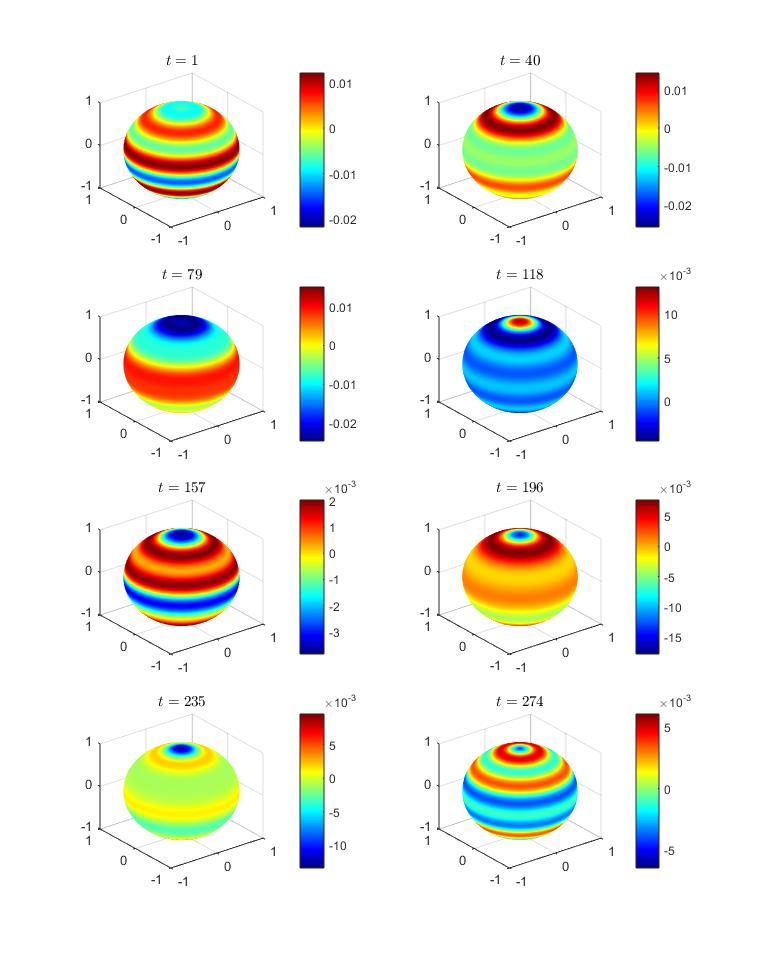}
\includegraphics[width=4cm,height=6.5cm]{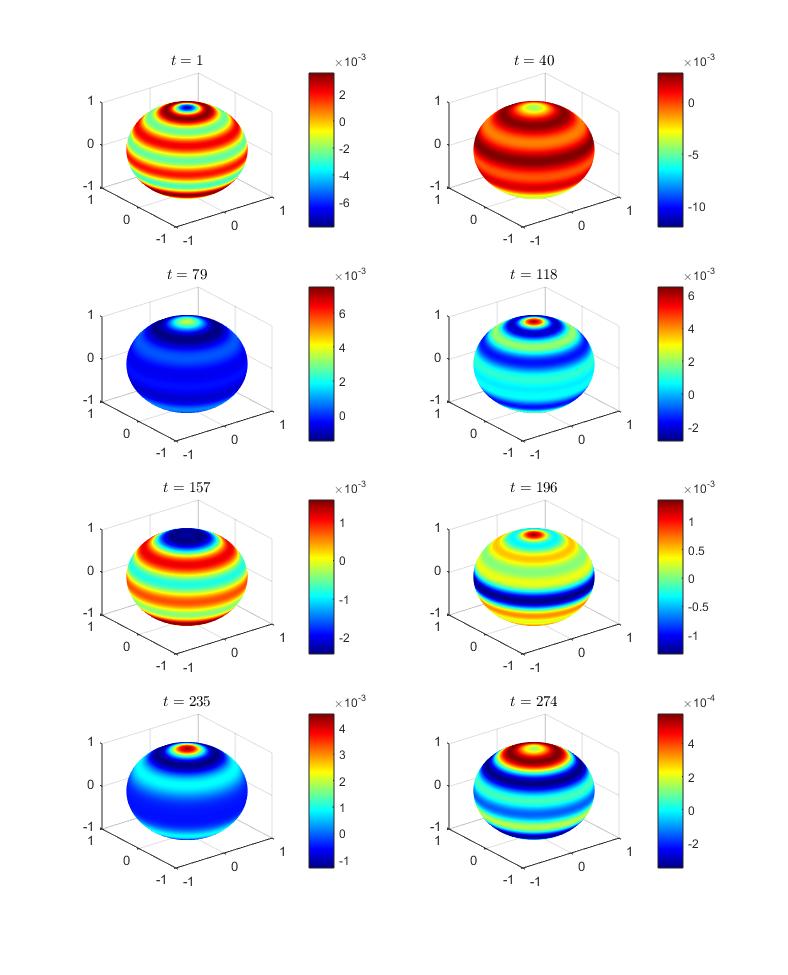}
\end{center}
\caption{\scriptsize{\emph{Subfamily 1}. Empirical mean of the bias, $\mathbb{D} = \left\{1, 64, 127, 190, 253, 316, 379, 442\right\}\subset  \mathbb{T}$ (top), $\mathbb{D} = \left\{1, 40, 79, 118, 157, 196, 235, 274\right\} \subset  \mathbb{T}$} (bottom). The following values of the  functional sample size $T$, number of spherical nodes  $N$, truncation parameter  $TR$, prior hyperparameter sample size $M$, and
 number of replicates $R$ are displayed:
$T=500, N=50, TR=\log(T)\simeq 6, M=100, R=400$ (top-left); $T=500, N=150, TR=\log(T)\simeq 6, M=50, R=200$  (top-right); $T=300, N=150, TR=\log(T)\simeq 6, M=50, R=400$
(bottom-left);   $T=300, N=200,   TR=[T^{\varrho}]_{-}= 10, \varrho=1/2.45, M=50, R=400$ (bottom-center);
 $T=300, N=250, TR=[T^{\varrho}]_{-}= 10, \varrho=1/2.45, M=50, R=400$ (bottom-right)}
 \label{F7P}
\end{figure}

Bias analysis in $L^{2}(\mathbb{S}_{2},d\nu)$-norm is also performed  (see  equation (\ref{decbv})).   The empirical mean of the bias over the $R$ replicates is shown in Figure \ref{F7P},  for different sets $\mathbb{D}$,  under  several  scenarios specified by the parameter values:
$T=500, N=50,    TR=\log(T)\simeq 6, M=100, R=400$, at the top-left; $T=500, N=150, TR=\log(T)\simeq 6, M=50, R=200$,  at the top-right; $T=300, N=150,  TR=\log(T)\simeq 6,  M=50, R=400$,
at the bottom-left;  $T=300, N=200, TR=[T^{\varrho}]_{-}= 10, M=50, R=400$,  at the bottom-center, and
 $T=300, N=250, TR=[T^{\varrho}]_{-}= 10, M=50, R=400$,  at the bottom-right.
 A  similar behavior is displayed by  the bias empirical mean regarding the increase of the parameter $M$ (see  top plot groups of   Figure \ref{F7P}), as well as when $N$ increases  (see bottom plots of  Figure \ref{F7P}), under both,  logarithmic (top plots and bottom-left plot), and power-function truncation schemes (bottom-center and bottom-right). Smoothing induced by the empirical mean, based on  $R$ replicates,  hinders the magnitude of changes when  increasing the values of $T,N, TR$ and $M$. As before, the minimum threshold $R=200$ ensures a suitable approximation via Monte Carlo numerical integration, hence leading to small bias empirical mean  values.

\begin{figure}[h]
\begin{center}
\includegraphics[width=5cm,height=6.5cm]{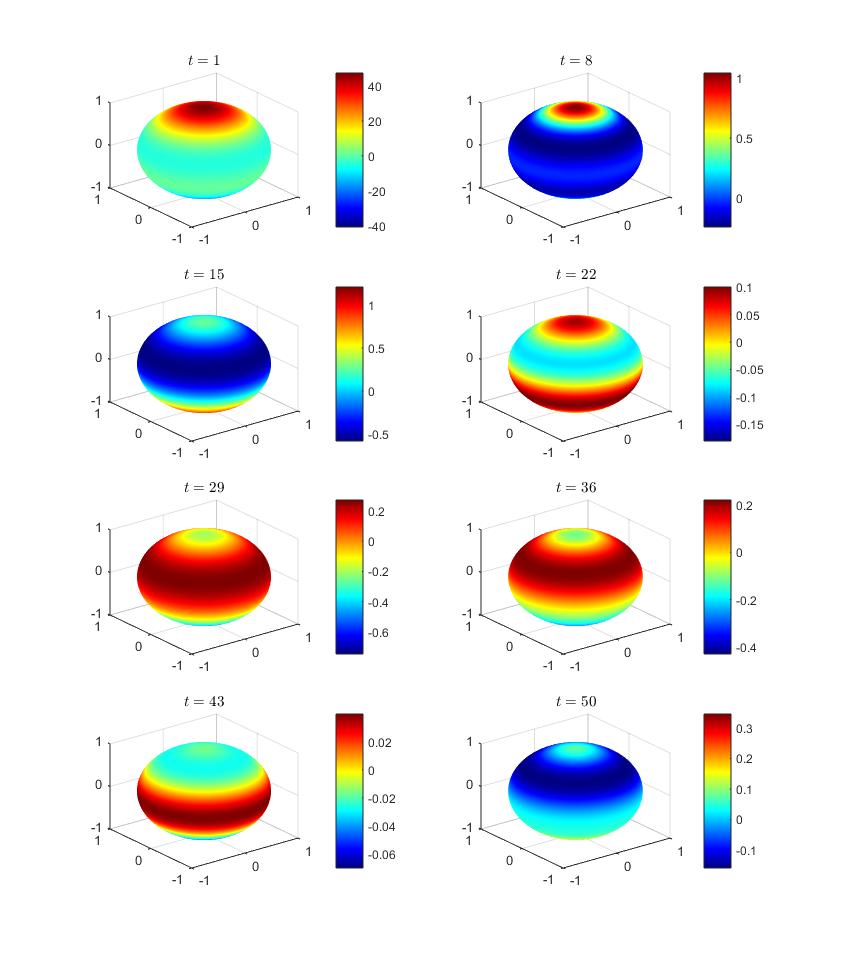}
\includegraphics[width=5cm,height=6.5cm]{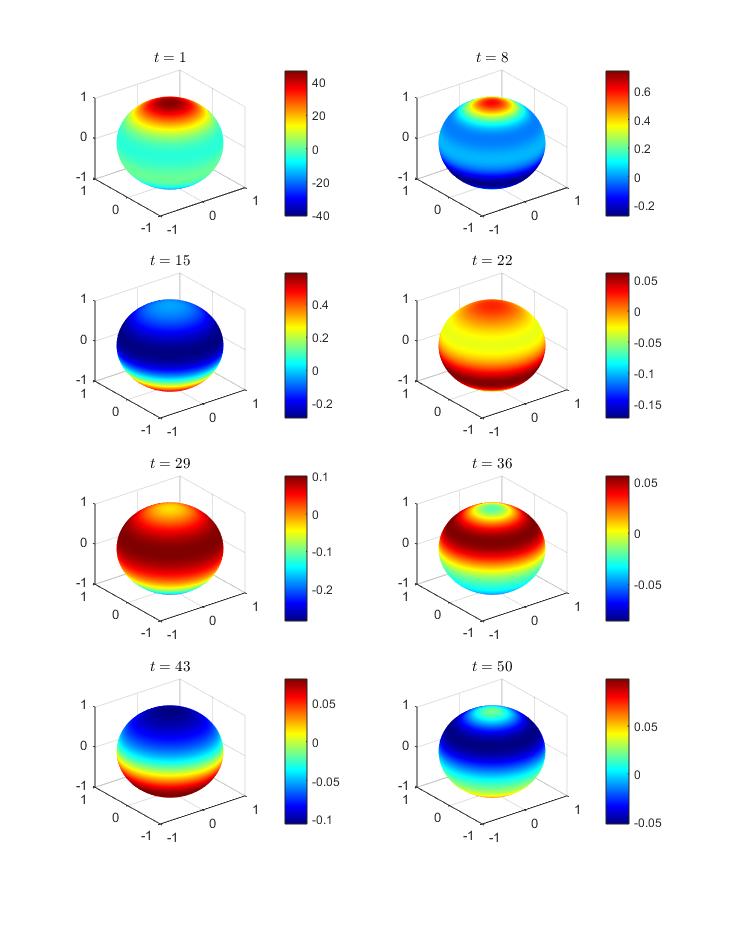}
\end{center}
\caption{\scriptsize{\emph{Subfamily 1}.  Original FGP sample value (left), and posterior
mean approximation (right), $\mathbb{D} = \left\{1, 8, 15, 22, 29, 36, 43, 50\right\} \subset  \mathbb{T},$  for $T=150, N=500,  TR=4, M=100,R=200$}}
 \label{F8P}
\end{figure}

Summarizing,  a good performance is observed under a logarithmic truncation scheme, when low spatial sampling frequency values are considered. Indeed, increasing the functional sample size, under this sparse spatial scenario, leads to an important asymptotic improvement in terms of EMQEs, the posterior  approximation of the linear  correlation in time of the FGP model, and the empirical mean of the time-varying spherical functional bias.  As expected, large prior hyperparameter sample sizes also improve the performance of the posterior predictor. Increasing the spatial sampling frequency requires higher truncation parameter values (fractional power law), particularly at high resolution levels, although the logarithmic truncation scheme still provides a good performance.  Regarding posterior time linear correlation reconstruction of the FGP, a better performance is obtained when the logarithmic truncation scheme is applied, ensuring consistency (see Chapter 8 in \cite{Bosq00}). We  work in the time-varying purely point spectral domain, leading to an important dimension reduction.  The   reconstruction of the posterior mean involves the application of  the inverse transform in terms of the   eigenfunctions  of the Laplace--Beltrami operator, requiring higher spatial sampling frequency to minimize  the numerical errors associated with the spatial discretization of the elements of this basis. Figure \ref{F8P} displays  FGP at the left, and posterior mean at the right, for the minimum threshold  $N=500$ for such a reconstruction, where  EBFGP has been implemented, for    $T=150, TR=4, M=100, R=200$.

\subsection{Subfamily 2}
\begin{figure}[h]
\begin{center}
\includegraphics[width=4cm,height=6.5cm]{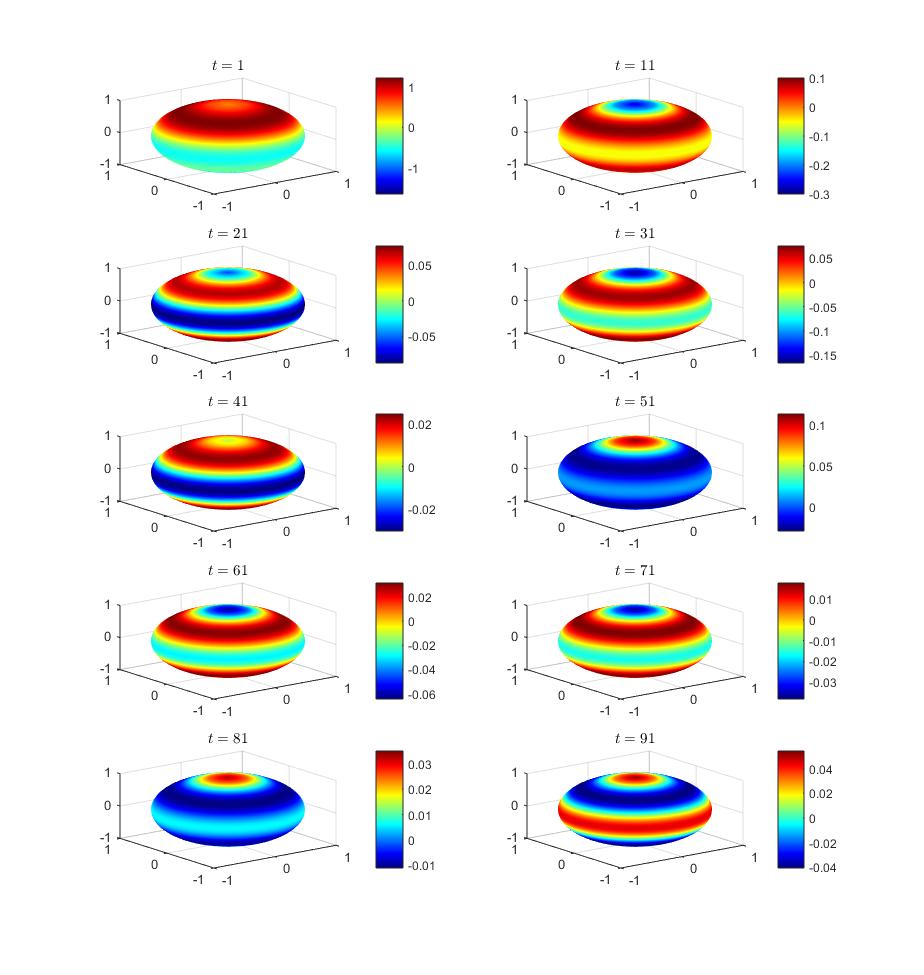}
\includegraphics[width=4cm,height=6.5cm]{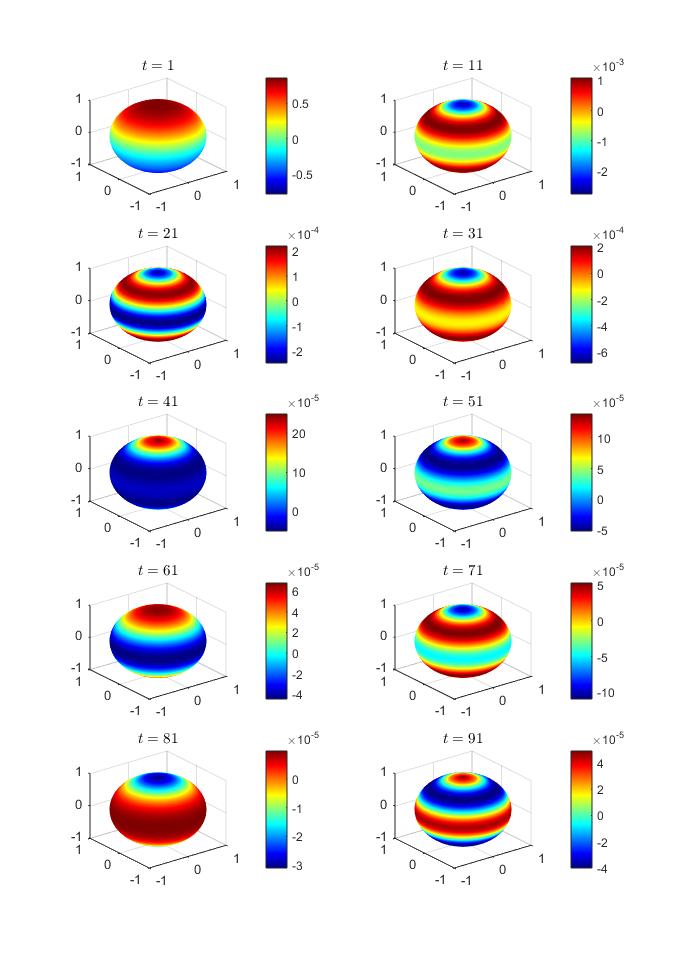}
\includegraphics[width=4cm,height=6.5cm]{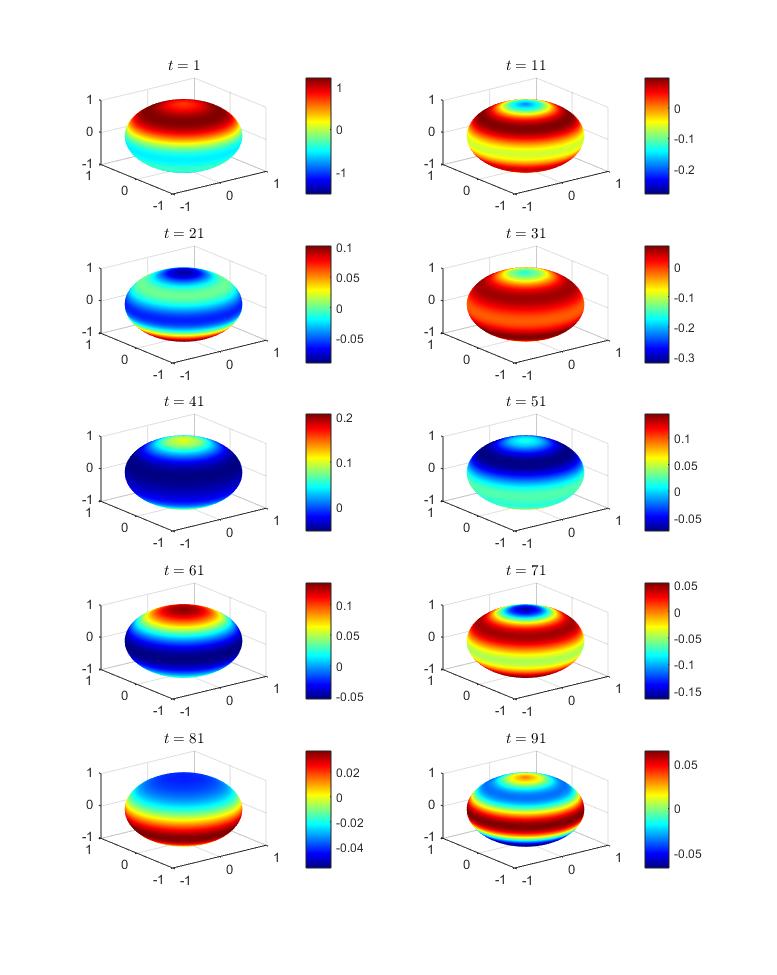}
\end{center}
\caption{\scriptsize{\emph{Subfamily 2}. Realization of conditional  FGP spherical functional time series model  (left), posterior spherical functional mean (center), and  observation spherical functional time series model (right),
$\mathbb{D}=\left\{1, 11, 21, 31, 41, 51, 61, 71, 81, 91\right\}\subset \mathbb{T}$ }}
 \label{F3P}
\end{figure}

One generation of the prior FGP in subfamily 2 is displayed   over \linebreak $\mathbb{D}=\left\{1, 11, 21, 31, 41, 51, 61, 71, 81, 91\right\}$  at the left  of  Figure  \ref{F3P}.  Conditionally to the ML-II estimates of the hyperparameter vector $(\varpi ,\alpha ,\beta , \sigma(t) )$,   the posterior spherical functional mean over  $\mathbb{D}$ is also plotted at the center of this figure. The  corresponding functional observation affected by additive noise with variance $\sigma^{2}(t)$ is shown at the right.   In these computations,  the values  $T=300, N=150,  TR=[T^{\varrho}]_{-}, \varrho=1/2.744, M=50, R=300$  have been considered.  As before, parameter $\varrho$ is fitted to  the decay velocity of covariance operator eigenvalues following the rules stated in Theorems 8.7--8.8 in  \cite{Bosq00} about consistency of projection correlation operator estimators.

\begin{figure}[h]
\begin{center}
 \includegraphics[width=6cm,height=8cm]{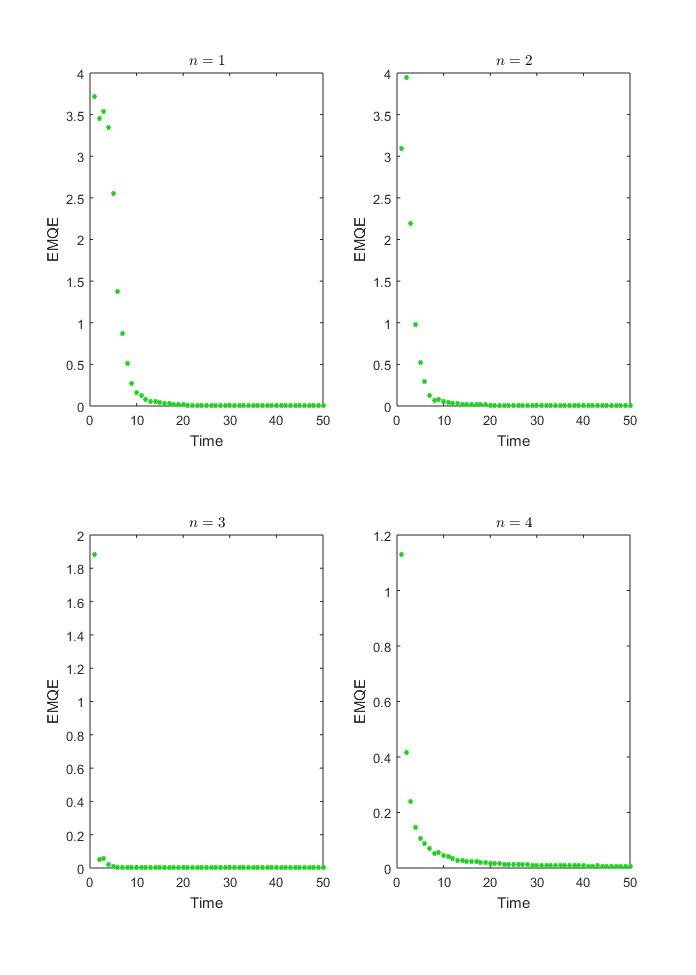}
\includegraphics[width=6cm,height=8cm]{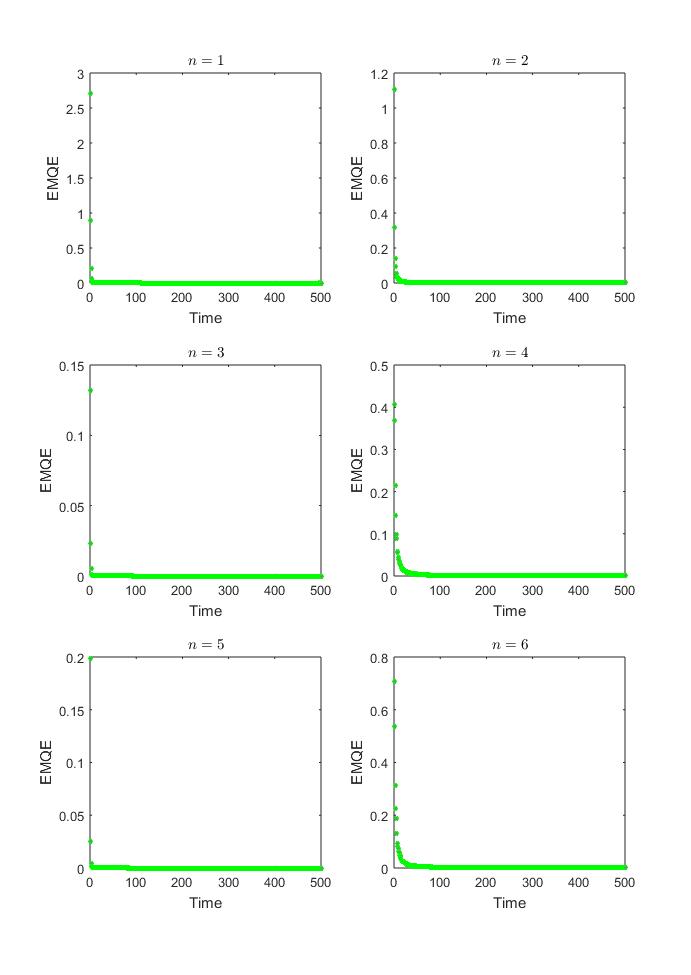}
\end{center}
\caption{\scriptsize{\emph{Subfamily 2}. Time-varying EMQEs.  $T=50,  N=500, TR=\log(T)\simeq 4,  M=100, R=400$
 (left); $T=500,  N=50, TR= \log(T)\simeq 6, M=100, R=400$  (right)}}
 \label{F5_A3B}
\end{figure}
\begin{figure}[h]
\begin{center}
\includegraphics[width=6cm,height=8cm]{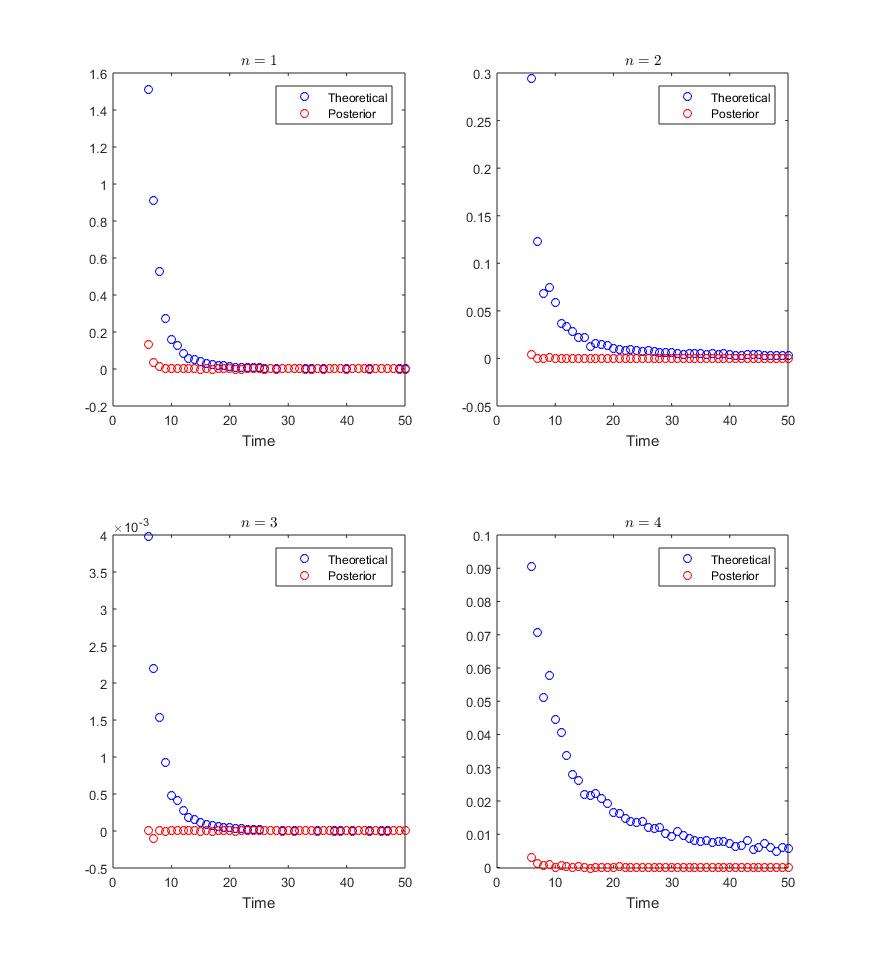}
\includegraphics[width=6cm,height=8cm]{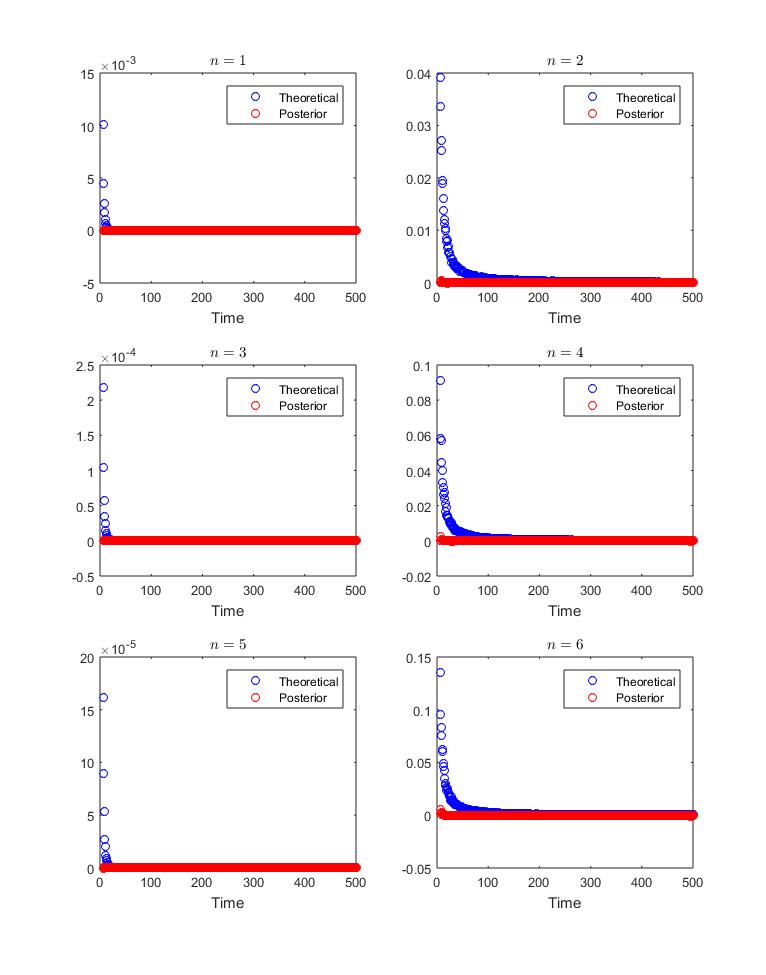}
\end{center}
\caption{\scriptsize{\emph{Subfamily 2}. FGP theoretical model (blue dotted line) and posterior time linear correlation reconstruction (red dotted line).  $T=50, N=500, TR=\log(T)\simeq 4, M=100,R=400$
 (left); $T=500,  N=50, TR= \log(T)\simeq 6, M=100, R=400$  (right)}}
 \label{F5_A3B2}
\end{figure}
\begin{figure}[h]
\begin{center}
\fbox{\includegraphics[width=4cm,height=6cm]{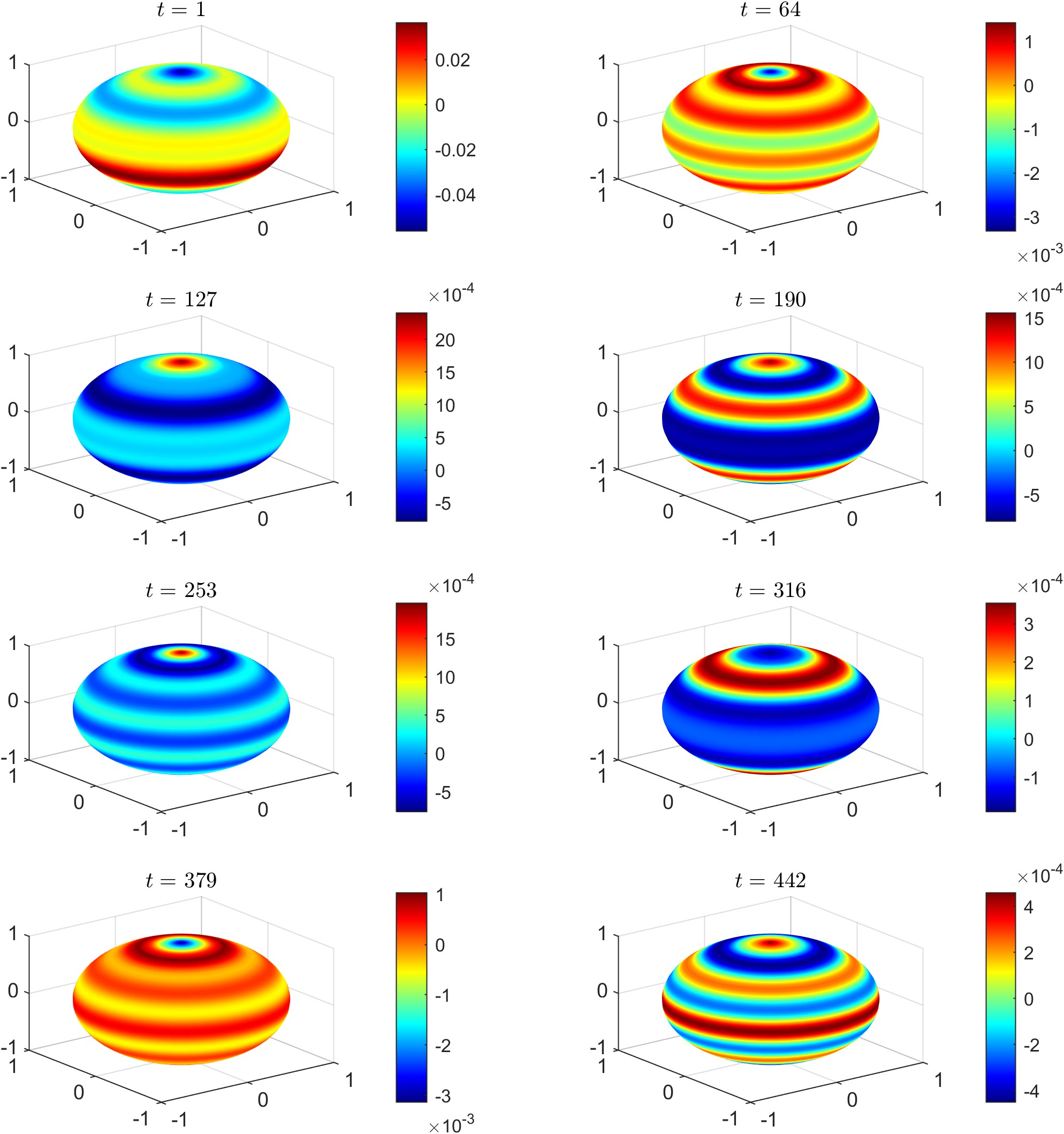}}
\fbox{\includegraphics[width=4cm,height=6cm]{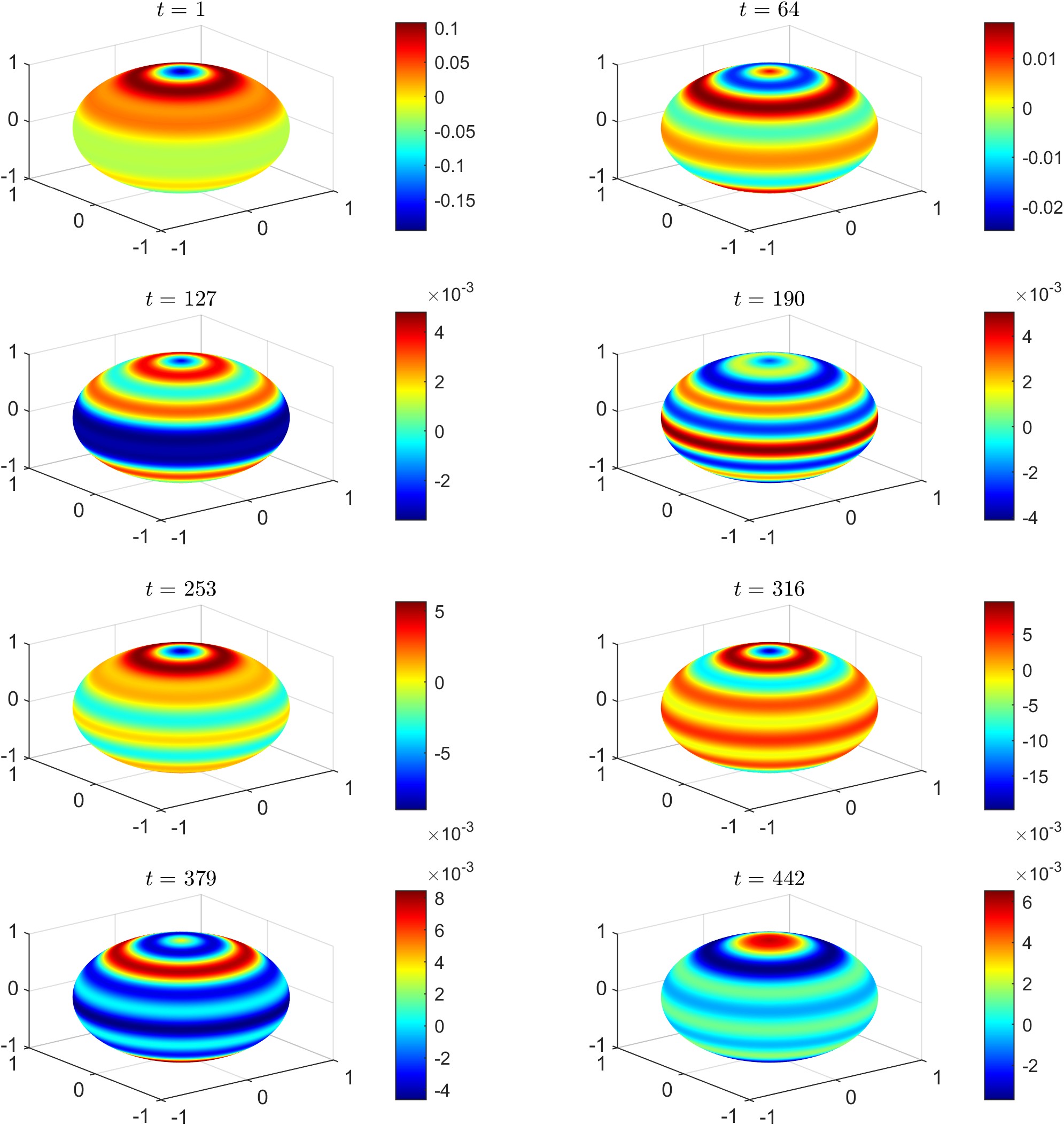}}\\
\fbox{\includegraphics[width=4cm,height=6cm]{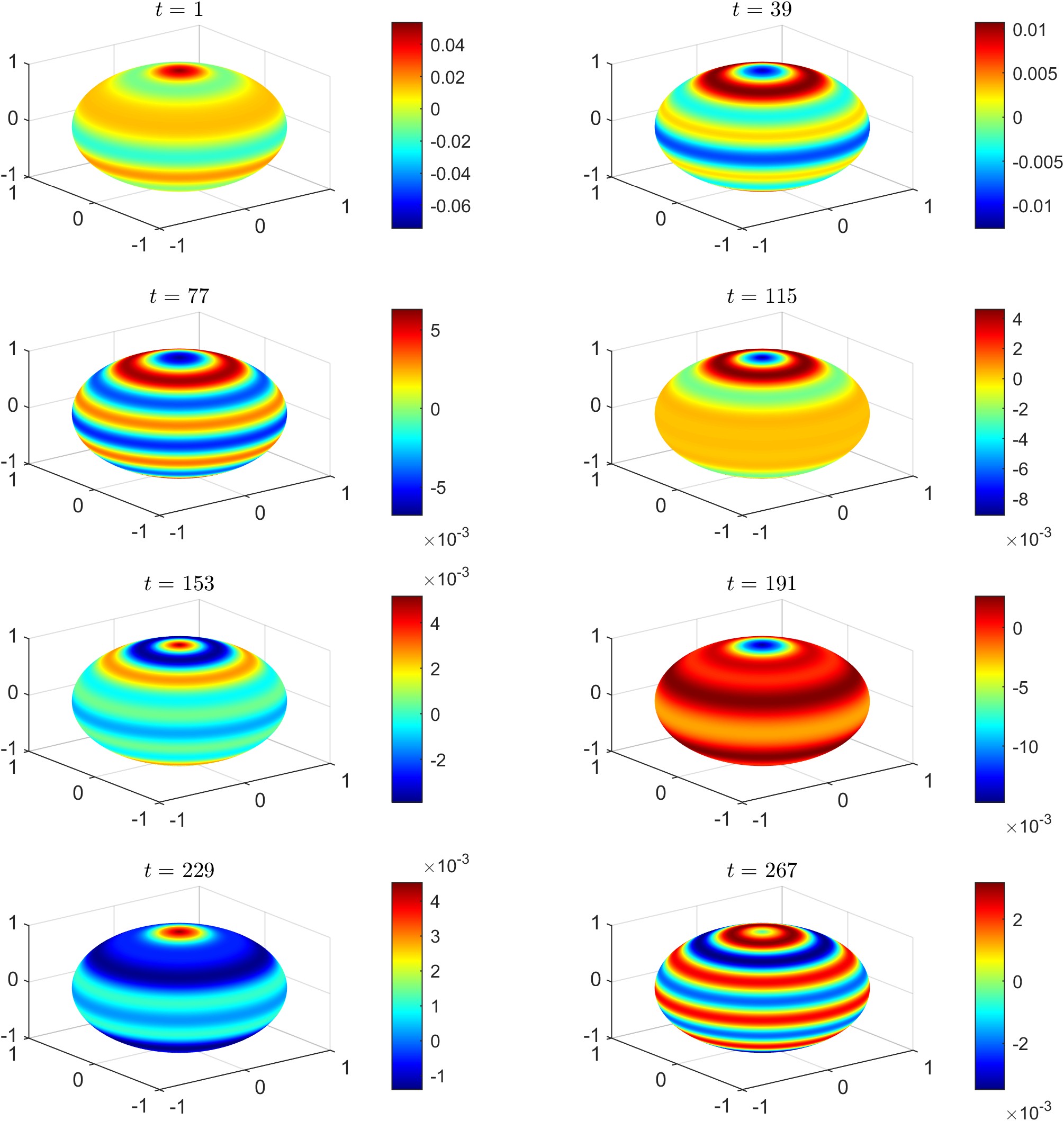}}
\fbox{\includegraphics[width=4cm,height=6cm]{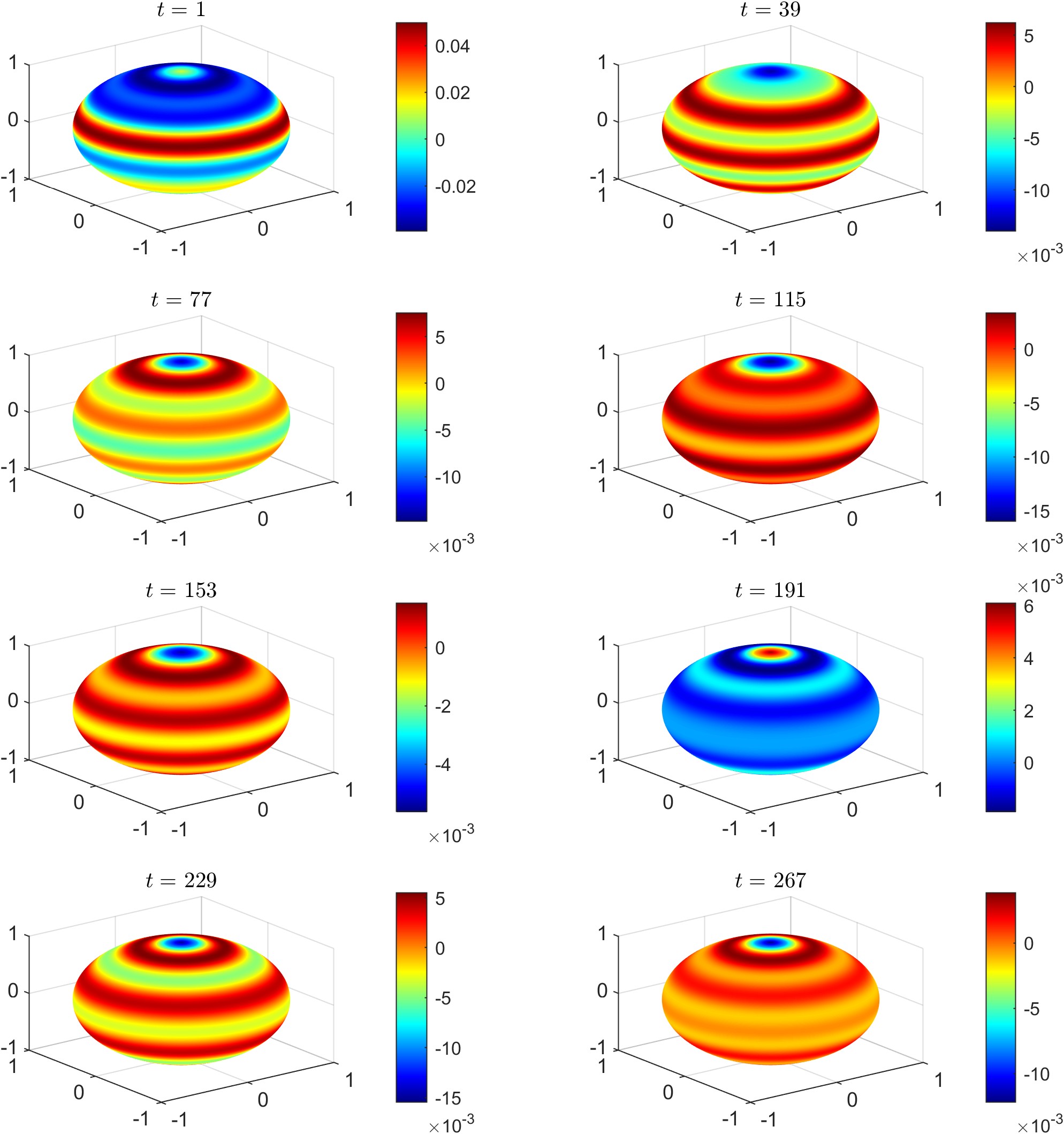}}\\
\fbox{\includegraphics[width=4cm,height=6cm]{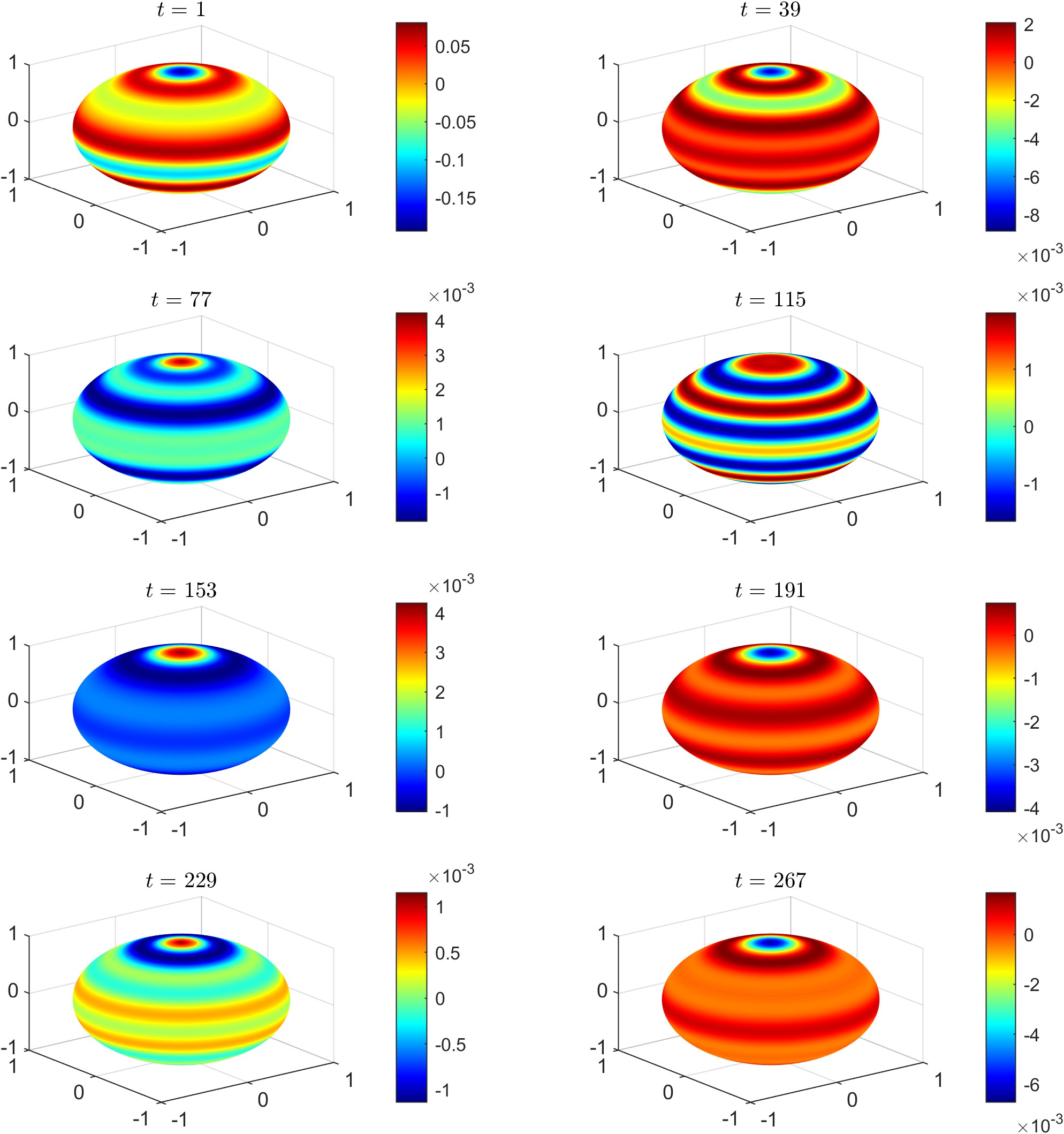}}
\fbox{\includegraphics[width=4cm,height=6cm]{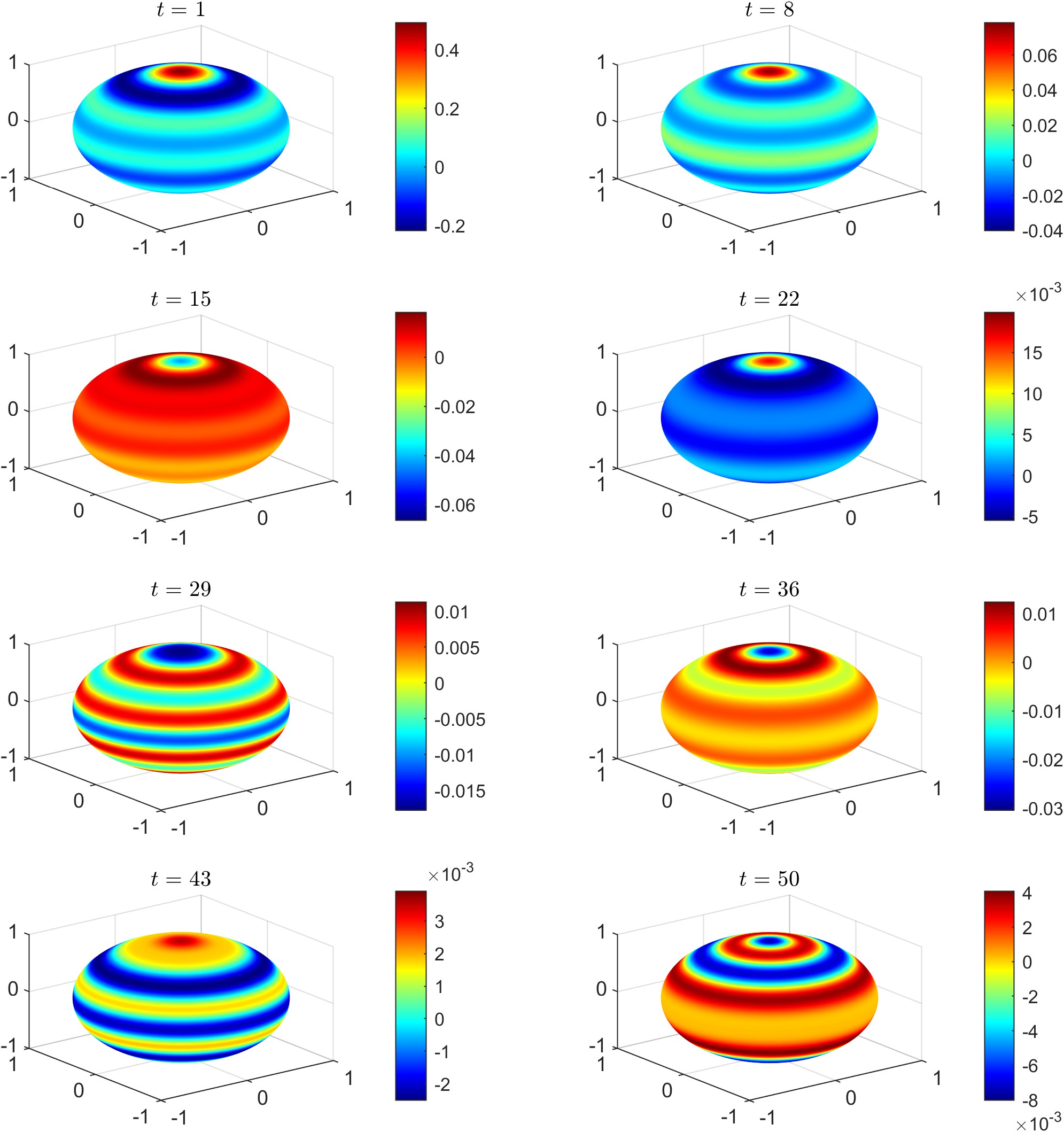}}
\end{center}
\caption{\scriptsize{\emph{Subfamily 2}.
Empirical mean of the bias over the $R$ replicates. $\mathbb{D} = \left\{1, 39, 77, 115, 153, 191, 229, 267\right\}\subset \mathbb{T},$ and
$T=500, N=50, TR=10, M=100, R=400$ (top-left); $T=500, N=150, TR=10, M=50, R=200$  (top-right); $T=300, N=150, TR=10, M=50, R=400$
(center-left);   $T=300, N=200, TR=10, M=50, R=400$
(center-right);
 $T=300, N=250,  TR=10,  M=50, R=400$
(bottom-left); $T=50, N=500, TR=10, M=100, R=200$
(bottom-right)}}
\label{F7_A3}
\end{figure}

The time-varying  EMQEs display a similar behavior regarding  interaction between  $T,N,TR,M, R$ parameter values. A stronger time edge effect
  is observed under all scenarios analyzed. Due to  the local spatial smoothness displayed by subfamily (\ref{A2}),   when   the number of spherical nodes $N$ is increased,  and  the functional sample size $T$ decreases,  for fixed common values of parameters $M$ and $R$,  multicollinearity and  information redundancy  is more pronounced, leading to spatiotemporal confounding or over-parameterization
 (see Figure \ref{F5_A3B}) (decreasing T means that a few temporal nodes refresh sample  information under this short-range dependence scenario).
This effect is also  illustrated  in terms of the  FGP posterior time linear correlation reconstruction, reversing the situation  when $T$ increases and $N$ decreases (see   Figure \ref{F5_A3B2}).  Note that the increasing of $TR$   in subfamily (\ref{A2}) produces an opposite effect to the one observed in subfamily (\ref{A1}), since  coarser spherical scales  perfectly fit this smooth local variation,  and high resolution levels are not required. Hence, under  subfamily (\ref{A2}), the best choice corresponds to the logarithmic truncation scheme, as shown in the numerical results displayed. We only use a power-function truncation rule in Figure \ref{F3P} where one realization of the FGP, the posterior mean and the functional observation process is plotted to visualize the effect of higher resolution levels in the displayed sample paths.
  In the bias analysis,  similar magnitudes to the ones displayed   for subfamily (\ref{A1}) are obtained. The largest  bias empirical mean  values are observed at high resolution levels rather than at coarser scales. As before, increasing parameter $N$ also increases bias empirical mean  values  (see Figure  \ref{F7_A3}).

\clearpage

\section{Synthetic data analysis}
\label{sda}
\begin{figure}[h]
\begin{center}
\includegraphics[width=5cm,height=3.5cm]{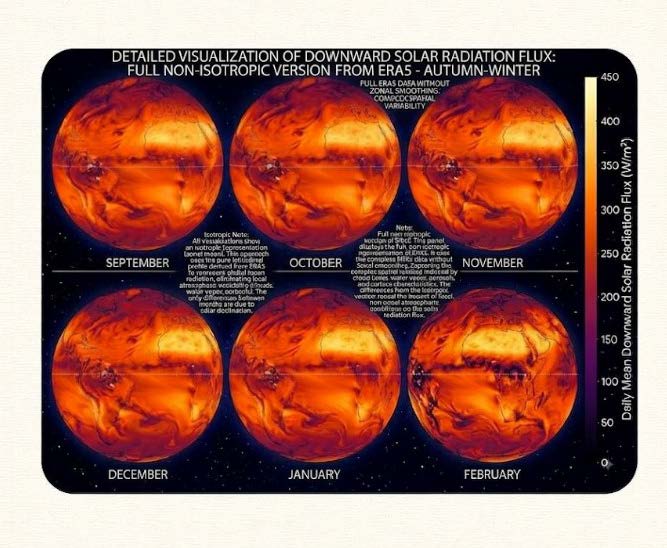}\hspace*{0.7cm}
\includegraphics[width=5cm,height=3.5cm]{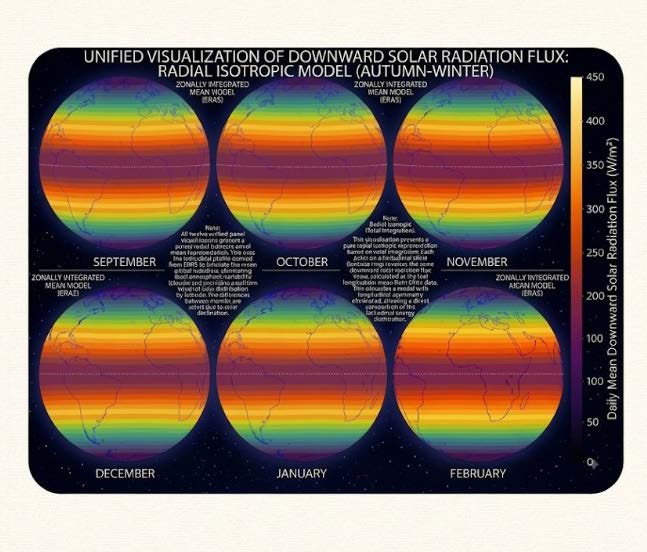}
\end{center}
\caption{\scriptsize{Original values (left), and axial (zonal) isotropic approximation (right) of hourly dataset of ERA5 downward solar radiation flux   (one  day at each month during the period September-February, corresponding to  autumn-winter)}}
 \label{f1app2}
\end{figure}

\begin{figure}[h]
\begin{center}
\includegraphics[width=5cm,height=6cm]{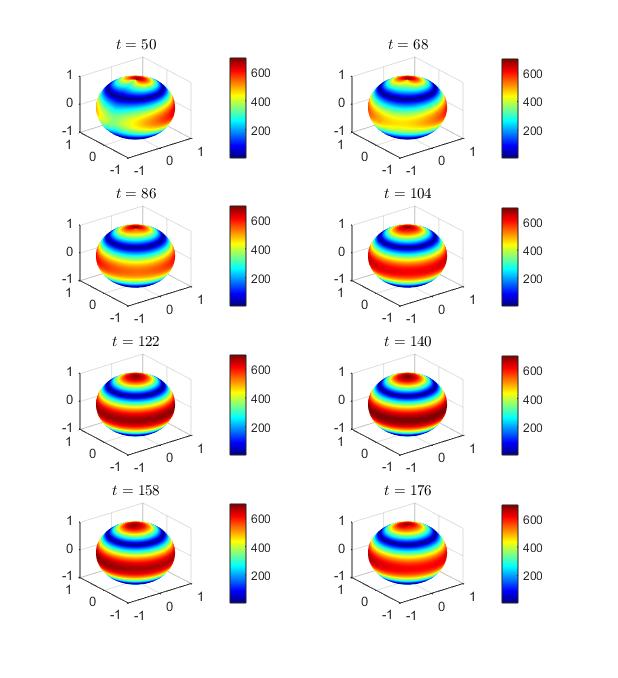}
\includegraphics[width=5cm,height=6cm]{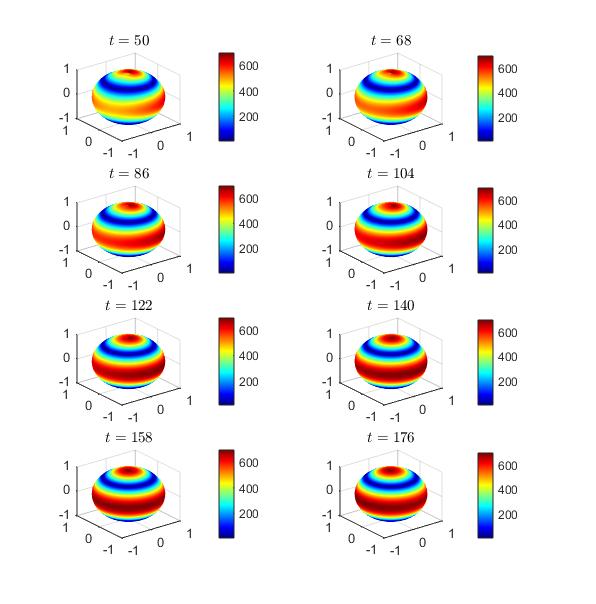}
\end{center}
\caption{\scriptsize{Downward solar radiation flux, $\mathbb{D} = \left\{50, 68, 86, 104, 122, 140, 158, 176\right\}\subset \mathbb{T}$. Unrestricted (left) and restricted (right) model to autumn-winter, involving different declination and  zenith  angles}}
 \label{f1app}
\end{figure}

In this section, we test the performance of the  time-adaptive   EBFGP regression  in the statistical analysis  of the nonlinear interaction between downward solar radiation flux  and  atmospheric pressure  at high cloud bottom. A   downward solar radiation flux   synthetic daily spherical functional dataset is generated,  based on a generative multi-physical  model (see equation (\ref{ephsi}) below). The computed values of downward solar radiation via this physical model share the solar geometry and atmospheric information, usually considered   in physical reanalyses like ERA5 (see Figure \ref{f1app2}).  Given that our primary goal is to analyze the nonlinear interaction between high cloud bottom pressure and  downward solar radiation, the  surface covariate information is ignored, being  considered as  irrelevant, to avoid some unnecessary noise in the exponential transfer function, modelling such a nonlinear interaction  (see equation (\ref{ephsi2}) below).  Uncertainty induced by these  spatiotemporal structural simplifications,  unobserved smoothly varying exogenous/endogenous covariates displaying weak/strong dependence
structures,  and measurement errors, are respectively represented, in  an  additive way, by an axial (or zonal)  isotropic temporal LRD spherical functional random effect,  and a  temporal  unstructured  spatiotemporal  noise. The axial  isotropy,  assumed in the generation of  the temporal LRD spherical functional random effect, is motivated by the spherical patterns displayed by downward solar radiation flux  during September-February
in autumn-winter  (see  Figures \ref{f1app2} and \ref{f1app}).

A $180\times 180$ meshgrid is built  in the  two-dimensional angle interval  $(0,\pi )\times (0,2\pi ),$ where  polar and azimuthal  angle nodes are respectively  located (see also \cite{Ovalle25}). The  zenith angle $A_{Z}$, involved in the physical generative model  (\ref{ephsi})  to compute  Solar Irradiance $I_{S}$,   depends on the time of the year, and on the  declination angle through a sinusoidal function also depending  on the day of the year. Specifically, Solar Irradiance $I_{S}$ is  computed from the following equation (see Figure \ref{f1app}):
\begin{equation}
I_{S}(t,\theta_{1}) =\frac{ 1361(0.8)}{\pi}\cos(A_{Z}(t, \theta_{1})),\quad t\in [0,183],\ \theta_{1}\in [0,\pi],
\label{ephsi}
\end{equation}
\noindent given in terms  of the Extraterrestrial Solar Constant  $G_0 = 1361$ W/$m^{2}$   (at the top  of the Earth's atmosphere), the
Clear-Sky  Index, $\mbox{CSI}=0.8$,
and the  time/spherical-varying zenith angle $A_{Z}(t,\theta_{1})$,  computed in radians from
$$A_{Z}(t,\theta_{1})= \arccos\left(\sin(\theta_{1}) \right) \sin(\theta_{2}(t)) + \cos(\theta_{1})\cos(\theta_{2}(t))\cos(\theta_{3}(t)),$$
\noindent  for  the values of the polar angle $\theta_{1}\in [0,\pi]$, where $\theta_{2}(t)$ is the time-varying declination angle, i.e.,  $\theta_{2}(t)=
\frac{23.45 \pi}{180}\sin\left(\frac{2\pi (t- 80)}{\mathcal{T}}\right),$
 and $\theta_{3}(t)=\frac{\pi t}{\mathcal{T}}$, with $\mathcal{T}=365$ for the unrestricted model (see Figure \ref{f1app} at the left), and $\mathcal{T}=183$ for the restricted model  (see Figure \ref{f1app} at the right), corresponding to the period September-February in autumn-winter.

\begin{figure}[h]
\begin{center}
\includegraphics[width=6cm,height=5.5cm]{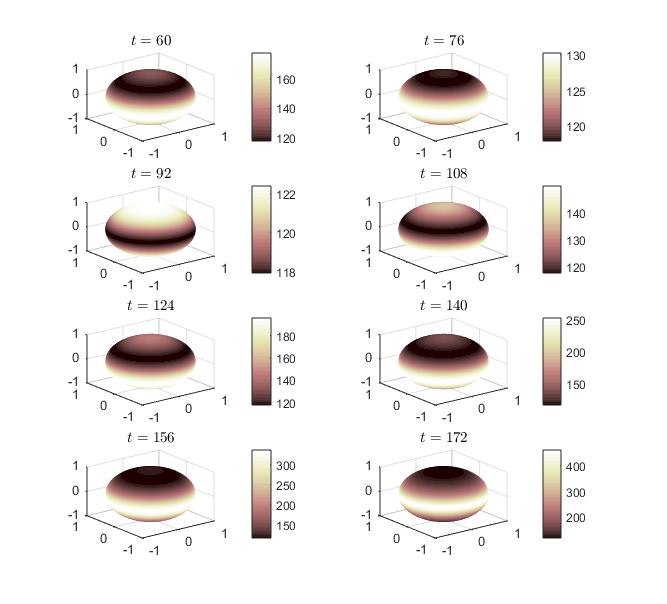}
\includegraphics[width=6cm,height=5.5cm]{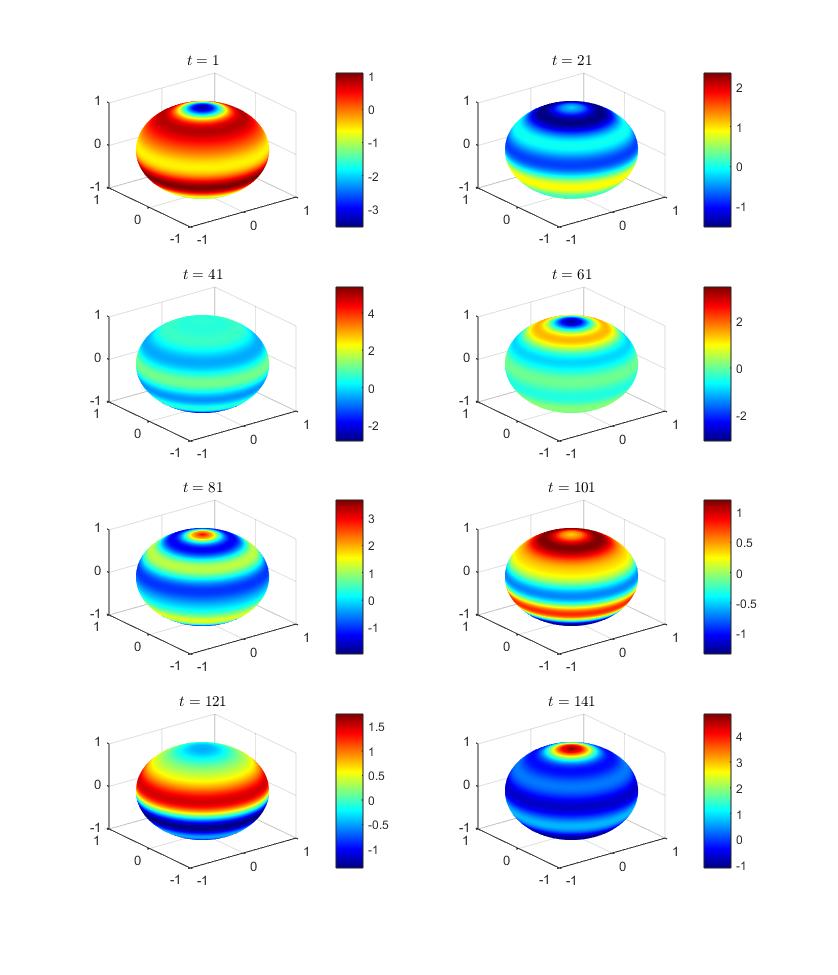}
\end{center}
\caption{\scriptsize{\emph{Temporal spherical functional regressor}. Zonal isotropic approximation to atmospheric pressure at high cloud bottom, from  inverted equation (\ref{ephsi2}),  $\mathbb{D} = \left\{60, 76, 92, 108, 124, 140, 156, 172\right\}\subset \mathbb{T}$ (left).  Unstructured spatiotemporal error term, $\mathbb{D} = \left\{1, 21, 41, 61, 81, 101, 121, 141\right\} \subset \mathbb{T}$ (right)}}
 \label{f1app3}
\end{figure}

 In Spatial Statistics,  atmospheric pressure at high cloud bottom is often leveraged as a core functional covariate  to predict, via a negative exponential link function,  downward surface solar radiation flux maps, since drastically reduces prediction variance comparing with simple linear regression modelling.  Thus, we consider the following radiative transfer model of negative exponential type, representing nonlinear relationship between  downward solar radiation flux $I_{S}(t,\theta_{1}),$ and observed spatiotemporal covariate $P_{A}(t,\theta_{1})$,  providing  high cloud bottom atmospheric pressure, at time $t$ and spherical position defined from polar angle $\theta_{1},$ assuming, as before, axial isotropy  (see  Figure \ref{f1app3} at the left):

\begin{equation}I_{S}(t,\theta_{1})=S_{0}\exp\left(- T_{I}(t,\theta_{1})(p-P_{A}(t,\theta_{1})) \right),\quad  t\in [0,183],\label{ephsi2}
\end{equation}

\medskip

\noindent where $S_{0}=  829.5$ W/$m^{2}$ is the Effective Attenuated/Broadband Solar Constant. This value represents an effective flux constant, tailored to the specific transfer model  (\ref{ephsi2}), which  approximates  the mean flux at the  top of the high cloud, considering  the  usual cirrostratus altitude  interval $(6000,12000)$ in meters.  The parameter $T_{I}(t, \theta_{1})=\frac{I_{O}(t, \theta_{1})}{p_{0}\cos(A_{Z}(t, \theta_{1}))}$ represents the thickness/extinction coefficient at the bottom of high cloud, with  $p_{0}=1013.25$ being the   reference atmospheric surface pressure. Here,  $p>P_{A}(t,\theta_{1} )$  denotes   the target surface  pressure  level where downward flux is computed. We will consider $p=p_{0}=1013.25$ such that the  vertical pressure thickness, i.e., atmospheric pressure gradient $\nabla P_{A}$, is  of 100hPa, which is the typical value for
  cirrostratus (also valid for spissatus or thick cirrus clouds). Note that, for  cirrostratus, the opacity index  $T_{I}(t, \theta_{1})$ values are usually in the interval $(0.0005,0.0045)$. As commented before, an  additive temporal LRD  spherical functional random effect  (in the restricted spherical  Gneiting class) is incorporated  to model  uncertainty arising from  these  structural simplifications,  and the lack of information by unobserved smoothly spatial  varying exogenous/endogenous covariates, displaying strong dependence in time.
 An  unstructured  spatiotemporal noise represents
 measurement errors  (see Figure \ref{f1app3} at the right).

\begin{remark}
In radiative transfer modelling (as in  EcRad in ERA5), atmospheric opacity is represented by total optical depth, computed as an integral over the atmospheric column, in terms of atmospheric covariates like cloud liquid and ice water path, aerosol optical depth, and water vapor column.
\end{remark}

\begin{figure}[h]
\begin{center}
\includegraphics[width=8cm,height=6.5cm]{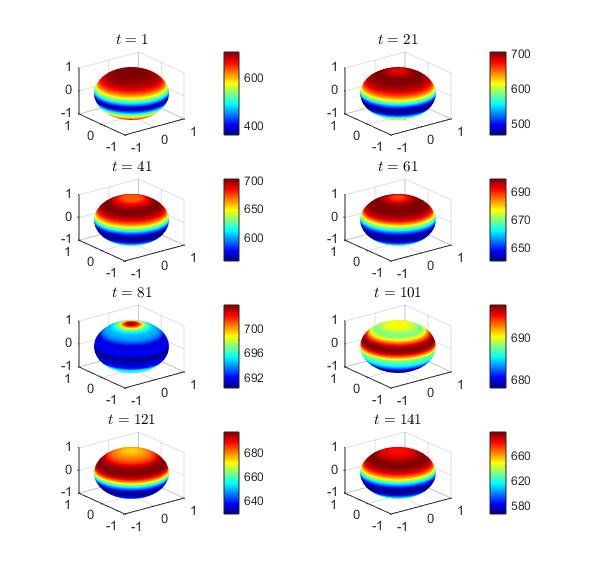}
 \end{center}
\caption{\scriptsize{One realization of the axial (zonal) isotropic temporal spherical functional observation process, $\mathbb{D} = \left\{1, 21, 41, 61, 81, 101, 121, 141\right\}\subset \mathbb{T}$}}
 \label{f1app5a}
\end{figure}

\begin{figure}[h]
 \begin{center}
\includegraphics[width=6cm,height=5.5cm]{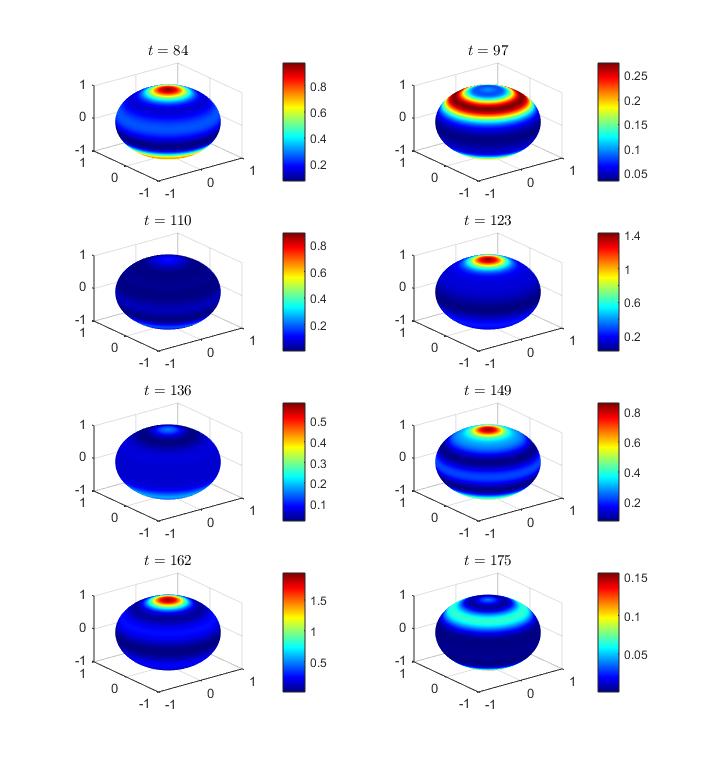}
\includegraphics[width=6cm,height=5.5cm]{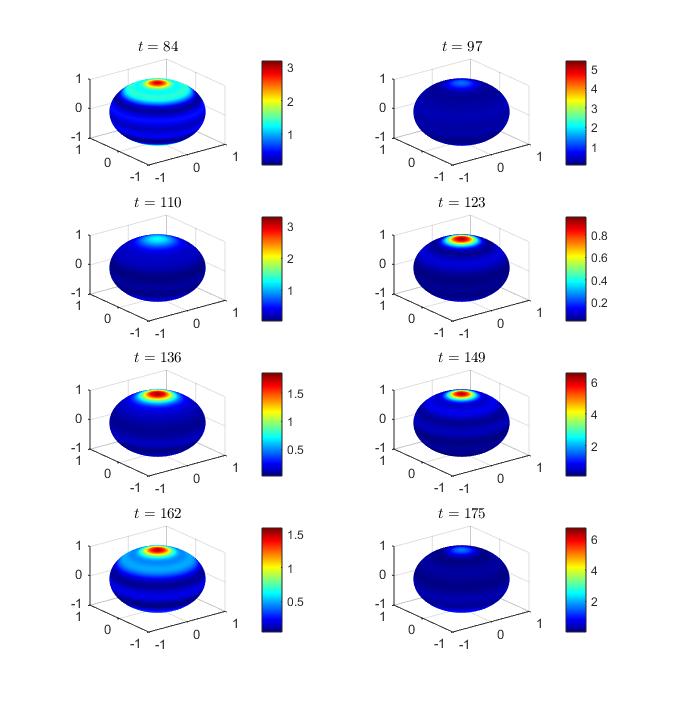}
\includegraphics[width=6cm,height=5.5cm]{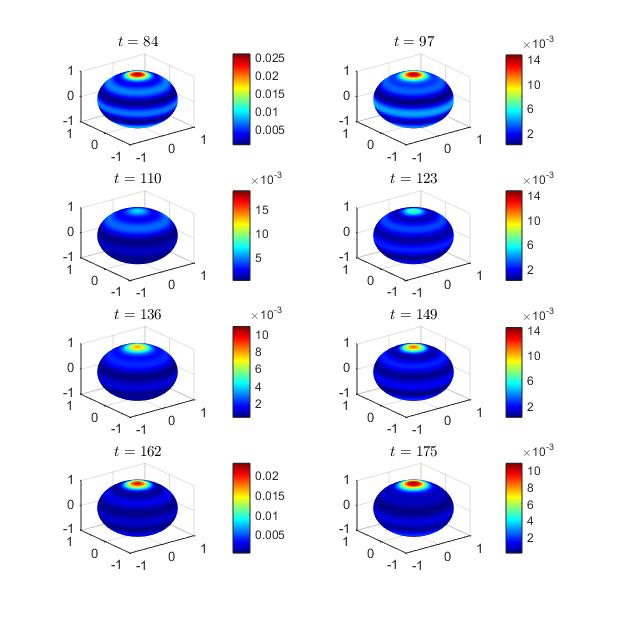}
\includegraphics[width=6cm,height=5.5cm]{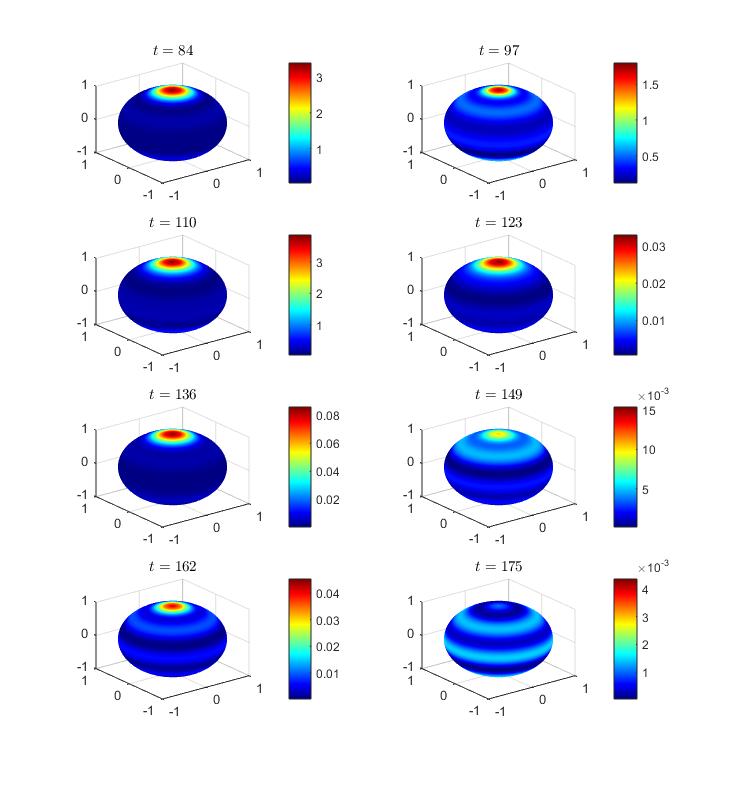}
 \end{center}
\caption{\scriptsize{5-fold cross-validation  errors, $\mathbb{D} = \left\{84, 97, 110, 123, 136, 149, 162, 175\right\}\subset \mathbb{T}$, for $T=183, N=180, TR=8, M=40, R=100,$ EBFGP based nonparametric series posterior  generalized   least-squares estimator (top-left) and linear functional predictor, non-incorporating the posterior covariance operator structure (top-right). For  $T=183, N=180, TR=8, M=60, R=200,$ EBFGP based nonparametric series posterior  generalized   least-squares estimator (bottom-left) and linear functional predictor, non-incorporating the posterior covariance operator structure (bottom-right)}}
 \label{f1app5}
\end{figure}

The performance of EBFGP posterior predictor  is tested by implementing 5-fold cross-validation in an infinite-dimensional framework,
from a generated simple random functional  sample of the conditional observation FGP model  (\ref{obGP}), based on physical equations  (\ref{ephsi})-(\ref{ephsi2}) (see Figure \ref{f1app5a}).
Note that   persistence in time and spatial dependence are incorporated  by adopting a functional sampling framework.  See \linebreak Section \ref{FGPeabb} providing  more details about the  steps involved in the implementation of the cross validation algorithm.  It is well-known that cross-validation techniques avoid overfitting and overestimation.  EBFGP regression approach  is  benchmarked against a linear functional predictor, non-incorporating the posterior covariance operator structure   (see Figure \ref{f1app5}, where  spherical functional 5-fold cross-validation quadratic errors  are  plotted).  One can observe a better performance of the EBFGP based  nonparametric series posterior  generalized   least-squares estimator  implemented in Step 9 below. Also,  for both functional predictors,  better results are observed when  the  size $M$ of the sample  from the hyperparameter  vector prior  distribution, and the  size $R$ of the simple random functional  sample  of spatiotemporal random object $Y_{t,\boldsymbol{\theta }(t),\sigma (t)}(\cdot )$ in (\ref{obGP}),  increase  (see Figure \ref{f1app5}).

\subsection{Validation of the FGP regression based predictor}
\label{FGPeabb}
This section provides a summary of the main steps of the $5$-fold cross validation algorithm implemented  from the generated simple random sample of the conditional observation FGP model  (\ref{obGP}), based on physical equations (\ref{ephsi}) and (\ref{ephsi2}).

\medskip

\begin{itemize}
\item[] \textbf{Step 1.}   Implementation of equations (\ref{ephsi}) and (\ref{ephsi2}).

\medskip

\item[] \textbf{Step 2.}  Generation of  a-priori samples of the covariance operator  hyperparameter vector $\boldsymbol{\theta }$  of the FGP, and of the   time-varying variance parameter $\sigma^{2}(t)$ of the additive functional white noise process (see equations (\ref{uet})-(\ref{eq22})).

    \medskip

\item[] \textbf{Step 3.} Given the hyperparameter values  from the prior  in  Step 2, a global simple random sample  of conditional observed  FGP  (\ref{obGP}) is generated. These generations are obtained  by additive random perturbation of the transfer equation (\ref{ephsi2}), in terms of a truncated temporal  LRD spherical   functional random effect, and a heteroscedastic  functional white noise.  High cloud bottom atmospheric pressure enters, in this transfer equation, as the observed temporal  spherical functional covariate. This covariate is  computed from the  inversion of (\ref{ephsi2}), in terms of  the downward solar radiation flux values, calculated from  the   generative physical model (\ref{ephsi}). The time-varying truncated Karhunen-Lo\'eve expansion (\ref{klexp}) is implemented in the generation of the temporal LRD functional random effect.

    \medskip

    \item[] \textbf{Step 4.} Integer-uniformly-distributed  based random splitting into training and target  functional  samples of the  global functional  sample  generated in Step 3 is performed, at each iteration of the $5$-fold cross validation.

        \medskip

        \item[] \textbf{Step 5.} Computation of the marginal likelihood of the data, given by  integral (\ref{eqmca}), is achieved via Monte Carlo numerical integration, based on the corresponding  training functional sample of conditional observation FGP  (\ref{obGP}), at each iteration of the $5$-fold cross validation.

            \medskip

            \item[] \textbf{Step 6.} Computation of the ML-II estimates $(\widehat{\boldsymbol{\theta}}_{t},\widehat{\sigma^{2}}(t))$ of the hyperparameter vector $(\boldsymbol{\theta},\sigma^{2}(t))$ is performed, based on the corresponding training functional samples at each iteration of the $5$-fold cross validation.

                \medskip

                \item[] \textbf{Step 7.}  The truncated projected  posterior functional mean of the predictive probability distribution, based on the computed  ML-II hyperparameter estimates  obtained  in Step 6, and the observation of the target functional sample, is  computed  at each iteration of the $5$-fold cross validation.

                    \medskip

                    \item[] \textbf{Step 8.}  The truncated projected   posterior covariance operator is calculated, based on the computed  ML-II hyperparameter estimates  in Step 6, and the observation of the target functional sample at each iteration of the $5$-fold cross validation.

                        \medskip

                        \item[] \textbf{Step 9.} A nonparametric series posterior  generalized   least-squares estimator  is implemented to approximate the  nonlinear exponential type regression function, from the posterior mean and covariance operator computed in Steps 7 and 8,  at each iteration of the $5$-fold cross validation.

                            \medskip

                             \item[] \textbf{Step 10.} The  $5$-fold cross validation error is finally computed.
\end{itemize}
\section{Final comments}
\label{FC}
The EBFGP  methodology  on manifolds, proposed in the time-varying angular spectral domain, leads to an important dimension reduction  in the time-adaptive  functional  regression context.   The hierarchical structure provides the proposed models with great versatility, in the purely point spectral domain, where the characteristic function based identification  of infinite-dimensional tight Gaussian measures with the infinite product of one-dimensional Gaussian measures  plays a fundamental role. This Gaussian context also  allows us to define suitable FGP priors, as well as posteriors conditioned to the ML-II hyperparameter estimates. Consistency of the posterior  predictor  can be analyzed in terms of  suitable truncation schemes, depending on the functional sample size (see \cite{Bosq00}). This posterior asymptotic analysis will be addressed in a subsequent work, being applied to the definition of time-adaptive credibility regions in an infinite-dimensional tight Gaussian measure framework. The   conditions on consistency  we will derive, involving a suitable functional sample-size-varying  truncation scheme, will mainly refer to the   sparsity (atom separation), and velocity decay of the  FGP posterior  angular  spectrum at each time, leading to an  efficient finite-dimensional  approximation of such  functional credibility regions.

The proposed FGP regression algorithm is implemented  in the pure point spectral domain, leading to an important dimension reduction, with substantial simplification of  mathematical complexity, since under isotropy (invariance under the group of isometries of the manifold), all the involved covariance kernels are diagonalized in terms of the Laplace--Beltrami eigenfunction basis $\left\{ S_{n,j}^{(d)}, j=1,\dots , \Gamma (n,d),\ n\in \mathbb{N}_{0}\right\}$. Sections \ref{ss}-\ref{sda} illustrate the fact that, even under  a weaker isotropy condition like  axial (zonal) isotropy, one can consider the truncated spherical harmonic basis, constituted by the  zonal functions associated with each one of the selected Laplace--Beltrami operator  eigenspaces, leading to a substantial dimension and  computational cost  reduction.  Specifically,  in the implemented scripts,  we  work in terms of the projections into these zonal functions.   However, the evaluation of the underlying  spatiotemporal parametric covariance family over the generated  prior hyperparameter samples is time consuming, involving a substantial increasing  of the runtimes, and strongly affecting the execution time of  the Empirical Bayes approach.

\medskip

\noindent \textbf{Acknowledgements}
This work has been partially supported by grants (M.D. Ruiz-Medina, J.M. Angulo, A.E. Madrid) PID2025-171143NBI00, (M.D. Ruiz-Medina, A. Torres-Signes) PID2022-142900NB-I00 and (A.E. Madrid, J.M. Angulo) PID2021-128077NB-I00, funded by MICIU / AEI/10.13039/501100011033 / ERDF, EU, and grant (M.D. Ruiz-Medina, J.M. Angulo) CEX2020-001105-M, funded by MICIU / AEI/10.13039/501100011033.


\begin{thebibliography}{4}
\bibitem{Adler81}
Adler, R.J. (1981). \emph{The geometry of random fields}. Wiley, Chichester, U.K.
\bibitem{Andersen}
Andersen, M.R., Vehtari, A., Winther, O. and Hansen, L.K. (2017). Bayesian inference for spatio-temporal spike-and-slab priors. \emph{J. Mach. Learn. Res.}  \textbf{18}, 1--58.
\bibitem{Bosq00}
Bosq, D. (2000).
\newblock {\em {Linear processes in function spaces: theory and applications}}.
  Volume 149.  {\em Lecture Notes in Statistics}.
\newblock {Springer-Verlag, New York}.
\bibitem{Carlin}
Carlin, B.P., Gelfand, A.E. and Banerjee, S. (2014). \emph{Hierarchical modelling and analysis of spatial data}. Chapman and Hall/CRC, London.
\bibitem{DautrayLions85}
Dautray, R. and   Lions, J.L. (1985). \emph{Mathematical analysis and
numerical methods for science and technology. Volume} \textbf{3}.
\emph{Spectral theory and applications}. Springer, New York.
\bibitem{DaPrato}
Da Prato, G.   and  Zabczyk,  J. (2009). \textit{Second Order Partial Differential
Equations in Hilbert Spaces}. Cambridge University Press, Cambridge.
\bibitem{Deisenroth}
Deisenroth, M.P., Fox, D. and Rasmussen, C.E.  (2015). Gaussian processes for data-efficient learning in robotics and control. \emph{IEEE Trans. Pattern Anal. Mach. Intell.} \textbf{37}, 408--423.
\bibitem{Diggle13}
Diggle, P.J. (2013). \emph{Statistical analysis of  spatial and spatiotemporal point patterns}. Chapman and Hall, London.
\bibitem{Gine1975} Gin\'e,  E.  (1975). The addition formula for the eigenfunctions of the Laplacian. \textit{Adv. Math.}  \textbf{18}, 102--107.
\bibitem{Gneiting02}
Gneiting, T. (2002).  Nonseparable,
stationary covariance functions for space-time data. \textit{J. Amer. Stat.
Assoc.}  \textbf{97}, 590--600.
\bibitem{Henning}
Hennig, P., Osborne, M.A. and  Girolami, M. (2015).  Probabilistic numerics and uncertainty in computations. \emph{Proc.  R. Soc. A: Math. Phys. Eng. Sci.} \textbf{471}, 20150142.
\bibitem{Hensman17}
Hensman, J., Durrande, N. and Solin, A.  (2017). Variational Fourier features for Gaussian processes. \emph{J. Mach. Learn. Res.} \textbf{18}, 5537--5588.
\bibitem{Hoffmann26}
 Hoffmann, A.G.,  Ekstrom, C.T. and  Jensen, A.K. (2026).
Computationally efficient multilevel Gaussian process
regression for functional data observed under completely
or partially regular sampling designs. \emph{TEST} \textbf{35},  211-231.
\bibitem{Kaufman10}
Kaufman, C.G. and Sain, S.R. (2010).  Bayesian functional ANOVA modeling using Gaussian
process prior distributions. \emph{Bayes. Anal.} \textbf{5}, 123--149.
\bibitem{Lazaro10}
L\'azaro-Gredilla, M., Qui\~{n}onero-Candela, J., Rasmussen, C.E. and Figueiras-Vidal, A.R. (2010). Sparse spectrum Gaussian process regression. \emph{J. Mach.  Learn. Res.} \textbf{11}, 1865--1881.
\bibitem{Lindgren22}
 Lindgren, F., Bolin, D. and Rue, H. (2022). The SPDE approach for Gaussian and non-Gaussian fields: 10 years and still running. \emph{Spatial Stat.} \textbf{50}, 100599.
 \bibitem{MaMalyarenko} Ma, C. and   Malyarenko, A.  (2020). Time-varying isotropic vector random fields on compact two-point homogeneous spaces. \textit{J. Theor. Probab.} \textbf{33}, 319--339.
 \bibitem{Neal97}
Neal, R.M.  (1997). Monte Carlo implementation of Gaussian process models for Bayesian regression and classification. arXiv preprint physics/9701026.
\bibitem{Ovalle23}
Ovalle-Mu\~noz, D.P.  and  Ruiz-Medina, M.D. (2024). LRD spectral analysis of multifractional functional time series on manifolds. \textit{TEST}
\textbf{33},   564--588.
\bibitem{Ovalle25}
Ovalle-Mu\~noz, D.P.  and  Ruiz-Medina, M.D. (2025). Climate change analysis from LRD manifold functional regression. \textit{Stoch.  Environ. Res. Risk Assess.} \textbf{39}, 1555--1580.
\bibitem{Quinonero05}
 Qui\~{n}onero-Candela, J. and  Rasmussen, C.E.  (2005). A unifying view of sparse approximate Gaussian process regression. \emph{J. Mach.  Learn. Res.} \textbf{6}, 1939--1959.
 \bibitem{Williams06}
Rasmussen, C.E. and Williams, C.K. (2006). \emph{Gaussian processes for machine learning}. MIT
Press, Cambridge.
 \bibitem{Rice91}
 Rice, J.A. and Silverman, B.W. (1991).
 Estimating the mean and covariance structure nonparametrically when the data are curves. \emph{J. R.  Stat. Soc. Ser. B} \textbf{53}, 233--243.
\bibitem{Riutort23}
Riutort-Mayol, G., B\"urkner, P.-C., Andersen, M.R., Solin, A. and  Vehtari, A. (2023). Practical Hilbert space approximate Bayesian Gaussian processes for probabilistic programming. \emph{Stat. Comput.} \textbf{33}, 17.
\bibitem{RuizMedina2022}
Ruiz-Medina, M.D. (2022). Spectral analysis of multifractional LRD functional time series. \textit{Fract. Calc. Appl. Anal.} \textbf{25}, 1426--1458.
\bibitem{RMCR25}
Ruiz-Medina, M.D. and Crujeiras, R.M. (2025). Testing LRD in the spectral domain for functional time
series in manifolds. \emph{Electron. J. Stat.} \textbf{19},  3601--3642.
\bibitem{RMCR26}
Ruiz-Medina, M.D. and Crujeiras, R.M. (2026). Long range dependence spectral  testing from  discretely observed  functional time series in manifolds. Accepted for publication in \emph{J. Multiv. Anal.} arXiv:2511.00518v2.
\bibitem{Solin13}
S\"arkk\"a, S., Solin, A. and Hartikainen, J. (2013). Spatiotemporal learning via infinite-dimensional Bayesian filtering and smoothing: a look at Gaussian process regression through Kalman filtering. \emph{IEEE Signal Process. Mag.} \textbf{30}, 51--61.
\bibitem{Simon}
 Simon, B. (2005). \emph{Trace Ideals and Their Applications}. Mathematical Surveys and Monographs, Vol. 120. American Mathematical Society, Providence, RI.
\bibitem{Solin20}
Solin, A. and  S\"arkk\"a, S. (2020). Hilbert space methods for reduced-rank Gaussian process
regression. \emph{Stat. Comput.} \textbf{30}, 419--446.
\bibitem{Triebel78}
Triebel, H.  (1978). \textit{Interpolation theory, function spaces, differential operators}, North-Holland, Amsterdam.
 \bibitem{Triebel97}
Triebel,  H. (1997).  \textit{Fractals and spectra}.
Birk\"auser, Basel.
\bibitem{Wilson14}
Wilson, A.G., Gilboa, E., Nehorai, A. and Cunningham, J.P. (2014). Fast kernel learning for
multidimensional pattern extrapolation. \emph{Adv.  Neural Inf.  Process. Syst.} \textbf{4},  3626--3634.\end{thebibliography}
\end{document}